\documentclass[10pt,journal,compsoc]{IEEEtran}
\pdfoutput=1

\usepackage{tikz, pgfplots} 
\usepackage{tabularx}
\usepackage[font=small,labelfont=bf]{caption}
\pgfplotsset{ 
  legend style = {font=\small\sffamily}, 
  label style = {font=\small\sffamily} 
}
\usepackage{scalefnt}
\usepackage{cite}
\usepackage[cmex10]{amsmath}
\ifCLASSOPTIONcompsoc
  \usepackage[caption=false,font=scriptsize,labelfont=sf,textfont=sf]{subfig}
\else
  \usepackage[caption=false,font=footnotesize]{subfig}
\fi
\usepackage{url}

\usepackage{booktabs} 
 \usepackage{amsmath}
\usepackage{amsfonts} 
\usepackage{graphicx}

\begin{document}
\newlength\figureheight
\newlength\figurewidth
\bstctlcite{IEEEexample:BSTcontrol}
%
\title{The Effect of Distortions on the Prediction of Visual Attention}
%
%
%

\author{Milind~S.~Gide, Samuel~F.~Dodge, and~Lina~J.~Karam,~\IEEEmembership{Fellow,~IEEE}
\thanks{M.~S.~Gide, S.~F.~Dodge and L.~J.~Karam are with the School
of Electrical, Computer and Energy Engineering, Arizona State University, Tempe, AZ, 85287-5706 USA~(e-mail:~mgide,~sfdodge,~karam@asu.edu)}

}

%
%

\markboth{Submitted to the IEEE Transactions on Pattern Analysis and Machine Intelligence }%
{Shell \MakeLowercase{\textit{et al.}}: Bare Demo of IEEEtran.cls for Journals}
%



\maketitle
\begin{abstract}
Existing saliency models have been designed and evaluated for predicting the saliency in distortion-free images. However, in practice, the image quality is affected by a host of factors at several stages of the image processing pipeline such as acquisition, compression and transmission. Several studies have explored the effect of distortion on human visual attention; however, none of them have considered the performance of visual saliency models in the presence of distortion.  
Furthermore, given that one potential application of visual saliency prediction is to aid pooling of objective visual quality metrics, it is important to compare the performance of existing saliency models on distorted images. In this paper, we evaluate several state-of-the-art visual attention models over different databases consisting of distorted images with various types of distortions such as blur, noise and compression with varying levels of distortion severity. This paper also introduces new improved performance evaluation metrics that are shown to overcome shortcomings in existing performance metrics. We find that the performance of most models improves with moderate and high levels of distortions as compared to the near distortion-free case. In addition, model performance is also found to decrease with an increase in image complexity.
\end{abstract}


%
\IEEEpeerreviewmaketitle

\section{Introduction}
%
%
%
%

\IEEEPARstart{V}{isual} attention (VA) is the broad area of research that focuses on understanding the mechanisms by which the human visual system (HVS) filters out the huge amount of visual information collected by the retina. It enables humans to focus on the most relevant and important information that helps them perform complex cognitive tasks such as object recognition and scene interpretation in real-time. This process is thought to be a combination of two inter-dependent mechanisms, a bottom-up pre-attentive one, that relies on low-level data driven features and a top-down one that is cognitive, high-level and task dependent. 

Over the past few decades, a lot of effort has gone into understanding the underlying bottom-up mechanisms of VA physiologically, psychologically as well as computationally. 
As a result, a large number of visual attention models (see~\cite{borjireview} for a detailed review) have been proposed that mimic pre-attentive vision. Each of these approaches computes the saliency of a pixel giving a measure of how much that pixel stands out from its surroundings. Traditionally, these VA models are compared with eye-tracking fixations to gage their performance.
A number of comprehensive comparative evaluation studies have been conducted in recent years \cite{borjieval,judd2012benchmark}, to benchmark existing state-of-the-art models in terms of their ability to predict human fixations obtained from eye-tracking data.  
The ability of a VA model to select relevant areas in images is very useful for data selection and reduction in applications such as object recognition~\cite{obj_recog_VAapp1},~\cite{obj_recog_VAapp2} segmentation~\cite{segment_VA_application1},~\cite{segment_VA_application2} amongst others.  

Another application of VA that has received considerable attention is image quality assessment (IQA). Images are subjected to a number of distortions at different stages of the image processing pipeline from acquisition to transmission. Thus, it becomes necessary to algorithmically detect the level of distortion affecting an image as perceived by a human observer. Several objective IQA algorithms have been proposed in the literature~\cite{Wang2004,Sheikh2006,Wang2003}. Most of these existing popular IQA algorithms pool information over all pixels to compute an image quality score. However, according to VA studies, humans tend to fixate only on certain salient areas in the image. Thus, pooling information only over perceptually important regions can improve performance. Based on this, a number of recent studies have explored the use of VA to improve IQA metrics~\cite{Engelke2011,Ninassi2007,Larson2008,HantaoLiu2011,Barland2006,Moorthy2009,Sadaka2008,Gkioulekas2010}. Some of  them~\cite{Ninassi2007,Larson2008,HantaoLiu2011} have used human eye-fixations to obtain saliency information that is used to weight the IQA metric scores at each pixel. Some others,~\cite{Barland2006,Moorthy2009,Sadaka2008,Gkioulekas2010}, use VA models to improve the pooling of the IQA metrics. Except for one study~\cite{Ninassi2007}, all others report a significant increase in IQA metric performance when combined with VA based pooling~\cite{Engelke2011},~\cite{Moorthy},~\cite{Liu}. Some other studies~\cite{TUDInteractions,Vu2008,Ninassi2007} show that human fixations significantly change when the task given to subjects changes from free-viewing to quality assessment which is consistent with results reported in the well known study by Yarbus~\cite{yarbus1967eye} which indicates significant changes in human scanpaths with a change in tasks. More recently, Mathe and  Sminchisescu~\cite{NIPS2013_5196} also demonstrated the influence of different tasks on gaze under an object and context recognition framework.
Given the potential of VA models in IQA algorithms in which the VA models are applied on distorted images, there are currently no evaluation studies in the literature that analyze the effect distortions have on the performance of state-of-the-art VA models. Also, given that almost all existing VA models have been developed for images without distortions, it is necessary to know how their performance changes due to distortions. This paper presents a comprehensive evaluation of 20 state-of-the-art VA models over distorted images obtained from two existing databases that provide eye-tracking data for a number of images with varying content and distortion types and levels. The change in performance over different levels and types of distortions is analyzed. As part of this evaluation, this paper also proposes new evaluation metrics that overcome shortcomings in existing performance metrics. 

In this paper we provide the first comprehensive comparative evaluation of VA models on distorted images. We first give an overview of 20 state-of-the-art models that we chose to evaluate in Section~\ref{sec:ExistingModels}. Next, in Section~\ref{sec:PerfEvalMetrics}, we describe the performance metrics used to evaluate VA models and show shortcomings in existing performance metrics and propose new metrics that improve these shortcomings. Finally, in Section~\ref{sec:PerfEvalwithDistortion}, we show and discuss the performance evaluation results and conclude the paper in Section~\ref{sec:Conclusion}.


 

\section{Visual Attention Models}
\label{sec:ExistingModels}

\begin{table*}[t]
\scriptsize
  \centering
   \caption{List of Evaluated Models: W and H represent the width and height of the input image and are set to 768 and 512, respectively, for determining computation time, except when noted otherwise.}
   \resizebox{0.7\textwidth}{!}{
    \begin{tabular}{ccccc}
    \toprule
    \textbf{Category} & \textbf{Model Name} & \textbf{Time Taken (seconds)} & \textbf{Platform} & \textbf{Resolution} \\
    \midrule
    \textit{Cognitive } & GAFFE~\cite{GAFFE} & 5.5794 & Matlab & $W \times H$ \\
          & ITTI1~\cite{Itti}  & 0.4160 & Matlab & $W \times H$ \\
          & ITTI2~\cite{Itti} & 4.9322 & Matlab & $W \times H$\\
          &       &       &       &  \\
    \textit{Graph Based} & GBVS~\cite{GBVS}  & 0.3515 & Matlab + MEX & $W \times H$ \\
    \textbf{} & GBVSNoCB~\cite{GBVS} & 0.5496 & Matlab + MEX & $W \times H$ \\
    \textbf{} &       &       &       &  \\
    \textit{Information Theoretic} & AIM~\cite{AIM}   & 22.9508 & Matlab & $\frac{W}{2} \times \frac{H}{2}$ \\
          & HouNIPS~\cite{HouNIPS} & 0.1960 & Matlab & $W \times H$ \\
          & GR~\cite{GR}    & 0.2926 & Matlab & $W \times H$ \\
          & SDSRG~\cite{SDSR}  & 1.2863 & Matlab & $W \times H$ \\
          & SDSRL~\cite{SDSR}  & 1.3263 & Matlab & $W \times H$ \\
          & SUN~\cite{SUN}   & 2.8115 & Matlab & $246 \times 331$ \\
          & Torralba~\cite{AudeTorralba} & 0.7909 & Matlab & $W \times H$ \\
          &       &       &       &  \\
    \textit{Machine Learning Based} & Judd~\cite{MIT} & 298   & Matlab + MEX+OpenCV & $200 \times 200$ \\
    \textbf{} &       &       &       &  \\
    \textit{Spectral} & FES~\cite{FES}   & 0.3593 & Matlab & $W \times H$\\
          & FTS~\cite{FTS}   & 1.7581 & Matlab & $W \times H$ \\
          & SigSal~\cite{SigSal} & 0.3852 & Matlab & $64 \times 48$ \\
          & SpectRes~\cite{SpectRes} & 0.0862 & Matlab & $W \times H$ \\
          &       &       &       &  \\
    \textit{Other} & AWS~\cite{AWS}   & 5.1655 & Matlab & $\frac{W}{2} \times \frac{H}{2}$\\
          & BMS~\cite{BMS}   & 2.4805 & Matlab + MEX & $W \times H$ \\
          & Context~\cite{Context} & 37.2912 & Matlab + MEX & $W \times H$\\
          & CovSal~\cite{CovSal} & 14.9193 & Matlab  & $W \times H$ \\
          & RandomCS~\cite{RandomCS} & 3.4434 & Matlab & $W \times H$ \\
          & SIMCoarse~\cite{SIM}   & 27.6926 & Matlab +  MEX + OpenCV & $W \times H$ \\
          & SIMFine~\cite{SIM}   & 28.7456 & Matlab +  MEX + OpenCV & $W \times H$ \\
    \bottomrule
    \end{tabular}%
}
 \label{table:models}%
\end{table*}%

 The VA models developed over the past few decades have mostly been targeted at modeling the bottom-up mechanisms of VA. This is because, top-down VA is highly dependent on the internal state of the observer and depends on factors like observer history which is very difficult to quantify and model. The bottom-up VA models work on the concept of a ``saliency map''  as proposed by Koch and Ullman~\cite{koch1987salmap}.  The bottom-up VA models can be categorized based on the type of features used to obtain the saliency maps.
 We choose 20 state-of-the-art models that well represent different categories.
 The models also vary in complexity and correspondingly their resource intensiveness.  The complete list of models along with the category, the time taken for each model to process a 768 x 512 image, the platform used, and the corresponding output resolution for each model is provided in Table~\ref{table:models}. To calculate the computation time, all the models were run on the same PC using Matlab. For some models, MEX files and OpenCV libraries are used to speed up execution times, as indicated in Table~\ref{table:models} under the platform column. An overview of the different model categories and corresponding models is provided in Sections~\ref{subsec:cognitivemodels} to \ref{subsec:othermodels}
  
 \subsection{Cognitive Models}
 \label{subsec:cognitivemodels}
 Most of the VA models are based on cognitive principles directly or indirectly. However, some models are more heavily based on neurophysiological findings and give an insight into the neurological mechanisms of VA.
  One of the first basic models was developed by Itti~et~al.~\cite{Itti} based on the feature integration theory (FIT)~\cite{treisman1980feature}.
 In this model, the image is first decomposed into color, intensity and orientation channels. Then each of these channels is represented by Gaussian pyramids to obtain the center-surround responses that can be used to enhance features that are different from their neighbors. The different channels are combined across scales and features with normalization to give the saliency maps, the peaks of which correspond to visually more salient locations. We use two implementations of this model: ITTI1 refers to the original model proposed by Itti et~al.~\cite{Itti} which is obtained via the implementation at \url{http://www.saliencytoolbox.net}, while ITTI2 refers to a higher resolution version of the same with Gaussian blurring and a center bias added to it, the implementation of which is found at \url{http://www.klab.caltech.edu/~harel/share/gbvs.php}. Another HVS based approach that models the mechanism of foveation is the Gaze Attentive Fixation Finding Engine (GAFFE) proposed by Rajashekar et al.~\cite{GAFFE}. GAFFE is a foveated analysis framework that uses four low-level local image saliency features, namely luminance, contrast, and bandpass outputs of both luminance and contrast, to form saliency maps and predict gaze fixations. It uses a sequential process in which the stimulus is foveated at the current fixation point and saliency features are obtained from patches from the foveated image to predict the next fixation point. The fixation points thus obtained can be converted to a saliency map by convolving them with a 2D Gaussian kernel with full-width-at-half-maximum equal to the visual angle corresponding to the foveal region.
 \subsection{Information Theory Based Models}
 \label{subsec:informationtheorybasedmodels}
 These models are based on the principle that regions of high saliency maximize information and are described by rarity of features. Bruce et al. (AIM)~\cite{AIM} used Shannon's self-information measure~\cite{ShannonsInformationMeasure} to get an indication of how unexpected the local image content is relative to its surroundings and saliency is taken to be proportional to the information content in a region in comparison with its surrounding. Zhang et al. (SUN)~\cite{SUN} used a similar approach by defining bottom-up saliency as the self-information of visual features; however, the self-information is described in terms of natural image statistics. Another similar saliency measure comprising of bottom-up and top-down components of saliency was proposed by Torralba (Torralba)~\cite{torralba2003modeling} in which the bottom-up saliency is defined to be inversely proportional to the probability of features. Hou and Zhang (HouNIPS)~\cite{HouNIPS} used the incremental coding length (ICL) concept to compute saliency wherein the objective is to maximize the entropy of the visual features. Mancas et al. (GR)~\cite{GR} also use self-information measures of the mean and variance of local image intensity, and the assumption that rare features attract visual attention. In the Saliency Detection by Self Resemblance (SDSR) approach proposed by Seo and Milanfar~\cite{SDSR}, local descriptors known as local regression kernels are computed over an image to measure the likeness or resemblance of a pixel with its surroundings.  Saliency is then computed using this self-resemblance measure at each point, which measures the statistical likelihood of a saliency value given its neighborhood. The global (SDSRG) and local (SDSRL) variants of the model are evaluated separately. In the model proposed by Tavakoli et al. (FES)~\cite{FES}, a center-surround approach using a Bayesian framework is adopted.  Saliency at a point is considered to be a binary random variable which is 1 if the point is salient and 0 otherwise.  The probability of a pixel being salient given feature values is considered to be the saliency.

 \subsection{Spectral Analysis Based Models}
 \label{subsec:spectralmodels}
 Models described in this category compute saliency in the frequency domain.  Hou and Zhang (SpectRes)~\cite{SpectRes} developed a method that models saliency by using the logarithm of the spectrum of an image and computing its deviation from the average log spectrum of several natural images, which acts as prior information. 
  Achanta et al. (FTS)~\cite{FTS} developed a frequency tuned saliency approach in which an image is first converted to the CIE Lab color space and then low-pass filtered with binomial kernel to eliminate fine texture details and noise from which the mean of the image is subtracted to get the final high resolution saliency map with  uniform salient regions. Another approach by Hou et al.~\cite{SigSal} develops an image signature based on the sign function of the discrete cosine transform (DCT) which is used to predict visually conspicuous locations in an image.

 \subsection{Graph Based Models}
 \label{subsec:graphbasedmodels}
 A graphical model is a framework which describes the conditional dependence of neighboring elements or nodes. Harel et al.~\cite{GBVS} introduced the Graph Based Visual Saliency (GBVS) model in which  feature maps are extracted at multiple resolutions for different image features like intensity, color, and orientation.  A fully connected graph, representing a Markov chain is constructed considering pixels of the feature maps as nodes. The weight between two nodes is assigned proportional to the similarity of feature values and spatial distance between them.
  On solving the Markov chain, the equilibrium distribution accumulates mass at nodes that have high dissimilarity with their surrounding nodes and the activation measure captures pairwise contrast.  
 The GBVS model inherently promotes high saliency values near the center of the images. For a fairer comparison, we also include the GBVS method with the center bias removed for performance evaluation purposes. This is treated as a separate model denoted by GBVSNoCB in the results. 
 
 \subsection{Machine Learning Based Models}
 \label{subsec:mlmodels}
 Machine learning can be used to model VA by using eye-tracking data to train classifiers to predict salient vs. non-salient regions in an image. One such approach by Judd et. al (Judd)~\cite{MIT} uses several low-level features as input to a linear SVM classifier including those in the original model by Itti et al.~\cite{Itti}, Gabor filtered values and a global gist based saliency map along with higher level top-down features like face, horizon and object detectors. The machine learning problem is formulated as a linear regression in which the optimal weights for each feature are learned by training a linear regression model on a training set for which the eye-tracking data is used.  
 
 \subsection{Other Models}
 \label{subsec:othermodels}
 Some of the newer approaches to VA are hybrid in the sense that they belong to more than one category or they cannot be readily classified into any of the aforementioned categories. 
 The Adaptive Spectral Whitening (AWS) VA model proposed by Garcia-Diaz et al.~\cite{AWS}
 uses variability in local energy as a measure of saliency. The image is first transformed into the CIE Lab color space and the luminance channel is decomposed by both a multi-orientation as well as multi-resolution filter bank whereas the a and b channels are decomposed into a multi-resolution filter bank. These responses are first decorrelated using PCA and then locally averaged to obtain the saliency map.
 In another approach by Goferman et. al~\cite{Context}, called Context Based Saliency (Context), local as well as global distinctiveness is considered as a contributing factor towards saliency. 
  The Saliency based on Image Manipulation (SIM) approach proposed by Margolin et al.~\cite{SIM}  introduces an object probability map  that groups together locations of high attention that are in close proximity to each other and also a pixel reciprocity measure that boosts the saliency of the pixels in close vicinity to highly salient locations. The coarse (SIMCoarse) and fine (SIMFine) versions of the model are evaluated separately.
 In the Boolean Map Saliency (BMS) approach proposed by Zhang and Scarloff~\cite{BMS}, the Gestalt principle of surroundedness is used to compute saliency.  In this approach, color-based feature maps are thresholded by variable thresholds to obtain boolean maps which are then used to obtain connected regions. Regions that exhibit surroundedness or are closed (i.e., regions not attached to the image boundary) are assigned a binary value of 1 and others are given a value of 0. Maps for a given threshold value are normalized using the $L_2$ norm and then averaged over different thresholds and different features to get the final saliency map. 
 Another approach by Vikram et al. called Random Center Surround Saliency (RandomCS)~\cite{RandomCS} deviates from the Feature Integration theory based approaches \cite{Itti,GAFFE} by utilizing a stimulus-influence-based approach in which saliency at a point is influenced by other randomly chosen points in the image. The saliency influence at a point by another point is described in terms of a contrast by taking the absolute difference in pixel values and normalizing with the distance between the two points. This process is repeated over the L, a and b components of the image. The Covariance Saliency (CovSal) method proposed by Erdem et al.~\cite{CovSal} uses covariance matrices of simple image features in the form of region covariance descriptors to capture local image structures and provide non-linear integration of the features. Saliency is obtained by finding the distance between the covariance matrices of a central region with its surrounding neighborhood regions using a non-Euclidean distance measure that is based on the eigenvalues and eigenvectors of the covariance matrices.	

\section{Performance Evaluation Metrics}
\label{sec:PerfEvalMetrics}

The most common evaluation strategy is to analyze the performance of a VA model in terms of several metrics. However, existing saliency metrics will often give different rankings to different models (see \cite{borjieval}). It is unclear as to which metrics are more useful. In this section we  describe existing metrics, propose four new metrics and show their advantage over the existing metrics.

We use the following common notation for all metrics. $G$ is the ground truth saliency map (also called fixation density map), $S$ is the saliency map produced by a model. ($x_i$,$y_i$) is the position of the $i$th  ground-truth fixation point, where $0 \leq i \leq N-1$.

\subsection{Existing Performance Metrics for VA models}
\label{existing_metrics}


\subsubsection{Linear Cross Correlation (CC)}
The linear cross correlation is a metric that measures the strength of a linear relationship between two variables.  When used for evaluating VA models, it is used in the following form: 

\begin{equation}
  \textnormal{CC}(G,S) = \frac{\sum_{x,y}(G(x,y)-\mu_G)(S(x,y)-\mu_S)}{\sqrt{\sigma_G^2\sigma_S^2}} 
  \label{equ:CC}
\end{equation}
where $G$ represents the fixation density map obtained by convolving the fixation map (which is a map of 1s at fixation points and 0s at non-fixated points) with a 2D Gaussian, and $S$ represents the predicted saliency map. $\mu_S$, $\mu_G$ are the means and $\sigma_G$, $\sigma_S$ are the standard deviations of the predicted saliency map and fixation density map, respectively. The $\textnormal{CC}$ value can vary between -1 and 1, with a value of 1 denoting a strong positive linear relationship, a value of -1 representing a strong negative linear relationship and a value of 0 denoting absence of correlation.

\subsubsection{Similarity (SIM) } 
The similarity measure was presented by Judd et al.~\cite{Judd_2012} as part of a benchmark for VA models. It corresponds to the histogram intersection between the predicted saliency map histogram $H_S$ and the fixation density map histogram $H_G$. 
It is defined as follows:
\begin{equation}
\textnormal{SIM}(S,G) =  \sum_{i=1}^{N} min(H_S(i),H_G(i));
\end{equation}
where $N$ is the number of histogram bins.

\subsubsection{Earth's Mover Distance (EMD)}
\label{EMD}
The Earth Mover's Distance (EMD)~\cite{Rubner} is a cross-bin histogram dissimilarity measure that doesn't require the histogram domains to be aligned. The EMD treats two histograms or distributions as two different ways in which the same amount of dirt has piled up over a region, and the distance is given by the minimum cost of changing one pile of dirt into the other. An improved variant of the EMD metric proposed by Pele and Werman~\cite{EMDHat} (denoted by $\widehat{\textnormal{EMD}}$), that works as a metric even for histograms that are not normalized, is given by:
\begin{eqnarray}
\label{eq:EMD}
\widehat{\textnormal{EMD}}(H_{S},H_{G}) = \left(\min_{\{f_{i,j}\}}\sum_{i,j}f_{i,j}d_{i,j}\right) + \nonumber\\
\lvert\sum_{i}H_{G}(i)-\sum_{j}H_S(j)\rvert	\max_{i,j} d_{i,j}  \\ 
\text{s.t} \ f_{i,j} \geq 0, \ \sum_{j}f_{i,j} \leq H_{G}(i), \ \sum_{i}f_{i,j} \leq H_{S}(j), \nonumber\\ 
\sum_{i,j} f_{i,j} = \min(\sum_{i}H_G(i),\sum_j H_S(j)) \nonumber 
\end{eqnarray}
where $H_S$ and $H_G$ represent the histograms of the predicted and ground-truth saliency maps, respectively, $f_{i,j}$ represents the amount transported from bin $i$  to bin $j$, and $d_{i,j}$ is the ground distance between bin $i$ and bin $j$ in the histograms.  For probability histograms  (total mass of 1), the $\textnormal{EMD}$ and $\widehat{\textnormal{EMD}}$ values are equal. $\widehat{\textnormal{EMD}}$ was used by Judd et al. in their VA  benchmark~\cite{Judd_2012} with the ground distance saturated to a maximum value as proposed in~\cite{FastEMD}.

\subsubsection{Area under the Receiver Operating Characteristic Curve using Ground Truth Fixations (AUC-F)}

For this evaluation method which is based on the receiver operating characteristics (ROC), the predicted saliency map is first thresholded. This thresholded map is denoted by $S_t$, where $t$ represents the threshold value which ranges from $0$ to $255$ for an 8-bit depth saliency map.

Then, the true positive rate (TPR) and false positive rate (FPR) are determined using the thresholded map and the ground-truth fixation points. The true positive rate is computed by counting the fraction of ground-truth fixation points that are properly detected in the thresholded saliency map as follows:

\begin{equation}
\textnormal{TPR(t)} = \frac{1}{N} \sum_{i=0}^{N-1} S_t(x_i,y_i)
\label{eq:TPR}
\end{equation}

To find the $\textnormal{FPR}$ we must compute the fraction of non-fixated points that are labeled as salient in the thresholded predicted saliency map. One approach \cite{borjieval} for computing the $\textnormal{FPR}$ is to have  the number of non-fixated points be equal to the number of fixated points. $N$ random points are chosen from the possible set of non-fixated pixels in the ground-truth. We use the notation $(\hat{x}_i,\hat{y}_i)$ to denote these non fixated points. Because these points are chosen randomly, the AUC test is performed 100 times each with different random non-fixated points, and the final result is the average of the individual trials.  Let $l$ be the index of the current trial and $L$ be the total number of trials. Also, let $(\hat{x}_{i,l},\hat{y}_{i,l})$, $1\leq i \leq N$ be the set of N non-fixated points at the $l^{th}$ trial. The $\textnormal{FPR}$ is then given by:

\begin{equation}
\textnormal{FPR}(t)=\frac{1}{L} \sum_{l=1}^{L} \textnormal{FPR}(t,l)\,,
\end{equation} 
 where
 \begin{equation}
 \textnormal{FPR}(t,l) = \frac{1}{N} \sum_{i=1}^{N} S_t(\hat{x}_{i,l},\hat{y}_{i,l})
 \end{equation}

By computing the {\em TPR} and {\em FPR} for every value of the threshold $t$ (from 0 to 255) we obtain a receiver operating characteristic curve (ROC). The area under this curve (AUC) gives a measure of the performance of the saliency map. A value of 1 means that the saliency map perfectly predicts the ground-truth, whereas a value of 0.5 is no better than random chance.

\subsubsection{Area under the Receiver Operating Characteristic Curve using Ground Truth Saliency (AUC-S)}
%

\begin{figure}[t]
	\centering
	\tiny
	\begin{tabular}{ccc}
		\includegraphics[width=0.1\textwidth]{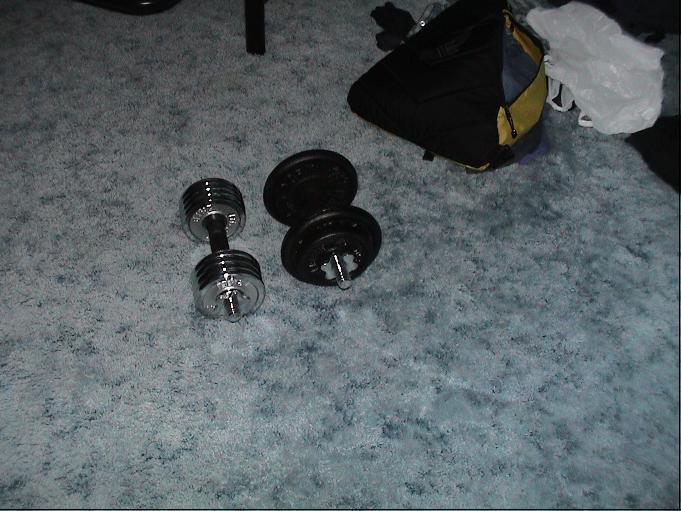} & 
		\includegraphics[width=0.1\textwidth]{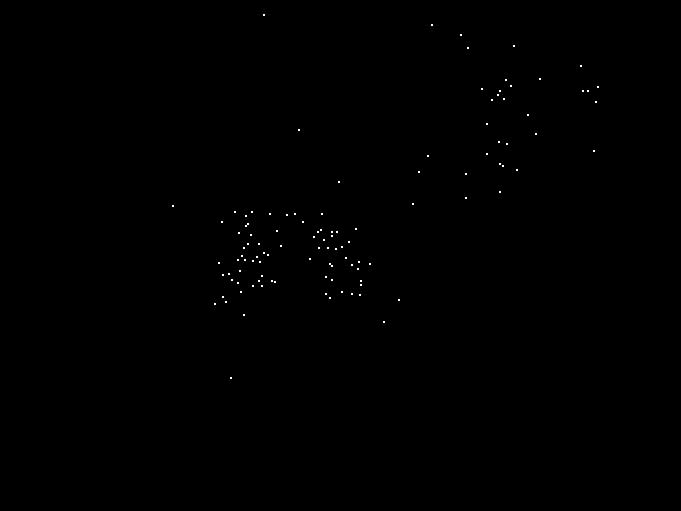} & 
		\includegraphics[width=0.1\textwidth]{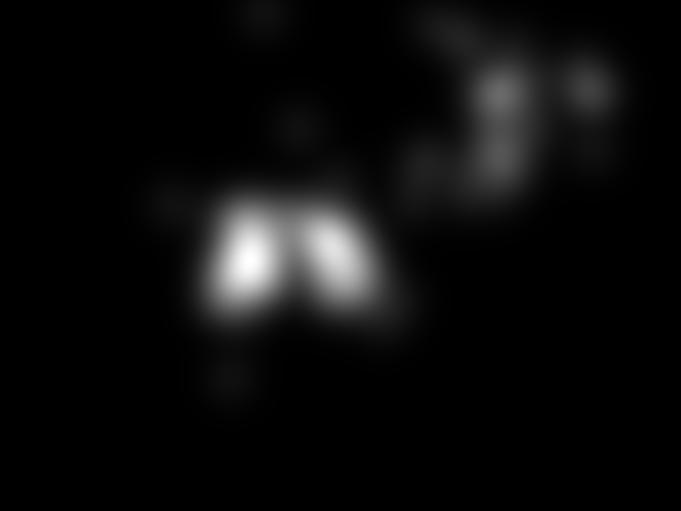} \\ 
		\textbf{(a)} Original Image & 
		\textbf{(b)} Fixation Map & 
		\textbf{(c)} Ground-Truth \\		
	\end{tabular}
	\vfill
	\begin{tabular}{cc}
		\includegraphics[width=0.1\textwidth]{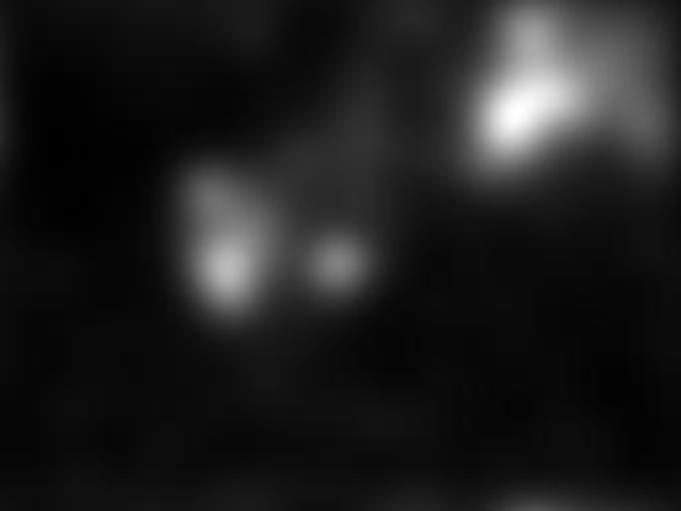} & 
		\includegraphics[width=0.1\textwidth]{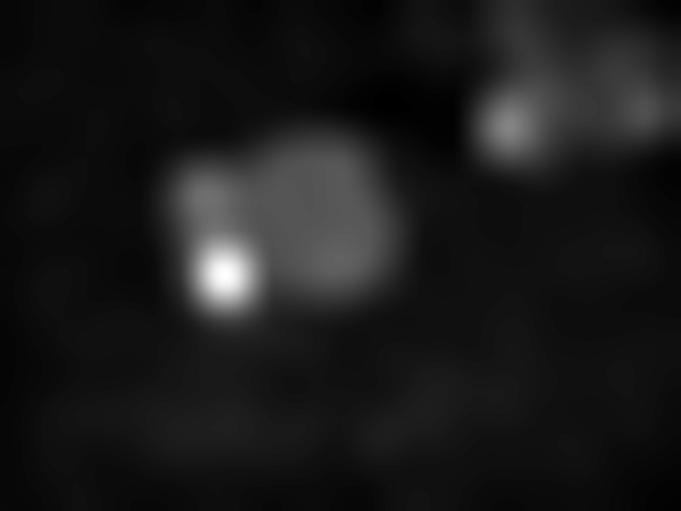} \\ 
		\textbf{(d)} AWS~\cite{AWS}, AUC-F~=~0.8882 &  
		\textbf{(e)} BMS~\cite{BMS}, AUC-F=0.8735 \\
		
	\end{tabular}
	\caption{Disadvantage of AUC-F in terms of evaluating a model's ability to detect the most salient regions.}
	\label{fig:FixVsSal}	
\end{figure}

The AUC-F metric gives equal importance to each ground truth fixation. As a result, the highly salient regions that attract more fixations per unit area in an image are treated the same as other less salient regions that attract fewer fixations per unit area. Figure~\ref{fig:FixVsSal} shows an example from the Toronto database~\cite{AIM} that illustrates this issue. As seen from the fixation map and the ground truth saliency map, which is obtained by placing 2D Gaussians with full-width-at-half-maximum (FWHM) corresponding to one degree of visual angle on each of the fixation points, the area with the dumbbells has a higher density of fixations as compared to the area with the bag. However, AUC-F will treat fixations near the dumbbells the same as those near the bag. As a consequence, some VA models whose saliency map incorrectly predicts the bag to be a more salient object than the dumbbells (e.g., AWS~\cite{AWS} in Figure~\ref{fig:FixVsSal}(d)) will get a higher AUC-F score than VA models (e.g., BMS~\cite{BMS} in Figure~\ref{fig:FixVsSal}(e)) whose saliency map is more in agreement with ground truth data.
A saliency map variant of the AUC (denoted by AUC-S) that uses the ground truth saliency maps instead of fixations attempts to rectify this problem. For this metric,  first, a fixed threshold $T=  0.5*\sigma$, where $\sigma$ is the standard deviation of $G$, is applied to $G$ to get a binary map $G_T=\{0,1\}$. Then, a variable threshold $t$ is applied to $S$ to get a binary map $S_t\in\{0,1\}$. Let $I_t=G_T \land S_t$ denote the intersection of both binary maps, $n(X)$ the number of non-zero points in a binary map $X$, and $l(X)$ the total number of points in $X$. The $TPR$ and $FPR$ can then be computed as follows:
\begin{eqnarray}
\textnormal{TPR(t)} = n(I_t)/n(G_T) \\
\textnormal{FPR(t)} = \frac{n(S_t)-n(I_t)}{l(G_T)-n(G_t)}
\end{eqnarray}

The TPR and FPR are computed for every value of the threshold $t$ to obtain a receiver operating characteristic curve (ROC) the area under which gives AUC-S. Like the AUC-F, for AUC-S, a value of 1 indicates the best prediction performance, whereas a value of 0.5 indicates chance performance. 

\subsubsection{Shuffled AUC (SAUC)}
The main limitation of the AUC-F and AUC-S metrics is that they are center biased. When using these metrics, a model's score can be improved simply by multiplying the predicted saliency map by a large Gaussian blob such that the predicted saliency near the edge of the image will be reduced\cite{SUN}. In fact, a simple Gaussian blob will often perform better than some saliency models using the AUC-F and AUC-S metrics. 
This center bias is inherent to eye tracking experiments. This occurs for two main reasons. First, the most salient regions of an image are often located near the center of the image. This is called the \emph{photographer's bias}. Secondly, the nature of eye tracking experiments yields a framing effect, where it is natural to look at the center of the scene because this point allows our vision system to capture the most information \cite{Tatler2005}.

\begin{figure}[t]
	\centering
	\tiny
	\begin{tabular}{cccc}
		\includegraphics[width=0.1\textwidth]{figures/AIMSAUC_failcase.jpg} &
		\includegraphics[width=0.1\textwidth]{figures/AIMfdm3.jpg} &
		\includegraphics[width=0.1\textwidth]{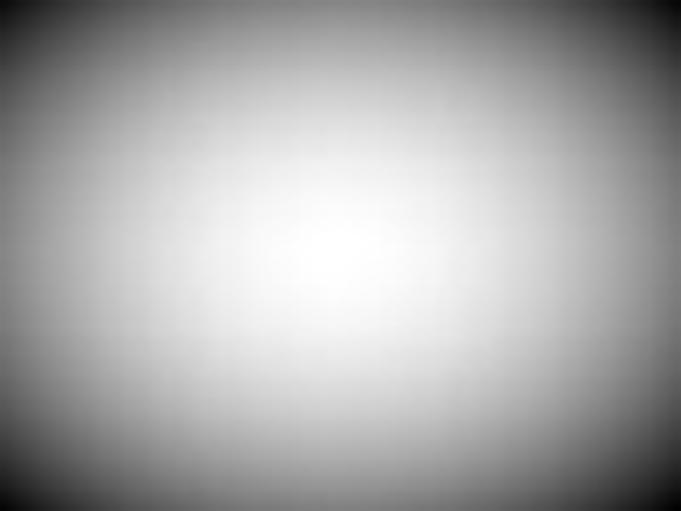} & 
		\includegraphics[width=0.1\textwidth]{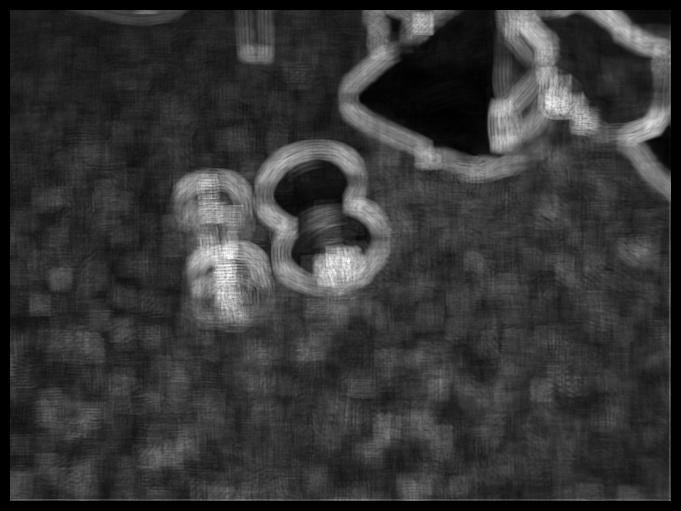} \\
		\textbf{(a)} Original Image &
		\textbf{(b)} Ground Truth & 
		\textbf{(c)} Centered Gaussian &
		\textbf{(d)} AIM~\cite{AIM} 
	\end{tabular}
	\footnotesize
	\begin{tabular}{|c|c c|}
		\hline
		& (c) & (d) \\ \hline
		AUC-F~\cite{borjieval} & 0.8079 & 0.7551 \\	
		AUC-S~\cite{Moorthy_Bovik} & 0.7991 & 0.7207 \\
		CC~\cite{Judd_2012} & 0.4399 & 0.3747 \\ 
		SAUC~\cite{borjieval} & 0.5823 & 0.6679 \\ \hline
			
	\end{tabular}
	\caption{Center bias problem in existing metrics that is rectified by the shuffled metrics like SAUC.}
	\label{fig:CBinAUCSandAUCF}
\end{figure}

The shuffled AUC (SAUC) metric was introduced to account for this center bias \cite{SUN}. An unbiased metric should give a low score to the central Gaussian blob model. The difference in the SAUC metric as compared to the AUC metric is that the non-fixated points $(\hat{x}_i,\hat{y}_i)$ are randomly chosen from the distribution of all the fixation points from all the other images in the dataset. Consequently, this will give the Gaussian blob model an SAUC of approximately 0.5, which is equivalent to chance. The center-bias issue for AUC-F and AUC-S is illustrated in Figure~\ref{fig:CBinAUCSandAUCF} which shows the AUC-S, AUC-F, CC and SAUC scores for a centered Gaussian Blob in comparison with the AIM model~\cite{AIM} for an image taken from the Toronto database~\cite{AIM}. The AUC-F, AUC-S and CC scores for the centered Gaussian are higher than that for the AIM model because of the center bias. The SAUC score results, for the centered Gaussian, in a score that is closer to 0.5 and that is lower than the AIM model's score.

%
%

%
\begin{figure}[t]
	\centering
	\tiny
	\begin{tabular}{ccc}
		\includegraphics[width=0.1\textwidth]{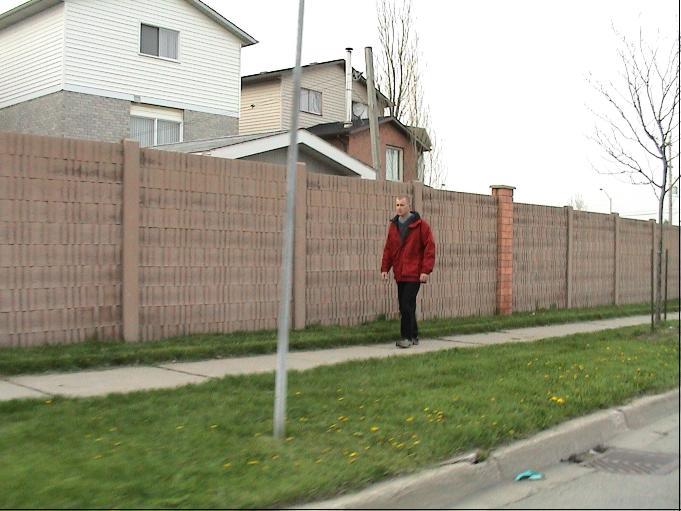} & 
		\includegraphics[width=0.1\textwidth]{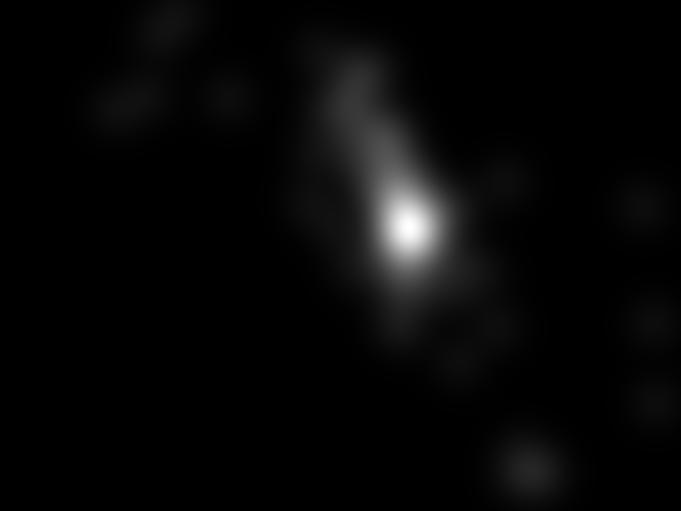} &
		\includegraphics[width=0.1\textwidth]{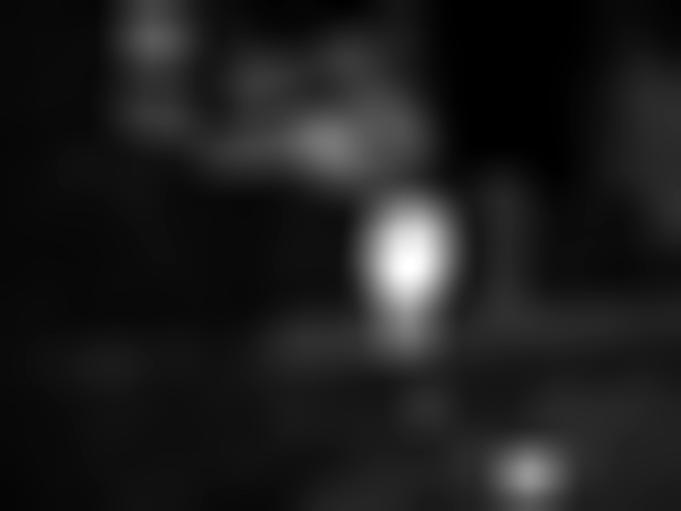} \\
		\textbf{(a)} Original Image &
		\textbf{(b)} Ground-Truth &
		\textbf{(c)} BMS~\cite{BMS} \\		
	\end{tabular}
	\footnotesize
	\begin{tabular}{|c|c c|}					
		\hline
		& (b) & (c) \\ \hline
		SAUC~\cite{borjieval} & 0.8052 & 0.8234 \\	
		NSS~\cite{NSS} & 4.1829 & 2.7881 \\	\hline		
	\end{tabular}
	\caption{Image from Toronto~\cite{AIM} database that shows the advantage of the  NSS metric over the SAUC metric. In terms of SAUC, the BMS model is incorrectly given a higher score than the ground truth saliency map.}
	\label{fig:snss_sauc_aim}
\end{figure}

\subsubsection{Normalized Scanpath Saliency (NSS)}

The normalized scanpath saliency (NSS) computes a score using ground-truth fixation points and the predicted saliency map as follows:
\begin{equation}
\text{NSS(x,y)} = \frac{1}{N}\sum_{i=1}^N \frac{S(x_i, y_i) - \mu_s}{\sigma_s}
\end{equation}
\noindent where $S$ is the saliency map to be tested, $(x_i,y_i)$ is the location of a fixation in the ground-truth eye tracking data, $\mu_s$ and $\sigma_s$ are, respectively, the mean and standard deviation of the predicted saliency map.

Compared to the NSS, the SAUC score, previously defined, suffers from the interpolation flaw (see~\cite{margolin2014evaluate} for more details) as a result of which models that output maps that are blurry and fuzzy tend to do better in terms of the SAUC metric. This can be observed in images with single, small salient regions and ground truth saliency maps that are sparse. One such image and its corresponding ground truth saliency map is shown in Figure~\ref{fig:snss_sauc_aim} along with the SAUC scores for the ground truth saliency map and a state-of-the-art model BMS~\cite{BMS}. The BMS model is incorrectly given a higher SAUC score than the ground truth saliency map as true positives are given more importance than false positives in the SAUC formulation. The NSS does not have this bias and has a penalty for false positives. Hence it is a more fair metric to analyze models in the case of sparse saliency maps. However, NSS suffers from the same problem as the previously introduced  AUC metrics (AUC-F and AUC-S) in that it is sensitive to center bias. To counter this center-bias we introduce a new modified NSS metric termed shuffled NSS (SNSS). The proposed SNSS is described in more details in Section~\ref{sec:ProposedMetrics}.

\subsubsection{Shuffled Kullback-Leibler divergence}
\begin{figure}[tb]        
\begin{center}
	\tiny
\begin{tabular}{c c c}            
  	\includegraphics[width=0.1\textwidth]{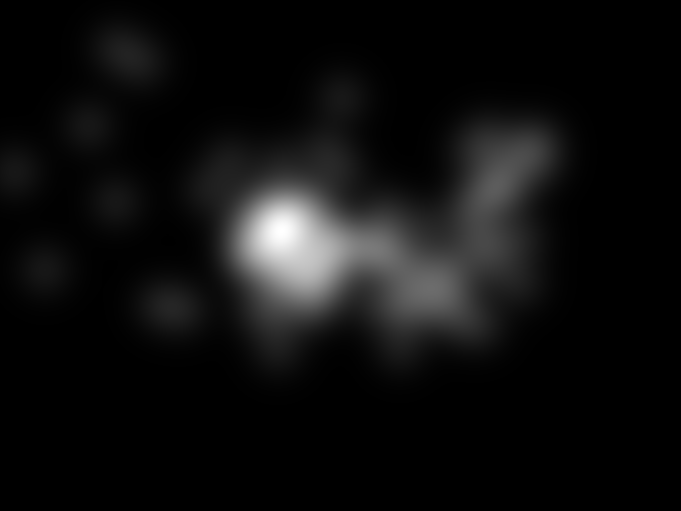} 			& \includegraphics[width=0.1\textwidth]{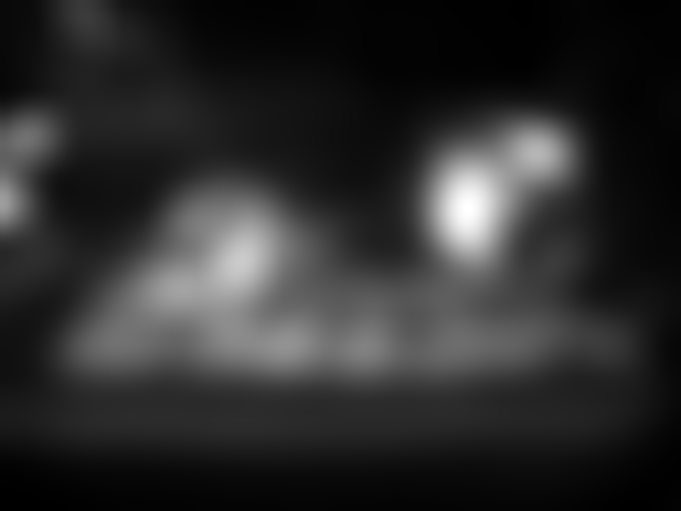} & \includegraphics[width=0.1\textwidth]{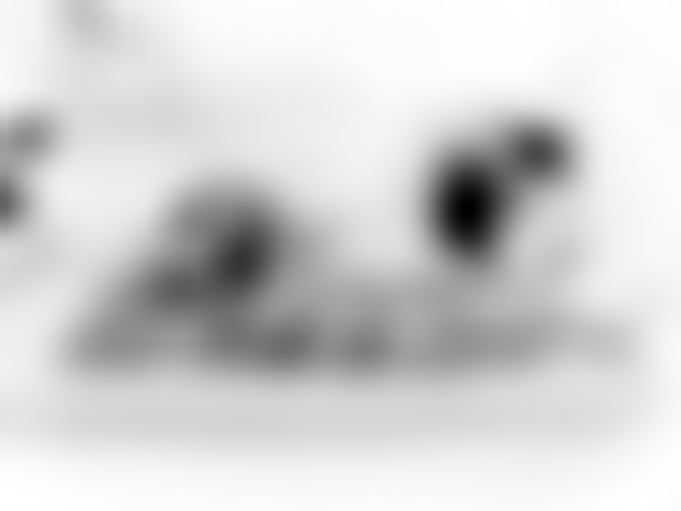}  \\
  	\textbf{(a)} Ground Truth 			& \textbf{(b)} Predicted Map S using BMS~\cite{BMS},			& \textbf{(c)} Inverted Map 	 	 \\ 
  	                & SKLD~\cite{SUN} = 9.36		& SKLD~\cite{SUN} = 9.46             \\
  	 				& Proposed SSKLD = 9.36			& Proposed SSKLD = -9.46			 \\ 
\end{tabular}
\end{center}
\caption{\textbf{Limitation of shuffled KLD from \cite{SUN}}. The original SKLD metric gives good scores for inverted saliency maps (c). Our proposed modified metric (SSKLD) gives a negative score for this inverted map.}
\label{fig:kldwrong}
\end{figure}


The Kullback-Leibler Divergence (KLD)~\cite{KLD} is widely used as a distance measure between two histograms. A shuffled KLD (SKLD) was used in~\cite{SUN} to compare histograms of the fixated and non-fixated points. As in the case of SAUC, the non-fixated points were chosen randomly from the distribution of fixation points of all other images in the dataset. If the histograms are very different then the SKLD score should be high and the saliency map is considered a good predictor of the ground-truth. 

The motivation behind including this shuffled KLD (SKLD) score in addition to the other metrics is described in~\cite{riche2013saliency}. The authors of~\cite{riche2013saliency} show that out of all other metrics the SKLD measure is the most decorrelated with the AUC and NSS measures and hence provides a non-redundant metric of performance. 

However, the SKLD metric presented in \cite{SUN} is flawed because the metric compares the distance between these histograms regardless of which histogram has a larger average value. Because of this, if the gray-level values of a predicted saliency are inverted, the SKLD metric will give the same score (Figure~\ref{fig:kldwrong}). Such inverted saliency maps should however be given a lower score as they predict the non-salient regions as salient instead of the salient regions.

We propose a modification of the SKLD metric that does not exhibit this problem. The proposed metric is referred to  as signed shuffled KLD or SSKLD for short  and is described in Section~\ref{sec:ProposedMetrics} . This metric can give negative scores if the mean of the distribution of the non-fixated points is higher than that of the fixated points. 

The KLD based measures, SKLD and SSKLD, cannot be classified as distance metrics as they do not have an upper bound and do not satisfy the triangular inequality. To resolve this issue we propose a  performance metric based on the square root of the symmetric and bounded variant of the KLD known as the Jensen-Shannon Divergence (JSD)~\cite{JSD} which can be considered as a distance measure. The proposed metric termed as the Shuffled Jensen-Shannon Distance (SJSD) is described in Section~\ref{sec:ProposedMetrics}.

\subsection{Proposed Performance Metrics for Evaluation Models}
\label{sec:ProposedMetrics}
The existing performance metrics described previously have either one or both of the following two shortcomings, center bias and false positives. Center bias is related to the tendency of some metrics to unfairly reward higher central activity and false positives relates to the tendency of the AUC metrics (AUC-F, AUC-S and SAUC) to not penalize false-positives which gives an unfairly high score to blurrier maps.  To resolve these issues we propose the following performance metrics.

\subsubsection{Shuffled Normalized Scanpath Saliency} 
\label{subsubsec:SNSS}
%
%
\begin{figure}
	\centering
	\tiny
	\begin{tabular}{cc}
		\includegraphics[width=0.1\textwidth]{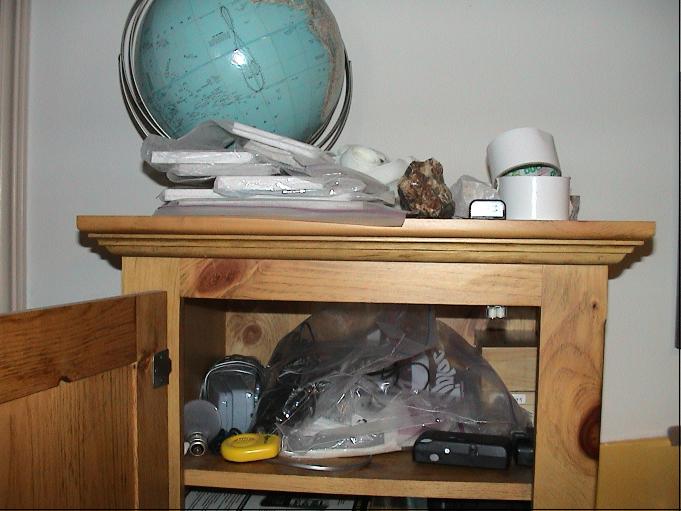} & 
		\includegraphics[width=0.1\textwidth]{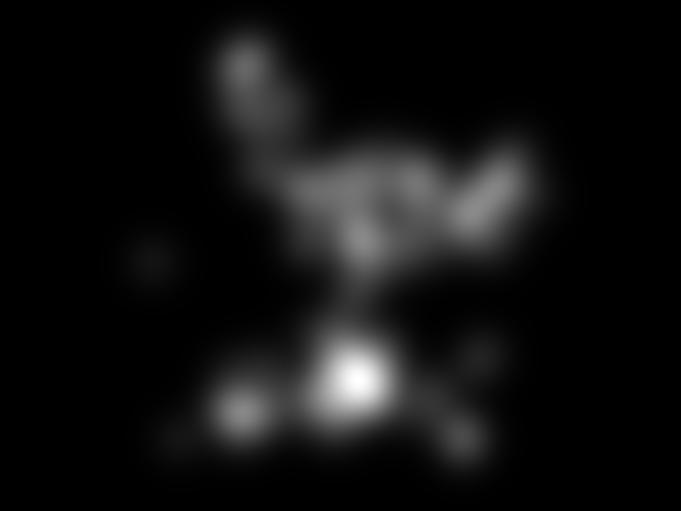} \\
		\textbf{(a)}Image &
		\textbf{(b)}Ground-truth	
	\end{tabular}
	\begin{tabular}{ccc}
		\includegraphics[width=0.1\textwidth]{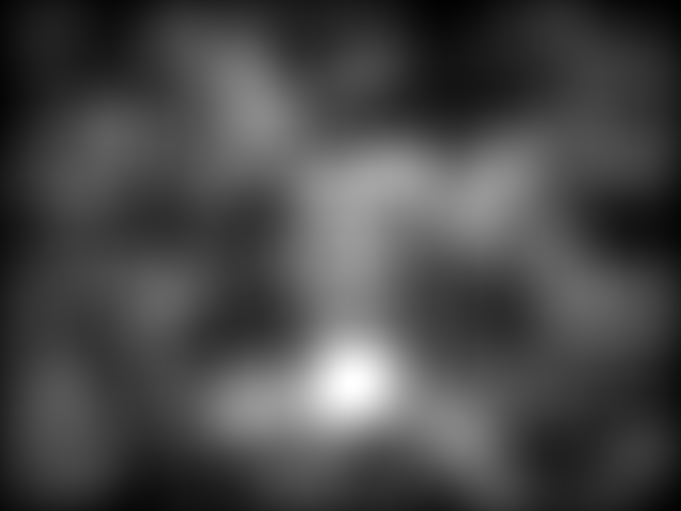} & 
		\includegraphics[width=0.1\textwidth]{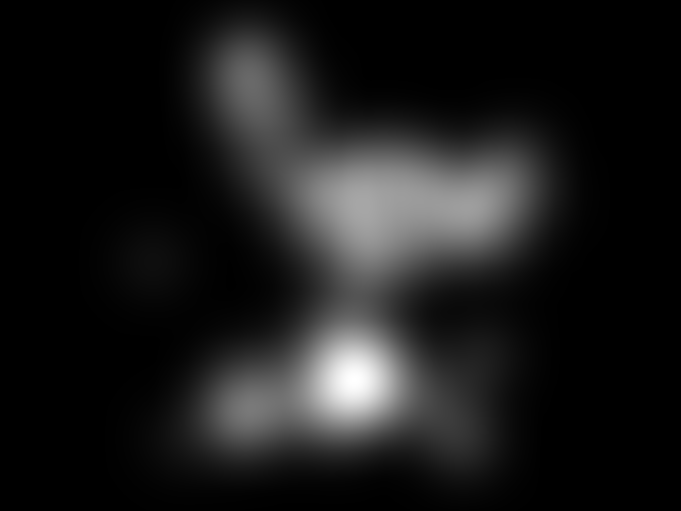} &
		\includegraphics[width=0.1\textwidth]{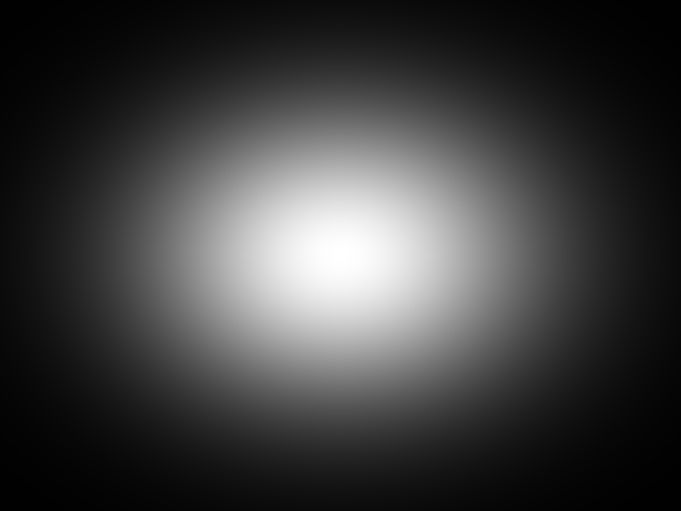} \\
		\textbf{(c)} &
		\textbf{(d)} & 
		\textbf{(e)} 
	\end{tabular}
	\footnotesize
	\begin{tabular}{| c | c c c |}
		\hline
		 & (c) & (d) & (e)\\ \hline
		 AUC~\cite{AUC} & 0.9056 & 0.9190 & 0.7993 \\
		 SAUC~\cite{borjieval} & 0.8145 & 0.8120 & 0.4603 \\
		 NSS~\cite{NSS}& 1.9630 & 2.2598 & 1.0790 \\
		 Proposed SNSS & 1.3557 & 1.6211 & -0.1400 \\ \hline		 			
	\end{tabular}
	\caption{\textbf{Advantage of the new SNSS metric}. The two saliency maps (c) and (d) have nearly identical SAUC scores, however it is clear that (d) is a ``better'' saliency map. The proposed SNSS metric that we introduce in this paper is a better measure of saliency prediction. Here (d) has a much higher SNSS than (c). Additionally, a centered Gaussian blob (e) will perform well on the NSS score, however using the proposed SNSS the same Gaussian blob receives a score close to 0.}
	\label{fig:snss}
\end{figure}

 The proposed shuffled NSS (SNSS) metric is based on the NSS~\cite{NSS} metric and is defined as follows: 
\begin{equation}
\text{SNSS} = \text{NSS}(x,y) - \text{NSS}(\hat{x},\hat{y})
\label{eq:SNSS}
\end{equation}

\noindent where $(x_i,y_i)$ and $(\hat{x}_i,\hat{y}_i)$ are, respectively, the ground-truth fixation and non-fixated points for the considered image. As before, the  non-fixated points are chosen randomly from the fixation points of all other images in the dataset. This computation is repeated 100 times and the final result is the average of these 100 trials. Figure \ref{fig:snss} shows the advantage of the SNSS metric over SAUC by creating two different synthetic saliency maps that are obtained by adding high and low amounts of background activity to the ground truth saliency map. The SAUC gives a higher score to the map with higher level of background activity whereas the SNSS correctly gives the map with lower background activity the higher score. SNSS also retains the invariance to center-bias that the SAUC has. 

\subsubsection{Signed Shuffled Kullback Leibler Divergence (SSKLD)}
The SSKLD is obtained by modifying the shuffled KLD measure by multiplying it with the  sign of the SNSS value defined in (\ref{eq:SNSS}).
Mathematically, 
\begin{equation}
	\textnormal{SSKLD} = sign\Big(\textnormal{SNSS}\left(S\right)\Big) \cdot \textnormal{KLD} \Big( H,\hat{H} \Big)
	\label{eq:hist}
\end{equation}

\noindent where $H$ is a histogram formed from $\left(x_i,y_i\right)$ and $\hat{H}$ is a histogram formed from $\left(\hat{x}_i,\hat{y}_i\right)$. Both  $H$ and $\hat{H}$ are normalized by dividing the bin frequencies with the total number of ground truth fixations. As indicated previously,  $(\hat{x}_i,\hat{y}_i)$ are chosen randomly from the ground-truth fixation points of all the other images in the dataset. KLD is the symmetric KLD distance given by:
\begin{equation}
	\textnormal{KLD}(H,\hat{H}) = \frac{ \sum_i H(i) \log \left( \frac{H(i)}{\hat{H}(i)} \right) + \sum_i \hat{H}(i) \log \left( \frac{\hat{H}(i)}{H(i)} \right)}{2}	
\end{equation}

In practice, a small $\epsilon$ is added to $\hat{H}$ and $H$ to avoid a division by zero and $\text{log}(0)$ errors.

As seen in Figure~\ref{fig:kldwrong}, the signed term in the proposed SSKLD metric enables it to penalize the inverted saliency map by giving it a negative score, whereas using the SKLD metric results in the same score for both the inverted and the non-inverted maps.

\subsubsection{Shuffled Jensen-Shannon Distance (SJSD)}
The Jensen-Shannon Divergence between two histograms $P$ and $Q$ is a symmetric divergence based on the KLD and is given by 
\begin{equation}
\textnormal{JSD}(P,Q) = \frac{D(P||M)+D(Q||M)}{2}
\end{equation}
where $M=\frac{P+Q}{2}$ is the mean of the two histograms and $D(P||Q) = \sum_i P(i)\log_2 \left( \frac{P(i)}{Q(i)}\right)$ is the non-symmetric Kullback-Leibler Divergence. When a logarithm of base 2 is used for the calculation of $D(P||Q)$, $0\le \textnormal{JSD} \le 1$. The square root of the $\textnormal{JSD}$ is a distance metric that satisfies the triangle inequality~\cite{JSDistance}. The proposed shuffled Jensen-Shannon Distance (SJSD) is defined as follows:
\begin{equation}
\textnormal{SJSD}(H,\hat{H}) = \left(\textnormal{JSD}(H,\hat{H})\right)^{\frac{1}{2}}
\end{equation}
where $H$ and $\hat{H}$ are the same as in (\ref{eq:hist}).


\begin{figure}[t]
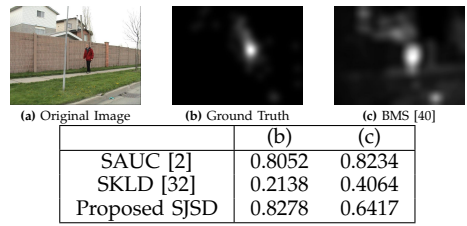

	\centering
	\tiny
	\begin{tabular}{ccc}
		\includegraphics[width=0.1\textwidth]{figures/AIMSAUC_failcase94.jpg} &
		\includegraphics[width=0.1\textwidth]{figures/AIMfdm94.jpg} &
		\includegraphics[width=0.1\textwidth]{figures/AIMpredmapBMS94.jpg} \\
		\textbf{(a)} Original Image &
		\textbf{(b)} Ground Truth & 
		\textbf{(c)} BMS~\cite{BMS}
	\end{tabular}
	\footnotesize
	\begin{tabular}{|c|c c|}
		\hline
		& (b) & (c) \\ \hline
		SAUC~\cite{borjieval} & 0.8052 & 0.8234 \\	
		SKLD~\cite{SUN} & 0.2138 & 0.4064 \\
		Proposed SJSD & 0.8278  & 0.6417	\\ \hline		
	\end{tabular}
	\caption{Image from Toronto~\cite{AIM} database that shows the advantage of the proposed SJSD metric over the SAUC and SKLD metrics. In terms of SAUC and SKLD, the BMS model is incorrectly given a higher score than the ground-truth saliency map. The proposed SJSD metric gives a higher score to the ground-truth saliency map as expected.}
	\label{fig:SJSD_adv}
\end{figure}

The SJSD measure improves on the SKLD and SAUC measures by penalizing false-positives as shown in Figure~\ref{fig:SJSD_adv}. It can be seen that SAUC and SKLD both incorrectly give the saliency map for the BMS~\cite{BMS}  a higher score than the ground truth fixation density map which is rectified by the proposed SJSD metric.

\subsubsection{Shuffled Earth Mover's Distance (SEMD)}
\begin{figure}[t]
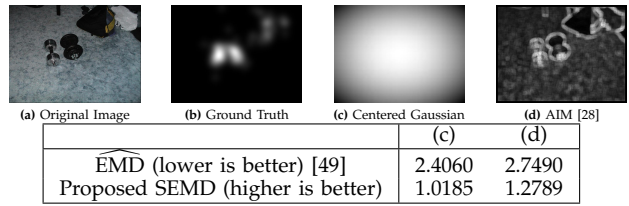

	\centering
	\tiny
	\begin{tabular}{cccc}
		\includegraphics[width=0.1\textwidth]{figures/AIMSAUC_failcase.jpg} &
		\includegraphics[width=0.1\textwidth]{figures/AIMfdm3.jpg} &
		\includegraphics[width=0.1\textwidth]{figures/AIMCenter.jpg} &
		\includegraphics[width=0.1\textwidth]{figures/AIMpredmapAIM3.jpg} \\
		\textbf{(a)} Original Image &
		\textbf{(b)} Ground Truth & 
		\textbf{(c)} Centered Gaussian &
		\textbf{(d)} AIM~\cite{AIM} 		
	\end{tabular}
	\footnotesize
	\begin{tabular}{|c|c c|}
		\hline
		& (c) & (d) \\ 
		\hline
		$\widehat{\textnormal{EMD}}$ (lower is better)~\cite{Judd_2012} & 2.4060 & 2.7490 \\ 
		Proposed SEMD (higher is better) & 1.0185 & 1.2789 \\ \hline
	\end{tabular}
	
	\caption{Center bias problem in the $\widehat{\textnormal{EMD}}$~\cite{Judd_2012} metric that is rectified by the proposed SEMD metric.}
	\label{fig:CBinEMD}
\end{figure}
As seen in Figure~\ref{fig:CBinEMD}, the disadvantage of using the  $\widehat{\textnormal{EMD}}$ metric defined in (\ref{eq:EMD}) to compare the histograms of the ground-truth and predicted saliency maps is that it is center-biased as it incorrectly assigns the centered Gaussian blob a higher score than the VA model. To rectify this issue, we propose a shuffled version of the $\widehat{\textnormal{EMD}}$ metric that is defined as follows:   
\begin{equation}
\textnormal{SEMD} = \widehat{\textnormal{EMD}}(H,\hat{H})
\label{Shuffled EMD}
\end{equation}
where $H$ and $\hat{H}$ are the same as in (\ref{eq:hist}).  Contrary to the original $\widehat{\textnormal{EMD}}$ of (\ref{eq:EMD}) (lower score means better), a higher SEMD score indicates better performance. The shuffled EMD score correctly assigns a lower score to the centered Gaussian blob as compared to the VA model. We use a fast implementation of the $\widehat{\textnormal{EMD}}$ metric proposed in~\cite{FastEMD} (code provided at ~\url{http://www.cs.huji.ac.il/~ofirpele/FastEMD/code/}) that makes use of a saturated ground distance in computing $\widehat{\textnormal{EMD}}$. 

\section{Performance Evaluation}
\label{sec:PerfEvalwithDistortion}

In order to evaluate the effect of different types and levels of distortions on the performance of VA models, predicted saliency maps are obtained for images that have been subjected to different levels and varying types of distortions. These are then compared with corresponding eye-tracking data results for which we use the proposed shuffled SNSS, SSKLD, SJSD and SEMD metrics introduced in Section~\ref{sec:ProposedMetrics} along with the widely used SAUC metric.  
As the different models produce saliency maps of differing  resolutions (see Table~\ref{table:models}),
we resize the saliency map given by each model to the size of the original image before computing the performance metric score as in~\cite{borjieval}. Also, as blurring the saliency map improves scores over certain metrics, we blur the predicted saliency map over different blur levels and choose the optimal blur level that maximizes the score as in~\cite{borjieval},~\cite{BMS}. For comparing the models with each other, we choose this maximum value at the optimum blur level. 

\subsection{Databases}
\subsubsection{TUD Interactions Database}
\begin{figure}[t]
\centering
\includegraphics[width=0.4\textwidth]{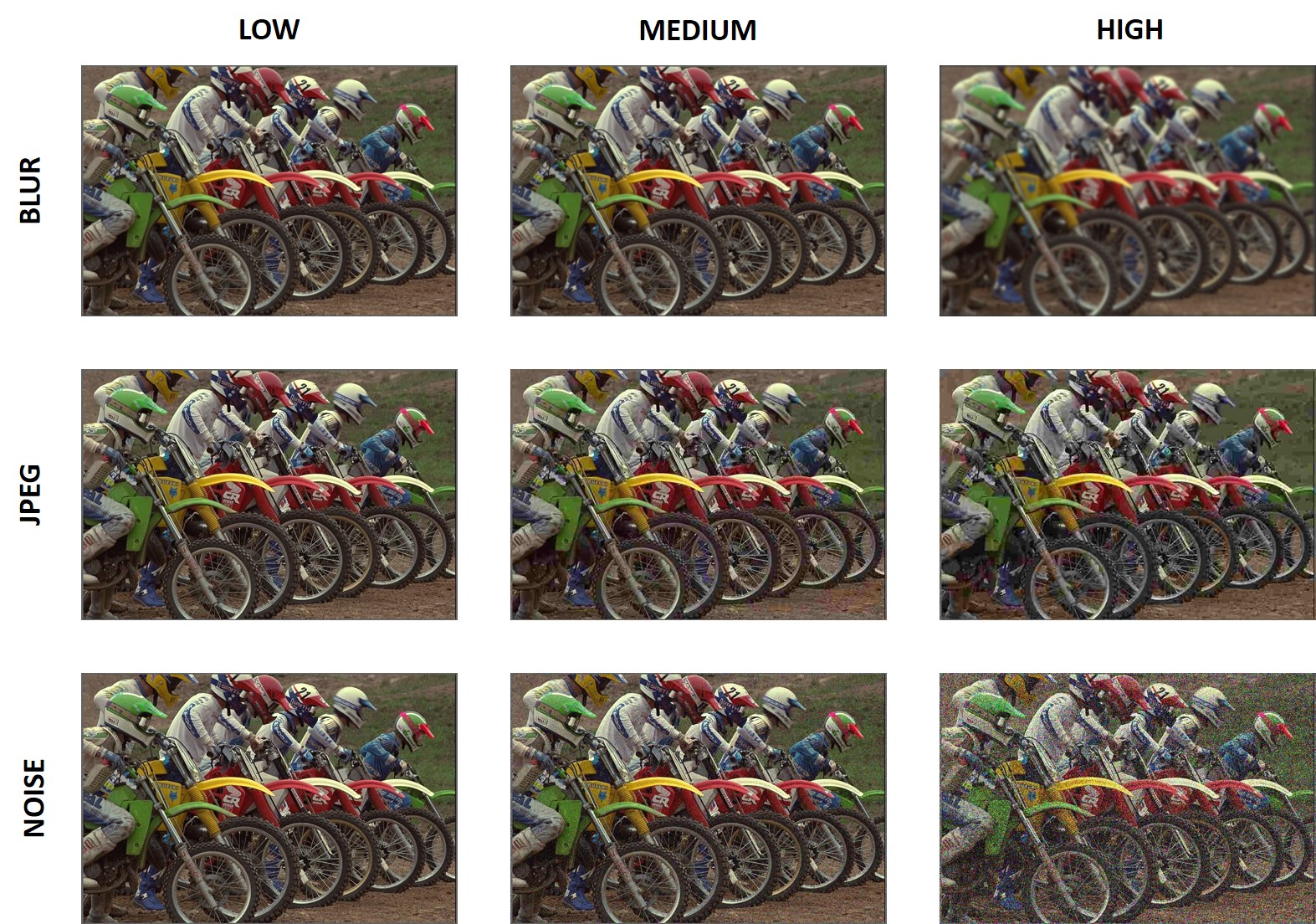}
\caption{Distorted versions of the Bikes image in the TUD database~\cite{TUD}. The level of distortion increases from left to right and the type of distortion changes from top to bottom, namely, Blur, JPEG and Gaussian Noise, respectively.}
\label{fig:TUD_ex}
\end{figure}

\begin{figure}[t]
\centering
\includegraphics[width=0.4\textwidth]{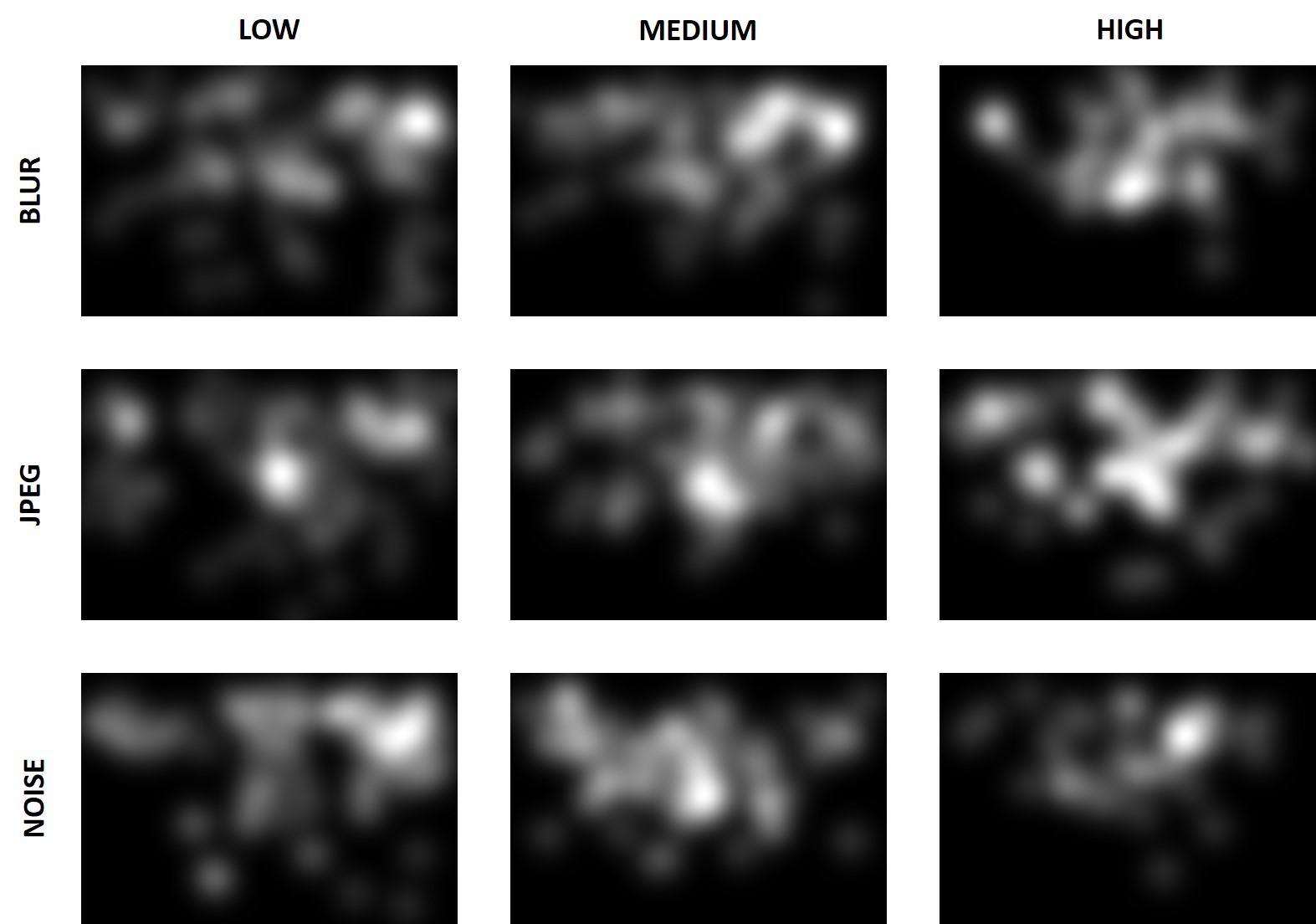}

\caption{Fixation density maps for distorted versions of the Bikes image in the TUD database~\cite{TUD}. The level of distortion increases from left to right and the type of distortion changes from top to bottom.}
\label{fig:TUD_Bikes_FDM}
\end{figure}

The TU Delft Interactions Database~\cite{TUDInteractions} is used as the ground-truth eye-tracking data for evaluating the performance over distorted images. This database includes  human eye movements recorded for 14 subjects while looking at 54 distorted stimuli. The different distortion types used are Gaussian blur, white noise and JPEG compression, with each type having three different levels of distortion (high, medium and low). The database provides MOS scores from the quality assessment task in addition to the saliency maps obtained from the recorded fixation points. Figure~\ref{fig:TUD_ex} shows the distorted versions (different types and different levels of distortion) for the Bikes image, and Figure~\ref{fig:TUD_Bikes_FDM} shows the corresponding fixation density maps for the images. The fixation density maps are obtained by placing a 2D Gaussian on the fixation point locations and normalizing the resulting map.

\subsubsection{Judd Low Resolution Database}
The Judd Low Resolution database~\cite{JuddLowRes} contains eye tracking data for original versions of  168 natural images and 25 pink noise images in addition to downsampled versions of the images varying from 4 to 512 in terms of image height. The images are shown to 64 observers after scaling them to a size of $860 \times 1024$. 
The natural images are divided into three categories, easy, medium and hard, depending on image complexity. The categorization is done using a subjective criteria in which first, each image is assigned a score that corresponds to the lowest resolution at which the image is understood by a subject. The images are then ranked in the ascending order and binned into three equal sized categories. The images with the lowest scores make up the easy category and those with the highest scores form the hard category with the remaining images making up the medium category. Based on the eye tracking data, it is observed that for the easy images, fixations stay consistent even for extremely low resolutions and are more centered. On the other hand, the hard images have more variations in eye fixations which are less centered for higher resolutions because there are more details on which subjects fixate.
As the input images are subsampled at different levels, the distortion that each image is affected with, in this case, is equivalent to blur. The lower the resolution of the image, the higher is the level of blur.

\begin{figure}[t!]
\centering
\includegraphics[width=0.45\textwidth]{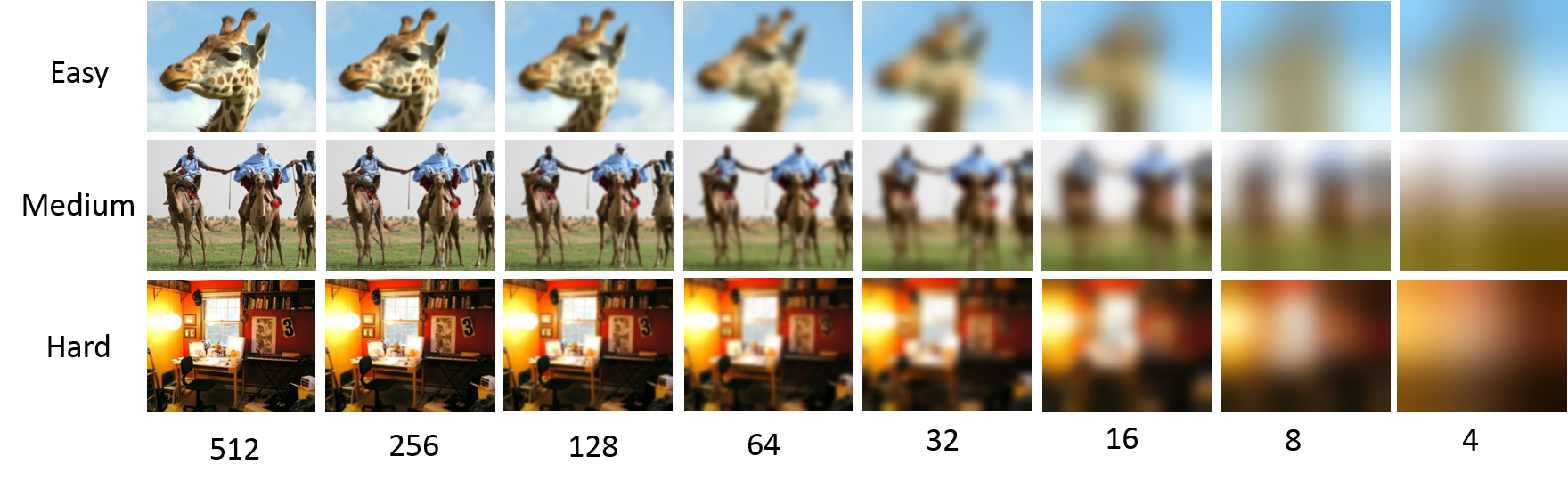}
\caption{Sample images covering all resolutions from the Judd Low Resolution Database~\cite{JuddLowRes} for each of the three categories: Easy, Medium and Hard.}
\label{fig:JuddLowRes}
\end{figure}

\begin{figure}[t!]
\centering
\includegraphics[width=0.45\textwidth]{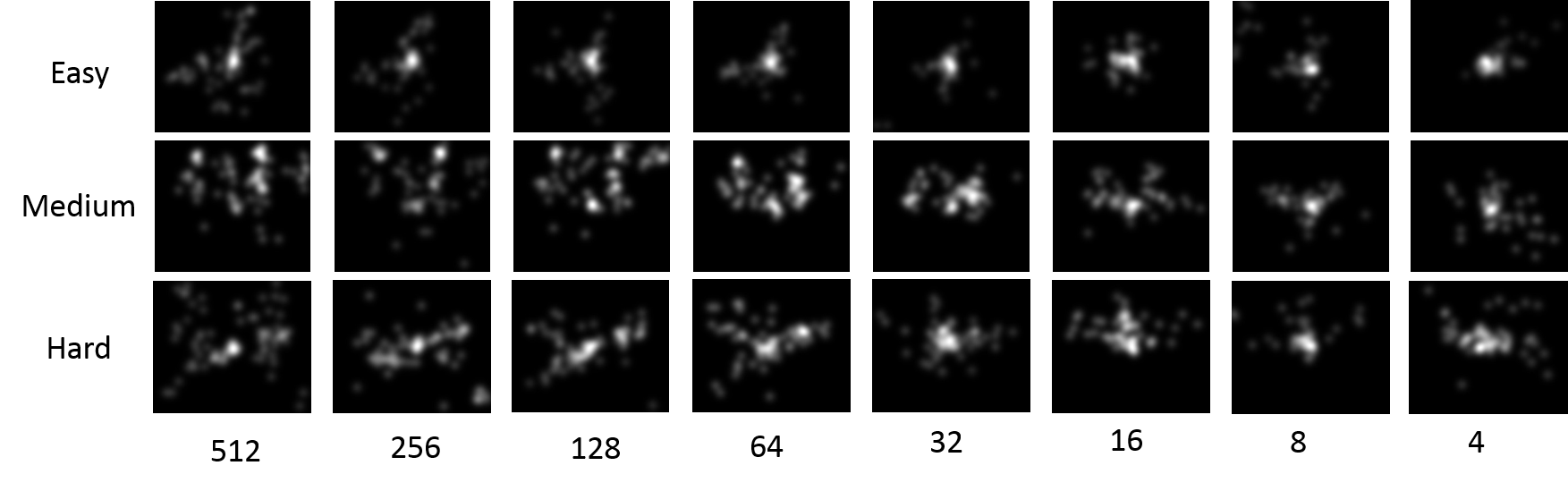}

\caption{Fixation density maps for images shown in Figure~\ref{fig:JuddLowRes} covering all resolutions from the Judd Low Resolution Database~\cite{JuddLowRes} for each of the three categories: Easy, Medium and Hard.}
\label{fig:JuddLowResFDM}
\end{figure}

Figure~\ref{fig:JuddLowRes} shows a sample image from each of the three categories for the different resolutions and Figure~\ref{fig:JuddLowResFDM} shows the corresponding fixation density maps that are obtained by convolving a 2D Gaussian kernel over all the fixations for the images and then normalizing the resulting maps. In our analysis, we only consider the natural images (1344 out of 1544) in the dataset and ignore the colored noise images because the considered saliency models  give a meaningful output  for images that have some inherent structure and which are not random in nature.

\subsection{Results}
\subsubsection{Performance over non-distorted images} 
\label{subsec:performance non-distorted}
We first validate the proposed metrics by using them to evaluate the VA models listed in Table~\ref{table:models} on the widely used Toronto~\cite{AIM} and MIT~\cite{MIT} databases which consist of non-distorted images. Figure~\ref{fig:TorontoMIT} shows the performance for the proposed metrics as well as the SAUC metric~\cite{borjieval}. To validate the consistency of the metrics over the two datasets, we adopt as in~\cite{riche2013saliency} the Kendall's coefficient of concordance (KCC) to compare the models' rankings provided by each metric over the two datasets. Table~\ref{tab:PerfConsistencyTorontoMIT} shows the obtained results. From Table~\ref{tab:PerfConsistencyTorontoMIT}, it can be seen that the proposed metrics provide consistent ratings in terms of models' rankings over the two datasets with the proposed SNSS metric showing the best consistency.   
\begin{figure*}
	\scalefont{0.6}
	\setlength{\figureheight}{0.07\textwidth}
	\setlength{\figurewidth}{0.3\textwidth}
	\centering
	\subfloat[SAUC]{
%
\begin{tikzpicture}

\begin{axis}[%
width=\figurewidth,
height=0.94843\figureheight,
at={(0\figurewidth,0\figureheight)},
scale only axis,
clip=false,
separate axis lines,
every outer x axis line/.append style={black},
every x tick label/.append style={font=\color{black}},
xmin=0,
xmax=25,
xtick={1,2,3,4,5,6,7,8,9,10,11,12,13,14,15,16,17,18,19,20,21,22,23,24},
xticklabels={\empty},
every outer y axis line/.append style={black},
every y tick label/.append style={font=\color{black}},
ymin=0.5,
ymax=0.75,
legend to name=badlegend,
legend style={legend pos=north east,legend cell align=center,align=center,font=\footnotesize,legend columns=1}
]
\addplot [color=blue,solid,line width=1.0pt,mark size=2.3pt,mark=asterisk,mark options={solid}]
  table[row sep=crcr]{%
1	0.684700758364161\\
2	0.710413291186334\\
3	0.70945308559947\\
4	0.542524629227758\\
5	0.686964160383078\\
6	0.589416113910797\\
7	0.581020343827812\\
8	0.595897106347302\\
9	0.619478251753461\\
10	0.620923449396726\\
11	0.681338666422198\\
12	0.682513869155317\\
13	0.609128391723198\\
14	0.599447136012995\\
15	0.620834111907826\\
16	0.575941630325078\\
17	0.700379484288894\\
18	0.702536815568237\\
19	0.696298381850869\\
20	0.6683316249365\\
21	0.670208564243203\\
22	0.676511171192454\\
23	0.656902624997977\\
24	0.678517024082637\\
};
\addlegendentry{Toronto};

\addplot [color=red,solid,line width=1.0pt,mark size=2.3pt,mark=asterisk,mark options={solid}]
  table[row sep=crcr]{%
1	0.692664991558388\\
2	0.703241369257717\\
3	0.706990723390186\\
4	0.587635666196183\\
5	0.686943822400017\\
6	0.621975205480359\\
7	0.587143466564781\\
8	0.628045559838633\\
9	0.646251858416437\\
10	0.669128167631475\\
11	0.686512592614881\\
12	0.677771547509849\\
13	0.638778784587306\\
14	0.607899132807593\\
15	0.649953033453578\\
16	0.606849342901679\\
17	0.675906396796027\\
18	0.671641978319247\\
19	0.683258704509767\\
20	0.598200403191638\\
21	0.595000039937643\\
22	0.668436044408819\\
23	0.579754768160801\\
24	0.6870645762818\\
};
\addlegendentry{MIT};

\node[left, align=right, inner sep=0mm, rotate=90, text=black]
at (rel axis cs:0.0357941834451902,-0.0424528301886792,0) {AIM};
\node[left, align=right, inner sep=0mm, rotate=90, text=black]
at (rel axis cs:0.0760626398210291,-0.0424528301886792,0) {AWS};
\node[left, align=right, inner sep=0mm, rotate=90, text=black]
at (rel axis cs:0.116331096196868,-0.0424528301886792,0) {BMS};
\node[left, align=right, inner sep=0mm, rotate=90, text=black]
at (rel axis cs:0.156599552572707,-0.0424528301886792,0) {CovSal};
\node[left, align=right, inner sep=0mm, rotate=90, text=black]
at (rel axis cs:0.196868008948546,-0.0424528301886792,0) {Context};
\node[left, align=right, inner sep=0mm, rotate=90, text=black]
at (rel axis cs:0.237136465324385,-0.0424528301886792,0) {SCIA};
\node[left, align=right, inner sep=0mm, rotate=90, text=black]
at (rel axis cs:0.275167785234899,-0.0424528301886792,0) {FTS};
\node[left, align=right, inner sep=0mm, rotate=90, text=black]
at (rel axis cs:0.315436241610738,-0.0424528301886792,0) {GAFFE};
\node[left, align=right, inner sep=0mm, rotate=90, text=black]
at (rel axis cs:0.355704697986577,-0.0424528301886792,0) {GBVS};
\node[left, align=right, inner sep=0mm, rotate=90, text=black]
at (rel axis cs:0.395973154362416,-0.0424528301886792,0) {GBVSNoCB};
\node[left, align=right, inner sep=0mm, rotate=90, text=black]
at (rel axis cs:0.436241610738255,-0.0424528301886792,0) {GR};
\node[left, align=right, inner sep=0mm, rotate=90, text=black]
at (rel axis cs:0.476510067114094,-0.0424528301886792,0) {HouNIPS};
\node[left, align=right, inner sep=0mm, rotate=90, text=black]
at (rel axis cs:0.514541387024609,-0.0424528301886792,0) {ITTI};
\node[left, align=right, inner sep=0mm, rotate=90, text=black]
at (rel axis cs:0.554809843400447,-0.0424528301886792,0) {ITTI2};
\node[left, align=right, inner sep=0mm, rotate=90, text=black]
at (rel axis cs:0.595078299776286,-0.0424528301886792,0) {Judd};
\node[left, align=right, inner sep=0mm, rotate=90, text=black]
at (rel axis cs:0.635346756152125,-0.0424528301886792,0) {RandomCS};
\node[left, align=right, inner sep=0mm, rotate=90, text=black]
at (rel axis cs:0.675615212527964,-0.0424528301886792,0) {SDSRL};
\node[left, align=right, inner sep=0mm, rotate=90, text=black]
at (rel axis cs:0.715883668903803,-0.0424528301886792,0) {SDSRG};
\node[left, align=right, inner sep=0mm, rotate=90, text=black]
at (rel axis cs:0.753914988814318,-0.0424528301886792,0) {SigSal};
\node[left, align=right, inner sep=0mm, rotate=90, text=black]
at (rel axis cs:0.794183445190157,-0.0424528301886792,0) {SIMFine};
\node[left, align=right, inner sep=0mm, rotate=90, text=black]
at (rel axis cs:0.834451901565995,-0.0424528301886792,0) {SIMCoarse};
\node[left, align=right, inner sep=0mm, rotate=90, text=black]
at (rel axis cs:0.874720357941834,-0.0424528301886792,0) {SpectRes};
\node[left, align=right, inner sep=0mm, rotate=90, text=black]
at (rel axis cs:0.914988814317673,-0.0424528301886792,0) {SUN};
\node[left, align=right, inner sep=0mm, rotate=90, text=black]
at (rel axis cs:0.955257270693512,-0.0424528301886792,0) {Torralba};
\end{axis}
\end{tikzpicture}%
\ref{badlegend}}	
    \subfloat[SNSS]{
%
\begin{tikzpicture}

\begin{axis}[%
width=\figurewidth,
height=0.94843\figureheight,
at={(0\figurewidth,0\figureheight)},
scale only axis,
clip=false,
separate axis lines,
every outer x axis line/.append style={black},
every x tick label/.append style={font=\color{black}},
xmin=0,
xmax=25,
xtick={1,2,3,4,5,6,7,8,9,10,11,12,13,14,15,16,17,18,19,20,21,22,23,24},
xticklabels={\empty},
every outer y axis line/.append style={black},
every y tick label/.append style={font=\color{black}},
ymin=0.2,
ymax=1.2
]
\addplot [color=blue,solid,line width=1.0pt,mark size=2.3pt,mark=asterisk,mark options={solid},forget plot]
  table[row sep=crcr]{%
1	0.63064498377852\\
2	1.00063035226107\\
3	1.16927160240646\\
4	0.290149411675749\\
5	0.890957144306964\\
6	0.640236881298359\\
7	0.285037886197931\\
8	0.297064466607153\\
9	0.478674928017182\\
10	0.45441170335791\\
11	0.667965169300779\\
12	0.934785632817778\\
13	0.416420752291807\\
14	0.568384233532141\\
15	0.450172057346951\\
16	0.364921537615805\\
17	0.810333366132713\\
18	0.947501515573448\\
19	0.983904653688433\\
20	0.819928506382878\\
21	0.835751713936985\\
22	0.774265856145854\\
23	0.595978779902279\\
24	0.628808882969029\\
};
\addplot [color=red,solid,line width=1.0pt,mark size=2.3pt,mark=asterisk,mark options={solid},forget plot]
  table[row sep=crcr]{%
1	0.597655475429455\\
2	0.954613595558\\
3	0.960047199970568\\
4	0.582397476600191\\
5	0.802517900380763\\
6	0.881420348271831\\
7	0.3193264529352\\
8	0.411470402835313\\
9	0.602566325538551\\
10	0.708107317987034\\
11	0.578919029870847\\
12	0.872575760990885\\
13	0.541271504779525\\
14	0.465597585551808\\
15	0.421454311741958\\
16	0.424596054116244\\
17	0.615685132557649\\
18	0.751279727417906\\
19	0.800437015156939\\
20	0.788824754900382\\
21	0.813187167353347\\
22	0.738965786821232\\
23	0.463890388880294\\
24	0.59820806828109\\
};
\node[left, align=right, inner sep=0mm, rotate=90, text=black]
at (rel axis cs:0.0357941834451902,-0.0471698113207547,0) {AIM};
\node[left, align=right, inner sep=0mm, rotate=90, text=black]
at (rel axis cs:0.0760626398210291,-0.0471698113207547,0) {AWS};
\node[left, align=right, inner sep=0mm, rotate=90, text=black]
at (rel axis cs:0.116331096196868,-0.0471698113207547,0) {BMS};
\node[left, align=right, inner sep=0mm, rotate=90, text=black]
at (rel axis cs:0.156599552572707,-0.0471698113207547,0) {CovSal};
\node[left, align=right, inner sep=0mm, rotate=90, text=black]
at (rel axis cs:0.196868008948546,-0.0471698113207547,0) {Context};
\node[left, align=right, inner sep=0mm, rotate=90, text=black]
at (rel axis cs:0.237136465324385,-0.0471698113207547,0) {SCIA};
\node[left, align=right, inner sep=0mm, rotate=90, text=black]
at (rel axis cs:0.275167785234899,-0.0471698113207547,0) {FTS};
\node[left, align=right, inner sep=0mm, rotate=90, text=black]
at (rel axis cs:0.315436241610738,-0.0471698113207547,0) {GAFFE};
\node[left, align=right, inner sep=0mm, rotate=90, text=black]
at (rel axis cs:0.355704697986577,-0.0471698113207547,0) {GBVS};
\node[left, align=right, inner sep=0mm, rotate=90, text=black]
at (rel axis cs:0.395973154362416,-0.0471698113207547,0) {GBVSNoCB};
\node[left, align=right, inner sep=0mm, rotate=90, text=black]
at (rel axis cs:0.436241610738255,-0.0471698113207547,0) {GR};
\node[left, align=right, inner sep=0mm, rotate=90, text=black]
at (rel axis cs:0.476510067114094,-0.0471698113207547,0) {HouNIPS};
\node[left, align=right, inner sep=0mm, rotate=90, text=black]
at (rel axis cs:0.514541387024609,-0.0471698113207547,0) {ITTI};
\node[left, align=right, inner sep=0mm, rotate=90, text=black]
at (rel axis cs:0.554809843400447,-0.0471698113207547,0) {ITTI2};
\node[left, align=right, inner sep=0mm, rotate=90, text=black]
at (rel axis cs:0.595078299776286,-0.0471698113207547,0) {Judd};
\node[left, align=right, inner sep=0mm, rotate=90, text=black]
at (rel axis cs:0.635346756152125,-0.0471698113207547,0) {RandomCS};
\node[left, align=right, inner sep=0mm, rotate=90, text=black]
at (rel axis cs:0.675615212527964,-0.0471698113207547,0) {SDSRL};
\node[left, align=right, inner sep=0mm, rotate=90, text=black]
at (rel axis cs:0.715883668903803,-0.0471698113207547,0) {SDSRG};
\node[left, align=right, inner sep=0mm, rotate=90, text=black]
at (rel axis cs:0.753914988814318,-0.0471698113207547,0) {SigSal};
\node[left, align=right, inner sep=0mm, rotate=90, text=black]
at (rel axis cs:0.794183445190157,-0.0471698113207547,0) {SIMFine};
\node[left, align=right, inner sep=0mm, rotate=90, text=black]
at (rel axis cs:0.834451901565995,-0.0471698113207547,0) {SIMCoarse};
\node[left, align=right, inner sep=0mm, rotate=90, text=black]
at (rel axis cs:0.874720357941834,-0.0471698113207547,0) {SpectRes};
\node[left, align=right, inner sep=0mm, rotate=90, text=black]
at (rel axis cs:0.914988814317673,-0.0471698113207547,0) {SUN};
\node[left, align=right, inner sep=0mm, rotate=90, text=black]
at (rel axis cs:0.955257270693512,-0.0471698113207547,0) {Torralba};
\end{axis}
\end{tikzpicture}%
}
    \qquad
    \subfloat[SEMD]{
%
\begin{tikzpicture}

\begin{axis}[%
width=\figurewidth,
height=0.94843\figureheight,
at={(0\figurewidth,0\figureheight)},
scale only axis,
clip=false,
separate axis lines,
every outer x axis line/.append style={black},
every x tick label/.append style={font=\color{black}},
xmin=0,
xmax=25,
xtick={1,2,3,4,5,6,7,8,9,10,11,12,13,14,15,16,17,18,19,20,21,22,23,24},
xticklabels={\empty},
every outer y axis line/.append style={black},
every y tick label/.append style={font=\color{black}},
ymin=1.5,
ymax=2
]
\addplot [color=blue,solid,line width=1.0pt,mark size=2.3pt,mark=asterisk,mark options={solid},forget plot]
  table[row sep=crcr]{%
1	1.64338509031908\\
2	1.76638519931995\\
3	1.74958195589975\\
4	1.72510244179004\\
5	1.75725045309066\\
6	1.73741507908367\\
7	1.56769090194788\\
8	1.60797584540712\\
9	1.66042314773871\\
10	1.63796075101129\\
11	1.68834332499272\\
12	1.69965388862652\\
13	1.62258653013711\\
14	1.52821869544828\\
15	1.53155009522497\\
16	1.64642809801448\\
17	1.76523474022663\\
18	1.81148737205295\\
19	1.69506691320106\\
20	1.69282306914694\\
21	1.7105287823076\\
22	1.69388934135172\\
23	1.64913333323714\\
24	1.6549970650543\\
};
\addplot [color=red,solid,line width=1.0pt,mark size=2.3pt,mark=asterisk,mark options={solid},forget plot]
  table[row sep=crcr]{%
1	1.83791930319304\\
2	1.97023995664449\\
3	1.94384464749594\\
4	1.95843357961907\\
5	1.9612715015234\\
6	1.97169690776575\\
7	1.78198351257388\\
8	1.83104170273886\\
9	1.86689564107895\\
10	1.89845301857858\\
11	1.86882393571981\\
12	1.92966106443518\\
13	1.86587237187284\\
14	1.68880856644599\\
15	1.75227088713458\\
16	1.87600649044935\\
17	1.90938088175371\\
18	1.95597230531432\\
19	1.88799176606812\\
20	1.7089499870117\\
21	1.72426899258405\\
22	1.91291619952155\\
23	1.56938152759454\\
24	1.85343246855385\\
};
\node[left, align=right, inner sep=0mm, rotate=90, text=black]
at (rel axis cs:0.0357941834451902,-0.0424528301886792,0) {AIM};
\node[left, align=right, inner sep=0mm, rotate=90, text=black]
at (rel axis cs:0.0760626398210291,-0.0424528301886792,0) {AWS};
\node[left, align=right, inner sep=0mm, rotate=90, text=black]
at (rel axis cs:0.116331096196868,-0.0424528301886792,0) {BMS};
\node[left, align=right, inner sep=0mm, rotate=90, text=black]
at (rel axis cs:0.156599552572707,-0.0424528301886792,0) {CovSal};
\node[left, align=right, inner sep=0mm, rotate=90, text=black]
at (rel axis cs:0.196868008948546,-0.0424528301886792,0) {Context};
\node[left, align=right, inner sep=0mm, rotate=90, text=black]
at (rel axis cs:0.237136465324385,-0.0424528301886792,0) {SCIA};
\node[left, align=right, inner sep=0mm, rotate=90, text=black]
at (rel axis cs:0.275167785234899,-0.0424528301886792,0) {FTS};
\node[left, align=right, inner sep=0mm, rotate=90, text=black]
at (rel axis cs:0.315436241610738,-0.0424528301886792,0) {GAFFE};
\node[left, align=right, inner sep=0mm, rotate=90, text=black]
at (rel axis cs:0.355704697986577,-0.0424528301886792,0) {GBVS};
\node[left, align=right, inner sep=0mm, rotate=90, text=black]
at (rel axis cs:0.395973154362416,-0.0424528301886792,0) {GBVSNoCB};
\node[left, align=right, inner sep=0mm, rotate=90, text=black]
at (rel axis cs:0.436241610738255,-0.0424528301886792,0) {GR};
\node[left, align=right, inner sep=0mm, rotate=90, text=black]
at (rel axis cs:0.476510067114094,-0.0424528301886792,0) {HouNIPS};
\node[left, align=right, inner sep=0mm, rotate=90, text=black]
at (rel axis cs:0.514541387024609,-0.0424528301886792,0) {ITTI};
\node[left, align=right, inner sep=0mm, rotate=90, text=black]
at (rel axis cs:0.554809843400447,-0.0424528301886792,0) {ITTI2};
\node[left, align=right, inner sep=0mm, rotate=90, text=black]
at (rel axis cs:0.595078299776286,-0.0424528301886792,0) {Judd};
\node[left, align=right, inner sep=0mm, rotate=90, text=black]
at (rel axis cs:0.635346756152125,-0.0424528301886792,0) {RandomCS};
\node[left, align=right, inner sep=0mm, rotate=90, text=black]
at (rel axis cs:0.675615212527964,-0.0424528301886792,0) {SDSRL};
\node[left, align=right, inner sep=0mm, rotate=90, text=black]
at (rel axis cs:0.715883668903803,-0.0424528301886792,0) {SDSRG};
\node[left, align=right, inner sep=0mm, rotate=90, text=black]
at (rel axis cs:0.753914988814318,-0.0424528301886792,0) {SigSal};
\node[left, align=right, inner sep=0mm, rotate=90, text=black]
at (rel axis cs:0.794183445190157,-0.0424528301886792,0) {SIMFine};
\node[left, align=right, inner sep=0mm, rotate=90, text=black]
at (rel axis cs:0.834451901565995,-0.0424528301886792,0) {SIMCoarse};
\node[left, align=right, inner sep=0mm, rotate=90, text=black]
at (rel axis cs:0.874720357941834,-0.0424528301886792,0) {SpectRes};
\node[left, align=right, inner sep=0mm, rotate=90, text=black]
at (rel axis cs:0.914988814317673,-0.0424528301886792,0) {SUN};
\node[left, align=right, inner sep=0mm, rotate=90, text=black]
at (rel axis cs:0.955257270693512,-0.0424528301886792,0) {Torralba};
\end{axis}
\end{tikzpicture}%
}
    \subfloat[SJSD]{
%
\begin{tikzpicture}

\begin{axis}[%
width=\figurewidth,
height=0.94843\figureheight,
at={(0\figurewidth,0\figureheight)},
scale only axis,
clip=false,
separate axis lines,
every outer x axis line/.append style={black},
every x tick label/.append style={font=\color{black}},
xmin=0,
xmax=25,
xtick={1,2,3,4,5,6,7,8,9,10,11,12,13,14,15,16,17,18,19,20,21,22,23,24},
xticklabels={\empty},
every outer y axis line/.append style={black},
every y tick label/.append style={font=\color{black}},
ymin=0.7,
ymax=0.95
]
\addplot [color=blue,solid,line width=1.0pt,mark size=2.3pt,mark=asterisk,mark options={solid},forget plot]
  table[row sep=crcr]{%
1	0.778406230561174\\
2	0.819165804374568\\
3	0.809645453301766\\
4	0.844442322834731\\
5	0.819683156708605\\
6	0.840024058031671\\
7	0.766687066634658\\
8	0.776346409262408\\
9	0.799724992830541\\
10	0.789097606171234\\
11	0.794887705764211\\
12	0.800831612271683\\
13	0.799317682500237\\
14	0.751624210827311\\
15	0.762890909191621\\
16	0.809515384423517\\
17	0.818296925058261\\
18	0.825769462149824\\
19	0.792910698494499\\
20	0.805555873904508\\
21	0.809703801293841\\
22	0.803133621042976\\
23	0.782315962256032\\
24	0.783420508828169\\
};
\addplot [color=red,solid,line width=1.0pt,mark size=2.3pt,mark=asterisk,mark options={solid},forget plot]
  table[row sep=crcr]{%
1	0.825894900137568\\
2	0.864675197397662\\
3	0.853920500839295\\
4	0.887920480232205\\
5	0.864050420649785\\
6	0.883693192726452\\
7	0.814837935514149\\
8	0.831121310370127\\
9	0.845738792951109\\
10	0.848101068825734\\
11	0.838474726874366\\
12	0.853915086326575\\
13	0.850988026780876\\
14	0.791527406359512\\
15	0.821253212843023\\
16	0.855689527602304\\
17	0.847887600127537\\
18	0.856860255119402\\
19	0.841525324369176\\
20	0.755726206509104\\
21	0.759802725345244\\
22	0.850682071566597\\
23	0.734045894411865\\
24	0.834341827918045\\
};
\node[left, align=right, inner sep=0mm, rotate=90, text=black]
at (rel axis cs:0.0357941834451902,-0.0424528301886792,0) {AIM};
\node[left, align=right, inner sep=0mm, rotate=90, text=black]
at (rel axis cs:0.0760626398210291,-0.0424528301886792,0) {AWS};
\node[left, align=right, inner sep=0mm, rotate=90, text=black]
at (rel axis cs:0.116331096196868,-0.0424528301886792,0) {BMS};
\node[left, align=right, inner sep=0mm, rotate=90, text=black]
at (rel axis cs:0.156599552572707,-0.0424528301886792,0) {CovSal};
\node[left, align=right, inner sep=0mm, rotate=90, text=black]
at (rel axis cs:0.196868008948546,-0.0424528301886792,0) {Context};
\node[left, align=right, inner sep=0mm, rotate=90, text=black]
at (rel axis cs:0.237136465324385,-0.0424528301886792,0) {SCIA};
\node[left, align=right, inner sep=0mm, rotate=90, text=black]
at (rel axis cs:0.275167785234899,-0.0424528301886792,0) {FTS};
\node[left, align=right, inner sep=0mm, rotate=90, text=black]
at (rel axis cs:0.315436241610738,-0.0424528301886792,0) {GAFFE};
\node[left, align=right, inner sep=0mm, rotate=90, text=black]
at (rel axis cs:0.355704697986577,-0.0424528301886792,0) {GBVS};
\node[left, align=right, inner sep=0mm, rotate=90, text=black]
at (rel axis cs:0.395973154362416,-0.0424528301886792,0) {GBVSNoCB};
\node[left, align=right, inner sep=0mm, rotate=90, text=black]
at (rel axis cs:0.436241610738255,-0.0424528301886792,0) {GR};
\node[left, align=right, inner sep=0mm, rotate=90, text=black]
at (rel axis cs:0.476510067114094,-0.0424528301886792,0) {HouNIPS};
\node[left, align=right, inner sep=0mm, rotate=90, text=black]
at (rel axis cs:0.514541387024609,-0.0424528301886792,0) {ITTI};
\node[left, align=right, inner sep=0mm, rotate=90, text=black]
at (rel axis cs:0.554809843400447,-0.0424528301886792,0) {ITTI2};
\node[left, align=right, inner sep=0mm, rotate=90, text=black]
at (rel axis cs:0.595078299776286,-0.0424528301886792,0) {Judd};
\node[left, align=right, inner sep=0mm, rotate=90, text=black]
at (rel axis cs:0.635346756152125,-0.0424528301886792,0) {RandomCS};
\node[left, align=right, inner sep=0mm, rotate=90, text=black]
at (rel axis cs:0.675615212527964,-0.0424528301886792,0) {SDSRL};
\node[left, align=right, inner sep=0mm, rotate=90, text=black]
at (rel axis cs:0.715883668903803,-0.0424528301886792,0) {SDSRG};
\node[left, align=right, inner sep=0mm, rotate=90, text=black]
at (rel axis cs:0.753914988814318,-0.0424528301886792,0) {SigSal};
\node[left, align=right, inner sep=0mm, rotate=90, text=black]
at (rel axis cs:0.794183445190157,-0.0424528301886792,0) {SIMFine};
\node[left, align=right, inner sep=0mm, rotate=90, text=black]
at (rel axis cs:0.834451901565995,-0.0424528301886792,0) {SIMCoarse};
\node[left, align=right, inner sep=0mm, rotate=90, text=black]
at (rel axis cs:0.874720357941834,-0.0424528301886792,0) {SpectRes};
\node[left, align=right, inner sep=0mm, rotate=90, text=black]
at (rel axis cs:0.914988814317673,-0.0424528301886792,0) {SUN};
\node[left, align=right, inner sep=0mm, rotate=90, text=black]
at (rel axis cs:0.955257270693512,-0.0424528301886792,0) {Torralba};
\end{axis}
\end{tikzpicture}%
}
    \subfloat[SSKLD]{
%
\begin{tikzpicture}

\begin{axis}[%
width=\figurewidth,
height=0.94843\figureheight,
at={(0\figurewidth,0\figureheight)},
scale only axis,
clip=false,
separate axis lines,
every outer x axis line/.append style={black},
every x tick label/.append style={font=\color{black}},
xmin=0,
xmax=25,
xtick={1,2,3,4,5,6,7,8,9,10,11,12,13,14,15,16,17,18,19,20,21,22,23,24},
xticklabels={\empty},
every outer y axis line/.append style={black},
every y tick label/.append style={font=\color{black}},
ymin=0.05,
ymax=0.4
]
\addplot [color=blue,solid,line width=1.0pt,mark size=2.3pt,mark=asterisk,mark options={solid},forget plot]
  table[row sep=crcr]{%
1	0.236334475257729\\
2	0.257345422213379\\
3	0.299147627702234\\
4	0.0622166154621486\\
5	0.239799904228963\\
6	0.213114140698118\\
7	0.127351866644052\\
8	0.194330136681152\\
9	0.182918907471967\\
10	0.189289088537462\\
11	0.279499110954524\\
12	0.314096438593862\\
13	0.187316842490308\\
14	0.283295563823402\\
15	0.199886225793763\\
16	0.136122821991764\\
17	0.219849422888071\\
18	0.3723224301281\\
19	0.289969698957268\\
20	0.277370586165601\\
21	0.255759468406728\\
22	0.268606313591695\\
23	0.213638384398674\\
24	0.22514372885593\\
};
\addplot [color=red,solid,line width=1.0pt,mark size=2.3pt,mark=asterisk,mark options={solid},forget plot]
  table[row sep=crcr]{%
1	0.254323189701408\\
2	0.197693054518464\\
3	0.255116579802249\\
4	0.0983268805950581\\
5	0.200525825021769\\
6	0.282117688426105\\
7	0.125294116118717\\
8	0.20915250521491\\
9	0.186279498150329\\
10	0.201223122279886\\
11	0.241706408885032\\
12	0.210291084339924\\
13	0.179227152779548\\
14	0.190641331942432\\
15	0.206587582040678\\
16	0.149956695591529\\
17	0.227549773634559\\
18	0.256334031200312\\
19	0.216724237347234\\
20	0.214223172989377\\
21	0.206761853452449\\
22	0.200012186791594\\
23	0.22694372277609\\
24	0.242404051433646\\
};
\node[left, align=right, inner sep=0mm, rotate=90, text=black]
at (rel axis cs:0.0357941834451902,-0.0471698113207547,0) {AIM};
\node[left, align=right, inner sep=0mm, rotate=90, text=black]
at (rel axis cs:0.0760626398210291,-0.0471698113207547,0) {AWS};
\node[left, align=right, inner sep=0mm, rotate=90, text=black]
at (rel axis cs:0.116331096196868,-0.0471698113207547,0) {BMS};
\node[left, align=right, inner sep=0mm, rotate=90, text=black]
at (rel axis cs:0.156599552572707,-0.0471698113207547,0) {CovSal};
\node[left, align=right, inner sep=0mm, rotate=90, text=black]
at (rel axis cs:0.196868008948546,-0.0471698113207547,0) {Context};
\node[left, align=right, inner sep=0mm, rotate=90, text=black]
at (rel axis cs:0.237136465324385,-0.0471698113207547,0) {SCIA};
\node[left, align=right, inner sep=0mm, rotate=90, text=black]
at (rel axis cs:0.275167785234899,-0.0471698113207547,0) {FTS};
\node[left, align=right, inner sep=0mm, rotate=90, text=black]
at (rel axis cs:0.315436241610738,-0.0471698113207547,0) {GAFFE};
\node[left, align=right, inner sep=0mm, rotate=90, text=black]
at (rel axis cs:0.355704697986577,-0.0471698113207547,0) {GBVS};
\node[left, align=right, inner sep=0mm, rotate=90, text=black]
at (rel axis cs:0.395973154362416,-0.0471698113207547,0) {GBVSNoCB};
\node[left, align=right, inner sep=0mm, rotate=90, text=black]
at (rel axis cs:0.436241610738255,-0.0471698113207547,0) {GR};
\node[left, align=right, inner sep=0mm, rotate=90, text=black]
at (rel axis cs:0.476510067114094,-0.0471698113207547,0) {HouNIPS};
\node[left, align=right, inner sep=0mm, rotate=90, text=black]
at (rel axis cs:0.514541387024609,-0.0471698113207547,0) {ITTI};
\node[left, align=right, inner sep=0mm, rotate=90, text=black]
at (rel axis cs:0.554809843400447,-0.0471698113207547,0) {ITTI2};
\node[left, align=right, inner sep=0mm, rotate=90, text=black]
at (rel axis cs:0.595078299776286,-0.0471698113207547,0) {Judd};
\node[left, align=right, inner sep=0mm, rotate=90, text=black]
at (rel axis cs:0.635346756152125,-0.0471698113207547,0) {RandomCS};
\node[left, align=right, inner sep=0mm, rotate=90, text=black]
at (rel axis cs:0.675615212527964,-0.0471698113207547,0) {SDSRL};
\node[left, align=right, inner sep=0mm, rotate=90, text=black]
at (rel axis cs:0.715883668903803,-0.0471698113207547,0) {SDSRG};
\node[left, align=right, inner sep=0mm, rotate=90, text=black]
at (rel axis cs:0.753914988814318,-0.0471698113207547,0) {SigSal};
\node[left, align=right, inner sep=0mm, rotate=90, text=black]
at (rel axis cs:0.794183445190157,-0.0471698113207547,0) {SIMFine};
\node[left, align=right, inner sep=0mm, rotate=90, text=black]
at (rel axis cs:0.834451901565995,-0.0471698113207547,0) {SIMCoarse};
\node[left, align=right, inner sep=0mm, rotate=90, text=black]
at (rel axis cs:0.874720357941834,-0.0471698113207547,0) {SpectRes};
\node[left, align=right, inner sep=0mm, rotate=90, text=black]
at (rel axis cs:0.914988814317673,-0.0471698113207547,0) {SUN};
\node[left, align=right, inner sep=0mm, rotate=90, text=black]
at (rel axis cs:0.955257270693512,-0.0471698113207547,0) {Torralba};
\end{axis}
\end{tikzpicture}%
}

%
\caption{Metric scores for the Toronto~\cite{AIM} and MIT~\cite{MIT} databases.}
\label{fig:TorontoMIT}
\end{figure*}
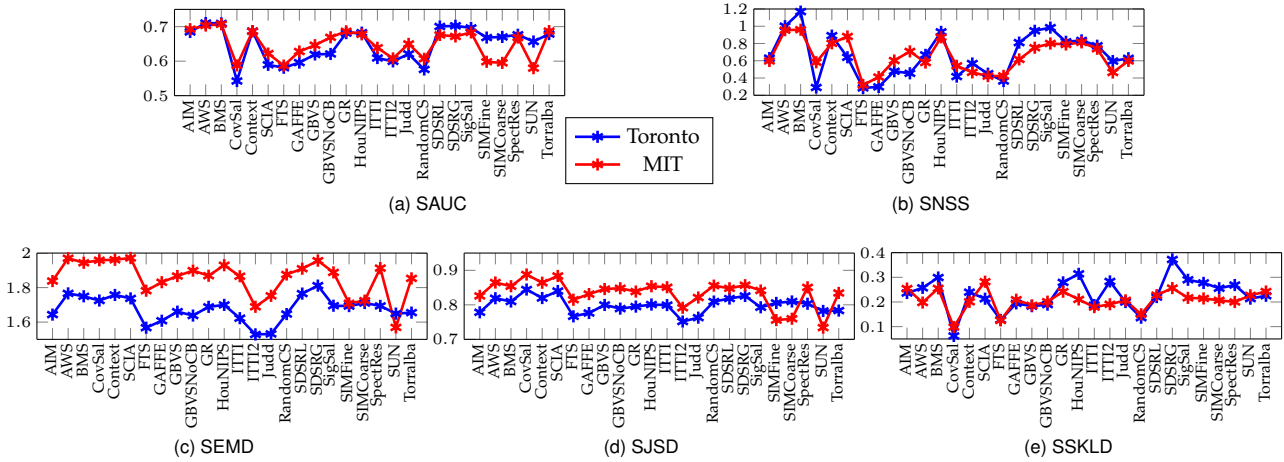
\begin{table}[t]
	\centering
	\caption{Consistency of performance for existing and proposed metrics over the MIT~\cite{MIT} and Toronto~\cite{AIM} datasets in terms of the Kendall's coefficient of concordance (KCC).}
	\begin{tabular}{cccccc}
		\toprule
		 & SAUC  & SNSS  & SSKLD & SEMD & SJSD    \\
		\midrule
		KCC & 0.8978 & \bf{0.9265}  & 0.7778  & 0.8704 & 0.8678 \\
		\bottomrule
	\end{tabular}%
	\label{tab:PerfConsistencyTorontoMIT}%
\end{table}%

\subsubsection{Variation in performance with varying distortion types and levels for the TUD Interactions database}
\label{subsec:performance distortion types}
As indicated previously, the TUD Interactions database has three different types of distortions: Gaussian blur, JPEG compression and Gaussian noise. Each of these distortion types are present in three different levels, high, medium and low. 
To check the performance over different distortion types, as well as the effect of the different levels of distortions have on the predicted saliency given by different models, we compute metric scores for each image and then average them over those set of images that belong to a particular distortion type and a particular level of distortion. 

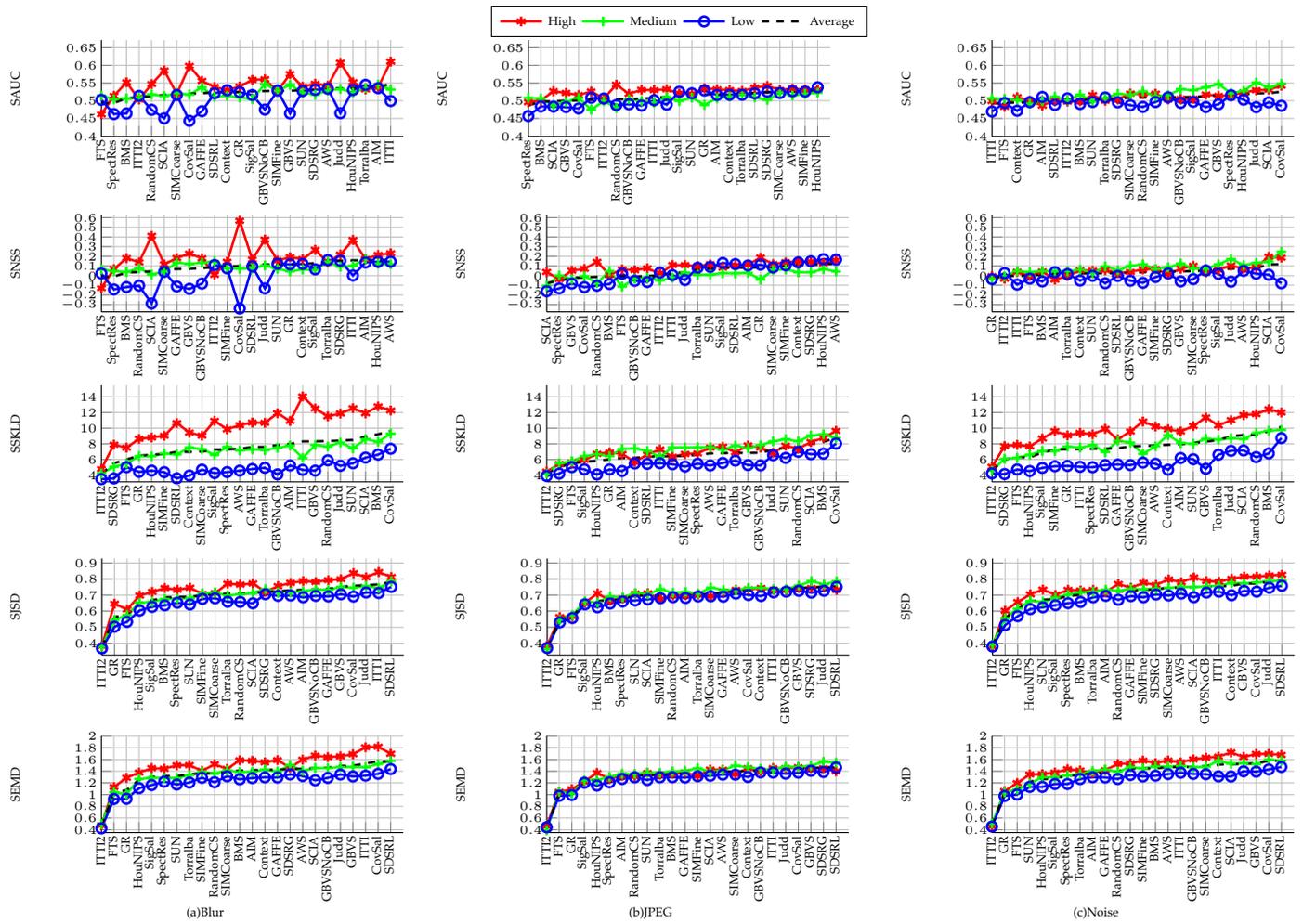
\begin{figure*}[]
	    \centering
	    \tiny
		\setlength{\figureheight}{0.08\textwidth}
		\setlength{\figurewidth}{0.25\textwidth}
	\begin{tabular}{ccc}    
%
\begin{tikzpicture}

\begin{axis}[%
width=\figurewidth,
height=\figureheight,
clip=false,
scale only axis,
xmin=1,
xmax=25,
xtick={1,2,3,4,5,6,7,8,9,10,11,12,13,14,15,16,17,18,19,20,21,22,23,24},
xticklabels={\empty},
xmajorgrids,
ymin=0.39883389939283,
ymax=0.671652895645101,
ylabel={SAUC},
ymajorgrids,
axis x line*=bottom,
axis y line*=left,
ytick = {0.4 ,0.45,0.5 ,0.55,0.6 ,0.65},ylabel style={font=\tiny},legend style={font=\tiny},
]
\addplot [color=red,solid,line width=1.0pt,mark=asterisk,mark options={solid},forget plot]
  table[row sep=crcr]{1	0.461633660795434\\
2	0.51379359794644\\
3	0.552031884825757\\
4	0.503597449538042\\
5	0.547114999410845\\
6	0.585720496672006\\
7	0.519463048596588\\
8	0.598478066014809\\
9	0.555929185565372\\
10	0.539038851348863\\
11	0.530122879026346\\
12	0.540401391923069\\
13	0.559038463961013\\
14	0.560666337609829\\
15	0.527190030851621\\
16	0.57543646526714\\
17	0.539755143713458\\
18	0.547937778280028\\
19	0.542506445278021\\
20	0.607301796345831\\
21	0.552401078908262\\
22	0.532146069186501\\
23	0.538084232491604\\
24	0.610593541495546\\
};
\addplot [color=green,solid,line width=1.0pt,mark=+,mark options={solid},forget plot]
  table[row sep=crcr]{1	0.514805571534366\\
2	0.504524146950264\\
3	0.506396672007674\\
4	0.509112017997125\\
5	0.518055262844187\\
6	0.51300185519863\\
7	0.521795966085044\\
8	0.517410748088204\\
9	0.53781886077685\\
10	0.511407914333663\\
11	0.513675586374218\\
12	0.509506820726585\\
13	0.504647727293309\\
14	0.547105891635007\\
15	0.528448941513039\\
16	0.54495763005149\\
17	0.52291486190148\\
18	0.517376889603766\\
19	0.529001058381411\\
20	0.533542510867529\\
21	0.523096493728165\\
22	0.535573570636627\\
23	0.545861053847528\\
24	0.531768274944011\\
};
\addplot [color=blue,solid,line width=1.0pt,mark=o,mark options={solid},forget plot]
  table[row sep=crcr]{1	0.502533192674978\\
2	0.462338756984712\\
3	0.464086279694542\\
4	0.513626458004269\\
5	0.474435067406195\\
6	0.45017355340161\\
7	0.516169304675874\\
8	0.443148777103145\\
9	0.470483351025448\\
10	0.520957766335136\\
11	0.529697256130968\\
12	0.5241425738387\\
13	0.516137940935909\\
14	0.474920216141273\\
15	0.5286485235275\\
16	0.464113060717046\\
17	0.527652541142055\\
18	0.530891948120584\\
19	0.534029008518082\\
20	0.464807317439792\\
21	0.530481978178069\\
22	0.544978814515916\\
23	0.535613060552233\\
24	0.499431326195519\\
};
\addplot [color=black,dashed,line width=1.0pt,forget plot]
  table[row sep=crcr]{1	0.492990808334926\\
2	0.493552167293805\\
3	0.507504945509324\\
4	0.508778641846479\\
5	0.513201776553742\\
6	0.516298635090748\\
7	0.519142773119169\\
8	0.519679197068719\\
9	0.521410465789223\\
10	0.523801510672554\\
11	0.524498573843844\\
12	0.524683595496118\\
13	0.52660804406341\\
14	0.527564148462036\\
15	0.528095831964053\\
16	0.528169052011892\\
17	0.530107515585664\\
18	0.53206887200146\\
19	0.535178837392505\\
20	0.535217208217717\\
21	0.535326516938165\\
22	0.537566151446348\\
23	0.539852782297122\\
24	0.547264380878359\\
};
\node[left, inner sep=0mm, rotate=90, text=black]
at (rel axis cs:-0.00202020202020202,-0.0298507462686567,0) {FTS};
\node[left, inner sep=0mm, rotate=90, text=black]
at (rel axis cs:0.0383838383838384,-0.0298507462686567,0) {SpectRes};
\node[left, inner sep=0mm, rotate=90, text=black]
at (rel axis cs:0.0808080808080808,-0.0298507462686567,0) {BMS};
\node[left, inner sep=0mm, rotate=90, text=black]
at (rel axis cs:0.121212121212121,-0.0298507462686567,0) {ITTI2};
\node[left, inner sep=0mm, rotate=90, text=black]
at (rel axis cs:0.163636363636364,-0.0298507462686567,0) {RandomCS};
\node[left, inner sep=0mm, rotate=90, text=black]
at (rel axis cs:0.204040404040404,-0.0298507462686567,0) {SCIA};
\node[left, inner sep=0mm, rotate=90, text=black]
at (rel axis cs:0.246464646464646,-0.0298507462686567,0) {SIMCoarse};
\node[left, inner sep=0mm, rotate=90, text=black]
at (rel axis cs:0.288888888888889,-0.0298507462686567,0) {CovSal};
\node[left, inner sep=0mm, rotate=90, text=black]
at (rel axis cs:0.329292929292929,-0.0298507462686567,0) {GAFFE};
\node[left, inner sep=0mm, rotate=90, text=black]
at (rel axis cs:0.371717171717172,-0.0298507462686567,0) {SDSRL};
\node[left, inner sep=0mm, rotate=90, text=black]
at (rel axis cs:0.412121212121212,-0.0298507462686567,0) {Context};
\node[left, inner sep=0mm, rotate=90, text=black]
at (rel axis cs:0.454545454545455,-0.0298507462686567,0) {GR};
\node[left, inner sep=0mm, rotate=90, text=black]
at (rel axis cs:0.496969696969697,-0.0298507462686567,0) {SigSal};
\node[left, inner sep=0mm, rotate=90, text=black]
at (rel axis cs:0.537373737373737,-0.0298507462686567,0) {GBVSNoCB};
\node[left, inner sep=0mm, rotate=90, text=black]
at (rel axis cs:0.57979797979798,-0.0298507462686567,0) {SIMFine};
\node[left, inner sep=0mm, rotate=90, text=black]
at (rel axis cs:0.62020202020202,-0.0298507462686567,0) {GBVS};
\node[left, inner sep=0mm, rotate=90, text=black]
at (rel axis cs:0.662626262626263,-0.0298507462686567,0) {SUN};
\node[left, inner sep=0mm, rotate=90, text=black]
at (rel axis cs:0.703030303030303,-0.0298507462686567,0) {SDSRG};
\node[left, inner sep=0mm, rotate=90, text=black]
at (rel axis cs:0.745454545454545,-0.0298507462686567,0) {AWS};
\node[left, inner sep=0mm, rotate=90, text=black]
at (rel axis cs:0.787878787878788,-0.0298507462686567,0) {Judd};
\node[left, inner sep=0mm, rotate=90, text=black]
at (rel axis cs:0.828282828282828,-0.0298507462686567,0) {HouNIPS};
\node[left, inner sep=0mm, rotate=90, text=black]
at (rel axis cs:0.870707070707071,-0.0298507462686567,0) {Torralba};
\node[left, inner sep=0mm, rotate=90, text=black]
at (rel axis cs:0.911111111111111,-0.0298507462686567,0) {AIM};
\node[left, inner sep=0mm, rotate=90, text=black]
at (rel axis cs:0.953535353535354,-0.0298507462686567,0) {ITTI};
\end{axis}
\end{tikzpicture}%
 & 
%
\begin{tikzpicture}

\begin{axis}[%
width=\figurewidth,
height=\figureheight,
clip=false,
scale only axis,
xmin=1,
xmax=25,
xtick={1,2,3,4,5,6,7,8,9,10,11,12,13,14,15,16,17,18,19,20,21,22,23,24},
xticklabels={\empty},
xmajorgrids,
ymin=0.39883389939283,
ymax=0.671652895645101,
ylabel={SAUC},
ymajorgrids,
axis x line*=bottom,
axis y line*=left,
legend style={at={(0.5,1.03)},anchor=south,legend columns=4,draw=black,fill=white,legend cell align=left},
ytick = {0.4 ,0.45,0.5 ,0.55,0.6 ,0.65},ylabel style={font=\tiny},legend style={font=\tiny},
]
\addplot [color=red,solid,line width=1.0pt,mark=asterisk,mark options={solid}]
  table[row sep=crcr]{1	0.495941106335809\\
2	0.500807536129448\\
3	0.526886661453403\\
4	0.522031711703209\\
5	0.516096988639112\\
6	0.526100231953054\\
7	0.506417156740167\\
8	0.545455259019126\\
9	0.519555870867191\\
10	0.530839413493922\\
11	0.529994717495584\\
12	0.53242588302016\\
13	0.520609895157124\\
14	0.519046898496093\\
15	0.535648218310601\\
16	0.532242086558814\\
17	0.530269256829053\\
18	0.527732408999589\\
19	0.536840867954945\\
20	0.541897278229359\\
21	0.530366076384401\\
22	0.534299361924891\\
23	0.531856147330465\\
24	0.530379774417523\\
};
\addlegendentry{High};

\addplot [color=green,solid,line width=1.0pt,mark=+,mark options={solid}]
  table[row sep=crcr]{1	0.508133192102048\\
2	0.50661969281314\\
3	0.486127241335889\\
4	0.496968617460677\\
5	0.507811919069835\\
6	0.474818216712801\\
7	0.499151932647852\\
8	0.479905209721921\\
9	0.50465297044918\\
10	0.499351384786298\\
11	0.494870853803642\\
12	0.510790269553636\\
13	0.499290277885077\\
14	0.508841889911966\\
15	0.488344910670054\\
16	0.507209272988164\\
17	0.511572161081394\\
18	0.516905157408552\\
19	0.509695078293992\\
20	0.502276338013934\\
21	0.522806823374403\\
22	0.514850685216088\\
23	0.519477745283203\\
24	0.522834845263606\\
};
\addlegendentry{Medium};

\addplot [color=blue,solid,line width=1.0pt,mark=o,mark options={solid}]
  table[row sep=crcr]{1	0.456895021430144\\
2	0.483767148480938\\
3	0.483369524025001\\
4	0.481950837396005\\
5	0.478034544353525\\
6	0.508379699271483\\
7	0.506318043776888\\
8	0.487663554249708\\
9	0.489974190085717\\
10	0.486392529865874\\
11	0.501601704236143\\
12	0.489069879488506\\
13	0.525413291825888\\
14	0.521049649671404\\
15	0.528985106761347\\
16	0.516048819792692\\
17	0.516811903018738\\
18	0.516728756778375\\
19	0.520626074429272\\
20	0.525397408535005\\
21	0.521822641575322\\
22	0.527023800757779\\
23	0.525946063742017\\
24	0.538114852918667\\
};
\addlegendentry{Low};

\addplot [color=black,dashed,line width=1.0pt]
  table[row sep=crcr]{1	0.486989773289334\\
2	0.497064792474509\\
3	0.498794475604764\\
4	0.500317055519964\\
5	0.500647817354157\\
6	0.503099382645779\\
7	0.503962377721636\\
8	0.504341340996918\\
9	0.504727677134029\\
10	0.505527776048698\\
11	0.508822425178456\\
12	0.510762010687434\\
13	0.515104488289363\\
14	0.516312812693154\\
15	0.517659411914\\
16	0.51850005977989\\
17	0.519551106976395\\
18	0.520455441062172\\
19	0.52238734022607\\
20	0.523190341592766\\
21	0.524998513778042\\
22	0.525391282632919\\
23	0.525759985451895\\
24	0.530443157533266\\
};
\addlegendentry{Average};

\node[left, inner sep=0mm, rotate=90, text=black]
at (rel axis cs:-0.00202020202020202,-0.0298507462686567,0) {SpectRes};
\node[left, inner sep=0mm, rotate=90, text=black]
at (rel axis cs:0.0383838383838384,-0.0298507462686567,0) {BMS};
\node[left, inner sep=0mm, rotate=90, text=black]
at (rel axis cs:0.0808080808080808,-0.0298507462686567,0) {SCIA};
\node[left, inner sep=0mm, rotate=90, text=black]
at (rel axis cs:0.121212121212121,-0.0298507462686567,0) {GBVS};
\node[left, inner sep=0mm, rotate=90, text=black]
at (rel axis cs:0.163636363636364,-0.0298507462686567,0) {CovSal};
\node[left, inner sep=0mm, rotate=90, text=black]
at (rel axis cs:0.204040404040404,-0.0298507462686567,0) {FTS};
\node[left, inner sep=0mm, rotate=90, text=black]
at (rel axis cs:0.246464646464646,-0.0298507462686567,0) {ITTI2};
\node[left, inner sep=0mm, rotate=90, text=black]
at (rel axis cs:0.288888888888889,-0.0298507462686567,0) {RandomCS};
\node[left, inner sep=0mm, rotate=90, text=black]
at (rel axis cs:0.329292929292929,-0.0298507462686567,0) {GBVSNoCB};
\node[left, inner sep=0mm, rotate=90, text=black]
at (rel axis cs:0.371717171717172,-0.0298507462686567,0) {GAFFE};
\node[left, inner sep=0mm, rotate=90, text=black]
at (rel axis cs:0.412121212121212,-0.0298507462686567,0) {ITTI};
\node[left, inner sep=0mm, rotate=90, text=black]
at (rel axis cs:0.454545454545455,-0.0298507462686567,0) {Judd};
\node[left, inner sep=0mm, rotate=90, text=black]
at (rel axis cs:0.496969696969697,-0.0298507462686567,0) {SigSal};
\node[left, inner sep=0mm, rotate=90, text=black]
at (rel axis cs:0.537373737373737,-0.0298507462686567,0) {SUN};
\node[left, inner sep=0mm, rotate=90, text=black]
at (rel axis cs:0.57979797979798,-0.0298507462686567,0) {GR};
\node[left, inner sep=0mm, rotate=90, text=black]
at (rel axis cs:0.62020202020202,-0.0298507462686567,0) {AIM};
\node[left, inner sep=0mm, rotate=90, text=black]
at (rel axis cs:0.662626262626263,-0.0298507462686567,0) {Context};
\node[left, inner sep=0mm, rotate=90, text=black]
at (rel axis cs:0.703030303030303,-0.0298507462686567,0) {Torralba};
\node[left, inner sep=0mm, rotate=90, text=black]
at (rel axis cs:0.745454545454545,-0.0298507462686567,0) {SDSRL};
\node[left, inner sep=0mm, rotate=90, text=black]
at (rel axis cs:0.787878787878788,-0.0298507462686567,0) {SDSRG};
\node[left, inner sep=0mm, rotate=90, text=black]
at (rel axis cs:0.828282828282828,-0.0298507462686567,0) {SIMCoarse};
\node[left, inner sep=0mm, rotate=90, text=black]
at (rel axis cs:0.870707070707071,-0.0298507462686567,0) {AWS};
\node[left, inner sep=0mm, rotate=90, text=black]
at (rel axis cs:0.911111111111111,-0.0298507462686567,0) {SIMFine};
\node[left, inner sep=0mm, rotate=90, text=black]
at (rel axis cs:0.953535353535354,-0.0298507462686567,0) {HouNIPS};
\end{axis}
\end{tikzpicture}%
 & 
%
\begin{tikzpicture}

\begin{axis}[%
width=\figurewidth,
height=\figureheight,
clip=false,
scale only axis,
xmin=1,
xmax=25,
xtick={1,2,3,4,5,6,7,8,9,10,11,12,13,14,15,16,17,18,19,20,21,22,23,24},
xticklabels={\empty},
xmajorgrids,
ymin=0.39883389939283,
ymax=0.671652895645101,
ylabel={SAUC},
ymajorgrids,
axis x line*=bottom,
axis y line*=left,
ytick = {0.4 ,0.45,0.5 ,0.55,0.6 ,0.65},ylabel style={font=\tiny},legend style={font=\tiny},
]
\addplot [color=red,solid,line width=1.0pt,mark=asterisk,mark options={solid},forget plot]
  table[row sep=crcr]{1	0.499190657676522\\
2	0.482853501143649\\
3	0.510241486923367\\
4	0.499951719733332\\
5	0.486179555738442\\
6	0.501898319871465\\
7	0.497671873702257\\
8	0.49892707508022\\
9	0.515998830656108\\
10	0.499978269921958\\
11	0.50387939931008\\
12	0.520457901185003\\
13	0.518961502127011\\
14	0.520112173185397\\
15	0.503814155664537\\
16	0.504996077396396\\
17	0.505573868536213\\
18	0.517679808774928\\
19	0.513722001077916\\
20	0.515268219259437\\
21	0.523433543232784\\
22	0.529651917679723\\
23	0.53121093917001\\
24	0.542514533886668\\
};
\addplot [color=green,solid,line width=1.0pt,mark=+,mark options={solid},forget plot]
  table[row sep=crcr]{1	0.507043154561433\\
2	0.505306458460785\\
3	0.502193043180288\\
4	0.491689638516822\\
5	0.498762308540512\\
6	0.511293802360338\\
7	0.501499947019655\\
8	0.516939994888068\\
9	0.497461707713197\\
10	0.507243529611835\\
11	0.51939803002712\\
12	0.517254280453503\\
13	0.526479644697841\\
14	0.51268038785816\\
15	0.515751638233754\\
16	0.532810659150625\\
17	0.529739727333086\\
18	0.537884707545845\\
19	0.54708354284819\\
20	0.521133443438036\\
21	0.528088574459097\\
22	0.551741359117338\\
23	0.537553637597406\\
24	0.548222187758615\\
};
\addplot [color=blue,solid,line width=1.0pt,mark=o,mark options={solid},forget plot]
  table[row sep=crcr]{1	0.469460811668073\\
2	0.493780028464146\\
3	0.471667633468324\\
4	0.49612120713785\\
5	0.510671735071611\\
6	0.488126952198492\\
7	0.506731626928325\\
8	0.490957294151281\\
9	0.496453220210181\\
10	0.509226913094459\\
11	0.495534390673675\\
12	0.487285973585987\\
13	0.482462860545575\\
14	0.496465871021726\\
15	0.509711757484602\\
16	0.493961501565089\\
17	0.497131375271292\\
18	0.482038034701078\\
19	0.490096111881669\\
20	0.515637366413287\\
21	0.503082913977642\\
22	0.481689256402176\\
23	0.495359085107892\\
24	0.48584941024382\\
};
\addplot [color=black,dashed,line width=1.0pt,forget plot]
  table[row sep=crcr]{1	0.491898207968676\\
2	0.49397999602286\\
3	0.49470072119066\\
4	0.495920855129335\\
5	0.498537866450188\\
6	0.500439691476765\\
7	0.501967815883412\\
8	0.502274788039856\\
9	0.503304586193162\\
10	0.505482904209417\\
11	0.506270606670292\\
12	0.508332718408164\\
13	0.509301335790142\\
14	0.509752810688427\\
15	0.509759183794298\\
16	0.510589412704036\\
17	0.510814990380197\\
18	0.51253418367395\\
19	0.516967218602592\\
20	0.51734634303692\\
21	0.518201677223174\\
22	0.521027511066413\\
23	0.521374553958436\\
24	0.525528710629701\\
};
\node[left, inner sep=0mm, rotate=90, text=black]
at (rel axis cs:-0.00202020202020202,-0.0298507462686567,0) {ITTI};
\node[left, inner sep=0mm, rotate=90, text=black]
at (rel axis cs:0.0383838383838384,-0.0298507462686567,0) {FTS};
\node[left, inner sep=0mm, rotate=90, text=black]
at (rel axis cs:0.0808080808080808,-0.0298507462686567,0) {Context};
\node[left, inner sep=0mm, rotate=90, text=black]
at (rel axis cs:0.121212121212121,-0.0298507462686567,0) {GR};
\node[left, inner sep=0mm, rotate=90, text=black]
at (rel axis cs:0.163636363636364,-0.0298507462686567,0) {AIM};
\node[left, inner sep=0mm, rotate=90, text=black]
at (rel axis cs:0.204040404040404,-0.0298507462686567,0) {SDSRL};
\node[left, inner sep=0mm, rotate=90, text=black]
at (rel axis cs:0.246464646464646,-0.0298507462686567,0) {ITTI2};
\node[left, inner sep=0mm, rotate=90, text=black]
at (rel axis cs:0.288888888888889,-0.0298507462686567,0) {BMS};
\node[left, inner sep=0mm, rotate=90, text=black]
at (rel axis cs:0.329292929292929,-0.0298507462686567,0) {SUN};
\node[left, inner sep=0mm, rotate=90, text=black]
at (rel axis cs:0.371717171717172,-0.0298507462686567,0) {Torralba};
\node[left, inner sep=0mm, rotate=90, text=black]
at (rel axis cs:0.412121212121212,-0.0298507462686567,0) {SDSRG};
\node[left, inner sep=0mm, rotate=90, text=black]
at (rel axis cs:0.454545454545455,-0.0298507462686567,0) {SIMCoarse};
\node[left, inner sep=0mm, rotate=90, text=black]
at (rel axis cs:0.496969696969697,-0.0298507462686567,0) {RandomCS};
\node[left, inner sep=0mm, rotate=90, text=black]
at (rel axis cs:0.537373737373737,-0.0298507462686567,0) {SIMFine};
\node[left, inner sep=0mm, rotate=90, text=black]
at (rel axis cs:0.57979797979798,-0.0298507462686567,0) {AWS};
\node[left, inner sep=0mm, rotate=90, text=black]
at (rel axis cs:0.62020202020202,-0.0298507462686567,0) {GBVSNoCB};
\node[left, inner sep=0mm, rotate=90, text=black]
at (rel axis cs:0.662626262626263,-0.0298507462686567,0) {SigSal};
\node[left, inner sep=0mm, rotate=90, text=black]
at (rel axis cs:0.703030303030303,-0.0298507462686567,0) {GAFFE};
\node[left, inner sep=0mm, rotate=90, text=black]
at (rel axis cs:0.745454545454545,-0.0298507462686567,0) {GBVS};
\node[left, inner sep=0mm, rotate=90, text=black]
at (rel axis cs:0.787878787878788,-0.0298507462686567,0) {SpectRes};
\node[left, inner sep=0mm, rotate=90, text=black]
at (rel axis cs:0.828282828282828,-0.0298507462686567,0) {HouNIPS};
\node[left, inner sep=0mm, rotate=90, text=black]
at (rel axis cs:0.870707070707071,-0.0298507462686567,0) {Judd};
\node[left, inner sep=0mm, rotate=90, text=black]
at (rel axis cs:0.911111111111111,-0.0298507462686567,0) {SCIA};
\node[left, inner sep=0mm, rotate=90, text=black]
at (rel axis cs:0.953535353535354,-0.0298507462686567,0) {CovSal};
\end{axis}
\end{tikzpicture}%
 \\
%
\begin{tikzpicture}

\begin{axis}[%
width=\figurewidth,
height=\figureheight,
clip=false,
scale only axis,
xmin=1,
xmax=25,
xtick={1,2,3,4,5,6,7,8,9,10,11,12,13,14,15,16,17,18,19,20,21,22,23,24},
xticklabels={\empty},
xmajorgrids,
ymin=-0.375290086927764,
ymax=0.623604240329483,
ylabel={SNSS},
ymajorgrids,
axis x line*=bottom,
axis y line*=left,
ytick = {-0.3,-0.2,-0.1,0   ,0.1 ,0.2 ,0.3 ,0.4 ,0.5 ,0.6 },ylabel style={font=\tiny},legend style={font=\tiny},
]
\addplot [color=red,solid,line width=1.0pt,mark=asterisk,mark options={solid},forget plot]
  table[row sep=crcr]{1	-0.130353034642989\\
2	0.0672696850905239\\
3	0.180838141700289\\
4	0.137253013107377\\
5	0.411700257990596\\
6	0.116702511773336\\
7	0.181493832507006\\
8	0.22765048314449\\
9	0.178503408457361\\
10	0.00815151392144014\\
11	0.142672069283801\\
12	0.566912945754075\\
13	0.165453529731212\\
14	0.372720560402854\\
15	0.156281517271422\\
16	0.190773355751653\\
17	0.170836314448366\\
18	0.267084034884007\\
19	0.136366755828599\\
20	0.217771078781542\\
21	0.369048549688545\\
22	0.171144839066293\\
23	0.214496160669434\\
24	0.23154525004645\\
};
\addplot [color=green,solid,line width=1.0pt,mark=+,mark options={solid},forget plot]
  table[row sep=crcr]{1	0.065343711118737\\
2	0.0453389179244342\\
3	0.0383777264594501\\
4	0.0758390110511601\\
5	0.0106584920135423\\
6	0.0441926805559889\\
7	0.134614681874934\\
8	0.120720552910391\\
9	0.133030096569133\\
10	0.146890032945705\\
11	0.0744759786820817\\
12	0.0709705734595941\\
13	0.0738626501391972\\
14	0.104775267511359\\
15	0.0697152503721927\\
16	0.0444694761302872\\
17	0.0629724888911119\\
18	0.0583275893527867\\
19	0.143520621823411\\
20	0.0911795217514716\\
21	0.101379684907415\\
22	0.17227975408932\\
23	0.121744467930678\\
24	0.118317644996976\\
};
\addplot [color=blue,solid,line width=1.0pt,mark=o,mark options={solid},forget plot]
  table[row sep=crcr]{1	0.0213679932415904\\
2	-0.143556087464512\\
3	-0.119079953829117\\
4	-0.104382079143368\\
5	-0.289392342884366\\
6	0.0394562976790899\\
7	-0.112645649871364\\
8	-0.13786743094345\\
9	-0.0828621426807245\\
10	0.109049604379429\\
11	0.0746051435795953\\
12	-0.341172806297967\\
13	0.0955532233448311\\
14	-0.133163000658669\\
15	0.122620228517005\\
16	0.116490319358644\\
17	0.121034928487955\\
18	0.0647844056216578\\
19	0.160469621975931\\
20	0.153426522314279\\
21	0.00128018479161785\\
22	0.135237737815714\\
23	0.143253310981904\\
24	0.144880631015066\\
};
\addplot [color=black,dashed,line width=1.0pt,forget plot]
  table[row sep=crcr]{1	-0.0145471100942205\\
2	-0.0103158281498514\\
3	0.0333786381102071\\
4	0.0362366483383897\\
5	0.0443221357065908\\
6	0.066783830002805\\
7	0.0678209548368585\\
8	0.0701678683704769\\
9	0.0762237874485898\\
10	0.0880303837488579\\
11	0.0972510638484925\\
12	0.0989035709719007\\
13	0.11162313440508\\
14	0.114777609085182\\
15	0.116205665386873\\
16	0.117244383746861\\
17	0.118281243942478\\
18	0.130065343286151\\
19	0.146785666542647\\
20	0.154125707615764\\
21	0.157236139795859\\
22	0.159554110323776\\
23	0.159831313194006\\
24	0.164914508686164\\
};
\node[left, inner sep=0mm, rotate=90, text=black]
at (rel axis cs:-0.00202020202020202,-0.0298507462686567,0) {FTS};
\node[left, inner sep=0mm, rotate=90, text=black]
at (rel axis cs:0.0383838383838384,-0.0298507462686567,0) {SpectRes};
\node[left, inner sep=0mm, rotate=90, text=black]
at (rel axis cs:0.0808080808080808,-0.0298507462686567,0) {BMS};
\node[left, inner sep=0mm, rotate=90, text=black]
at (rel axis cs:0.121212121212121,-0.0298507462686567,0) {RandomCS};
\node[left, inner sep=0mm, rotate=90, text=black]
at (rel axis cs:0.163636363636364,-0.0298507462686567,0) {SCIA};
\node[left, inner sep=0mm, rotate=90, text=black]
at (rel axis cs:0.204040404040404,-0.0298507462686567,0) {SIMCoarse};
\node[left, inner sep=0mm, rotate=90, text=black]
at (rel axis cs:0.246464646464646,-0.0298507462686567,0) {GAFFE};
\node[left, inner sep=0mm, rotate=90, text=black]
at (rel axis cs:0.288888888888889,-0.0298507462686567,0) {GBVS};
\node[left, inner sep=0mm, rotate=90, text=black]
at (rel axis cs:0.329292929292929,-0.0298507462686567,0) {GBVSNoCB};
\node[left, inner sep=0mm, rotate=90, text=black]
at (rel axis cs:0.371717171717172,-0.0298507462686567,0) {ITTI2};
\node[left, inner sep=0mm, rotate=90, text=black]
at (rel axis cs:0.412121212121212,-0.0298507462686567,0) {SIMFine};
\node[left, inner sep=0mm, rotate=90, text=black]
at (rel axis cs:0.454545454545455,-0.0298507462686567,0) {CovSal};
\node[left, inner sep=0mm, rotate=90, text=black]
at (rel axis cs:0.496969696969697,-0.0298507462686567,0) {SDSRL};
\node[left, inner sep=0mm, rotate=90, text=black]
at (rel axis cs:0.537373737373737,-0.0298507462686567,0) {Judd};
\node[left, inner sep=0mm, rotate=90, text=black]
at (rel axis cs:0.57979797979798,-0.0298507462686567,0) {SUN};
\node[left, inner sep=0mm, rotate=90, text=black]
at (rel axis cs:0.62020202020202,-0.0298507462686567,0) {GR};
\node[left, inner sep=0mm, rotate=90, text=black]
at (rel axis cs:0.662626262626263,-0.0298507462686567,0) {Context};
\node[left, inner sep=0mm, rotate=90, text=black]
at (rel axis cs:0.703030303030303,-0.0298507462686567,0) {SigSal};
\node[left, inner sep=0mm, rotate=90, text=black]
at (rel axis cs:0.745454545454545,-0.0298507462686567,0) {Torralba};
\node[left, inner sep=0mm, rotate=90, text=black]
at (rel axis cs:0.787878787878788,-0.0298507462686567,0) {SDSRG};
\node[left, inner sep=0mm, rotate=90, text=black]
at (rel axis cs:0.828282828282828,-0.0298507462686567,0) {ITTI};
\node[left, inner sep=0mm, rotate=90, text=black]
at (rel axis cs:0.870707070707071,-0.0298507462686567,0) {AIM};
\node[left, inner sep=0mm, rotate=90, text=black]
at (rel axis cs:0.911111111111111,-0.0298507462686567,0) {HouNIPS};
\node[left, inner sep=0mm, rotate=90, text=black]
at (rel axis cs:0.953535353535354,-0.0298507462686567,0) {AWS};
\end{axis}
\end{tikzpicture}%
 & 
%
\begin{tikzpicture}

\begin{axis}[%
width=\figurewidth,
height=\figureheight,
clip=false,
scale only axis,
xmin=1,
xmax=25,
xtick={1,2,3,4,5,6,7,8,9,10,11,12,13,14,15,16,17,18,19,20,21,22,23,24},
xticklabels={\empty},
xmajorgrids,
ymin=-0.375290086927764,
ymax=0.623604240329483,
ylabel={SNSS},
ymajorgrids,
axis x line*=bottom,
axis y line*=left,
ytick = {-0.3,-0.2,-0.1,0   ,0.1 ,0.2 ,0.3 ,0.4 ,0.5 ,0.6 },ylabel style={font=\tiny},legend style={font=\tiny},
]
\addplot [color=red,solid,line width=1.0pt,mark=asterisk,mark options={solid},forget plot]
  table[row sep=crcr]{1	0.0374951541069412\\
2	-0.0379728721084319\\
3	0.0511657480316758\\
4	0.0693723312215175\\
5	0.142327461701612\\
6	0.00774559663918063\\
7	0.0636179718129617\\
8	0.0560200755124477\\
9	0.0761691399551728\\
10	0.0204508798496345\\
11	0.109353147649403\\
12	0.1083338719284\\
13	0.0982332290692218\\
14	0.117225203788224\\
15	0.0815740672381406\\
16	0.100839290855479\\
17	0.113235252474327\\
18	0.186072831228927\\
19	0.11856000090896\\
20	0.137989757352202\\
21	0.147947454727109\\
22	0.144019406315331\\
23	0.125606942243318\\
24	0.157877283305674\\
};
\addplot [color=green,solid,line width=1.0pt,mark=+,mark options={solid},forget plot]
  table[row sep=crcr]{1	-0.105080173781564\\
2	0.00174248882725301\\
3	-0.037137374112365\\
4	-0.00664324568190755\\
5	-0.0757373473265317\\
6	0.0465558819255886\\
7	-0.113988473579287\\
8	-0.0321791411323349\\
9	-0.0210804993225911\\
10	-0.049442245760012\\
11	-0.021480996336342\\
12	0.0319758265496789\\
13	0.00857115666244261\\
14	0.0104166310245941\\
15	0.0248145240835244\\
16	0.022352929397877\\
17	0.0293811889212174\\
18	-0.0410681237081354\\
19	0.0583783371138555\\
20	0.0646641299165908\\
21	0.0338312861738332\\
22	0.0335046639940261\\
23	0.0673663426336457\\
24	0.0422241785567212\\
};
\addplot [color=blue,solid,line width=1.0pt,mark=o,mark options={solid},forget plot]
  table[row sep=crcr]{1	-0.162607635972026\\
2	-0.13267208263132\\
3	-0.0822791831377443\\
4	-0.120303687231971\\
5	-0.104583147781942\\
6	-0.0868438395080035\\
7	0.0184211754097018\\
8	-0.0543355483957355\\
9	-0.0664327075789557\\
10	0.0303181810813161\\
11	0.00498900349842637\\
12	-0.0462236943412525\\
13	0.0810493537278766\\
14	0.0897959556324614\\
15	0.130940353675198\\
16	0.1186139553898\\
17	0.104053004045747\\
18	0.112855688784059\\
19	0.0837700151761449\\
20	0.106476116472482\\
21	0.136116576774763\\
22	0.148041742426036\\
23	0.169881957301726\\
24	0.165268792290072\\
};
\addplot [color=black,dashed,line width=1.0pt,forget plot]
  table[row sep=crcr]{1	-0.0767308852155498\\
2	-0.0563008219708329\\
3	-0.0227502697394778\\
4	-0.0191915338974538\\
5	-0.0126643444689537\\
6	-0.0108474536477448\\
7	-0.010649775452208\\
8	-0.0101648713385409\\
9	-0.00378135564879133\\
10	0.000442271723646201\\
11	0.0309537182704958\\
12	0.0313620013789422\\
13	0.0626179131531803\\
14	0.0724792634817599\\
15	0.0791096483322875\\
16	0.0806020585477185\\
17	0.0822231484804306\\
18	0.0859534654349502\\
19	0.0869027843996535\\
20	0.103043334580425\\
21	0.105965105891902\\
22	0.108521937578464\\
23	0.120951747392897\\
24	0.121790084717489\\
};
\node[left, inner sep=0mm, rotate=90, text=black]
at (rel axis cs:-0.00202020202020202,-0.0298507462686567,0) {SCIA};
\node[left, inner sep=0mm, rotate=90, text=black]
at (rel axis cs:0.0383838383838384,-0.0298507462686567,0) {SpectRes};
\node[left, inner sep=0mm, rotate=90, text=black]
at (rel axis cs:0.0808080808080808,-0.0298507462686567,0) {GBVS};
\node[left, inner sep=0mm, rotate=90, text=black]
at (rel axis cs:0.121212121212121,-0.0298507462686567,0) {CovSal};
\node[left, inner sep=0mm, rotate=90, text=black]
at (rel axis cs:0.163636363636364,-0.0298507462686567,0) {RandomCS};
\node[left, inner sep=0mm, rotate=90, text=black]
at (rel axis cs:0.204040404040404,-0.0298507462686567,0) {BMS};
\node[left, inner sep=0mm, rotate=90, text=black]
at (rel axis cs:0.246464646464646,-0.0298507462686567,0) {FTS};
\node[left, inner sep=0mm, rotate=90, text=black]
at (rel axis cs:0.288888888888889,-0.0298507462686567,0) {GBVSNoCB};
\node[left, inner sep=0mm, rotate=90, text=black]
at (rel axis cs:0.329292929292929,-0.0298507462686567,0) {GAFFE};
\node[left, inner sep=0mm, rotate=90, text=black]
at (rel axis cs:0.371717171717172,-0.0298507462686567,0) {ITTI2};
\node[left, inner sep=0mm, rotate=90, text=black]
at (rel axis cs:0.412121212121212,-0.0298507462686567,0) {ITTI};
\node[left, inner sep=0mm, rotate=90, text=black]
at (rel axis cs:0.454545454545455,-0.0298507462686567,0) {Judd};
\node[left, inner sep=0mm, rotate=90, text=black]
at (rel axis cs:0.496969696969697,-0.0298507462686567,0) {Torralba};
\node[left, inner sep=0mm, rotate=90, text=black]
at (rel axis cs:0.537373737373737,-0.0298507462686567,0) {SUN};
\node[left, inner sep=0mm, rotate=90, text=black]
at (rel axis cs:0.57979797979798,-0.0298507462686567,0) {SigSal};
\node[left, inner sep=0mm, rotate=90, text=black]
at (rel axis cs:0.62020202020202,-0.0298507462686567,0) {SDSRL};
\node[left, inner sep=0mm, rotate=90, text=black]
at (rel axis cs:0.662626262626263,-0.0298507462686567,0) {AIM};
\node[left, inner sep=0mm, rotate=90, text=black]
at (rel axis cs:0.703030303030303,-0.0298507462686567,0) {GR};
\node[left, inner sep=0mm, rotate=90, text=black]
at (rel axis cs:0.745454545454545,-0.0298507462686567,0) {SIMCoarse};
\node[left, inner sep=0mm, rotate=90, text=black]
at (rel axis cs:0.787878787878788,-0.0298507462686567,0) {SIMFine};
\node[left, inner sep=0mm, rotate=90, text=black]
at (rel axis cs:0.828282828282828,-0.0298507462686567,0) {Context};
\node[left, inner sep=0mm, rotate=90, text=black]
at (rel axis cs:0.870707070707071,-0.0298507462686567,0) {SDSRG};
\node[left, inner sep=0mm, rotate=90, text=black]
at (rel axis cs:0.911111111111111,-0.0298507462686567,0) {HouNIPS};
\node[left, inner sep=0mm, rotate=90, text=black]
at (rel axis cs:0.953535353535354,-0.0298507462686567,0) {AWS};
\end{axis}
\end{tikzpicture}%
 & 
%
\begin{tikzpicture}

\begin{axis}[%
width=\figurewidth,
height=\figureheight,
clip=false,
scale only axis,
xmin=1,
xmax=25,
xtick={1,2,3,4,5,6,7,8,9,10,11,12,13,14,15,16,17,18,19,20,21,22,23,24},
xticklabels={\empty},
xmajorgrids,
ymin=-0.375290086927764,
ymax=0.623604240329483,
ylabel={SNSS},
ymajorgrids,
axis x line*=bottom,
axis y line*=left,
ytick = {-0.3,-0.2,-0.1,0   ,0.1 ,0.2 ,0.3 ,0.4 ,0.5 ,0.6 },ylabel style={font=\tiny},legend style={font=\tiny},
]
\addplot [color=red,solid,line width=1.0pt,mark=asterisk,mark options={solid},forget plot]
  table[row sep=crcr]{1	-0.00659766429075758\\
2	-0.0330217816596546\\
3	0.0386341235424976\\
4	-0.00289778049568414\\
5	0.0414347682274803\\
6	-0.0428032735750164\\
7	-0.00415726874341989\\
8	0.0534385998455382\\
9	0.0509546181887578\\
10	0.0568678179442128\\
11	0.0282136056174831\\
12	0.0378160852838671\\
13	0.0654038071498931\\
14	0.0668915348038837\\
15	0.0143744272689366\\
16	0.070603175142207\\
17	0.10127723077392\\
18	0.0330396269159784\\
19	0.0598986327263005\\
20	0.104388822734892\\
21	0.069877197916185\\
22	0.0848429766633518\\
23	0.195549298300367\\
24	0.18527982386746\\
};
\addplot [color=green,solid,line width=1.0pt,mark=+,mark options={solid},forget plot]
  table[row sep=crcr]{1	-0.0344064512076831\\
2	-0.00858623951042968\\
3	0.0465798740938926\\
4	0.0371094058731905\\
5	0.0464108672489608\\
6	0.043964233780153\\
7	0.044378314295316\\
8	0.0612285659247184\\
9	0.0232678325683882\\
10	0.0950731843584169\\
11	0.0547220613447259\\
12	0.100241325329351\\
13	0.117199681081296\\
14	0.0673963934420625\\
15	0.0848089551530864\\
16	0.12492819845435\\
17	0.0714428728179284\\
18	0.0685527390992686\\
19	0.117020341316894\\
20	0.172182227004405\\
21	0.107233994974771\\
22	0.132618080124908\\
23	0.14364994401606\\
24	0.247279353552089\\
};
\addplot [color=blue,solid,line width=1.0pt,mark=o,mark options={solid},forget plot]
  table[row sep=crcr]{1	-0.0371840399700205\\
2	0.0257239491327349\\
3	-0.0930267539356541\\
4	-0.0318222114668429\\
5	-0.0603808187096238\\
6	0.0349854923291922\\
7	0.0151117618039117\\
8	-0.0512962758616503\\
9	-0.00243086658344372\\
10	-0.0795546635950989\\
11	-0.00387052801943848\\
12	-0.0530124106423193\\
13	-0.0746939202499947\\
14	-0.0153703653543146\\
15	0.02142662676308\\
16	-0.0600592638477565\\
17	-0.0364093360907105\\
18	0.0531702400997291\\
19	0.0173922728561139\\
20	-0.0633547073208181\\
21	0.0494432713097163\\
22	0.0195333986752608\\
23	0.00889954142072508\\
24	-0.0798897892738901\\
};
\addplot [color=black,dashed,line width=1.0pt,forget plot]
  table[row sep=crcr]{1	-0.0260627184894871\\
2	-0.00529469067911646\\
3	-0.00260425209975464\\
4	0.000796471303554492\\
5	0.00915493892227247\\
6	0.0120488175114429\\
7	0.0184442691186026\\
8	0.0211236299695354\\
9	0.0239305280579008\\
10	0.0241287795691769\\
11	0.0263550463142569\\
12	0.028348333323633\\
13	0.0359698559937314\\
14	0.0396391876305439\\
15	0.0402033363950343\\
16	0.0451573699162669\\
17	0.0454369225003792\\
18	0.0515875353716587\\
19	0.0647704156331028\\
20	0.0710721141394927\\
21	0.0755181547335573\\
22	0.0789981518211735\\
23	0.116032927912384\\
24	0.11755646271522\\
};
\node[left, inner sep=0mm, rotate=90, text=black]
at (rel axis cs:-0.00202020202020202,-0.0298507462686567,0) {GR};
\node[left, inner sep=0mm, rotate=90, text=black]
at (rel axis cs:0.0383838383838384,-0.0298507462686567,0) {ITTI2};
\node[left, inner sep=0mm, rotate=90, text=black]
at (rel axis cs:0.0808080808080808,-0.0298507462686567,0) {ITTI};
\node[left, inner sep=0mm, rotate=90, text=black]
at (rel axis cs:0.121212121212121,-0.0298507462686567,0) {FTS};
\node[left, inner sep=0mm, rotate=90, text=black]
at (rel axis cs:0.163636363636364,-0.0298507462686567,0) {BMS};
\node[left, inner sep=0mm, rotate=90, text=black]
at (rel axis cs:0.204040404040404,-0.0298507462686567,0) {AIM};
\node[left, inner sep=0mm, rotate=90, text=black]
at (rel axis cs:0.246464646464646,-0.0298507462686567,0) {Torralba};
\node[left, inner sep=0mm, rotate=90, text=black]
at (rel axis cs:0.288888888888889,-0.0298507462686567,0) {Context};
\node[left, inner sep=0mm, rotate=90, text=black]
at (rel axis cs:0.329292929292929,-0.0298507462686567,0) {SUN};
\node[left, inner sep=0mm, rotate=90, text=black]
at (rel axis cs:0.371717171717172,-0.0298507462686567,0) {RandomCS};
\node[left, inner sep=0mm, rotate=90, text=black]
at (rel axis cs:0.412121212121212,-0.0298507462686567,0) {SDSRL};
\node[left, inner sep=0mm, rotate=90, text=black]
at (rel axis cs:0.454545454545455,-0.0298507462686567,0) {GBVSNoCB};
\node[left, inner sep=0mm, rotate=90, text=black]
at (rel axis cs:0.496969696969697,-0.0298507462686567,0) {GAFFE};
\node[left, inner sep=0mm, rotate=90, text=black]
at (rel axis cs:0.537373737373737,-0.0298507462686567,0) {SIMFine};
\node[left, inner sep=0mm, rotate=90, text=black]
at (rel axis cs:0.57979797979798,-0.0298507462686567,0) {SDSRG};
\node[left, inner sep=0mm, rotate=90, text=black]
at (rel axis cs:0.62020202020202,-0.0298507462686567,0) {GBVS};
\node[left, inner sep=0mm, rotate=90, text=black]
at (rel axis cs:0.662626262626263,-0.0298507462686567,0) {SIMCoarse};
\node[left, inner sep=0mm, rotate=90, text=black]
at (rel axis cs:0.703030303030303,-0.0298507462686567,0) {SpectRes};
\node[left, inner sep=0mm, rotate=90, text=black]
at (rel axis cs:0.745454545454545,-0.0298507462686567,0) {SigSal};
\node[left, inner sep=0mm, rotate=90, text=black]
at (rel axis cs:0.787878787878788,-0.0298507462686567,0) {Judd};
\node[left, inner sep=0mm, rotate=90, text=black]
at (rel axis cs:0.828282828282828,-0.0298507462686567,0) {AWS};
\node[left, inner sep=0mm, rotate=90, text=black]
at (rel axis cs:0.870707070707071,-0.0298507462686567,0) {HouNIPS};
\node[left, inner sep=0mm, rotate=90, text=black]
at (rel axis cs:0.911111111111111,-0.0298507462686567,0) {SCIA};
\node[left, inner sep=0mm, rotate=90, text=black]
at (rel axis cs:0.953535353535354,-0.0298507462686567,0) {CovSal};
\end{axis}
\end{tikzpicture}%
 \\
%
\begin{tikzpicture}

\begin{axis}[%
width=\figurewidth,
height=\figureheight,
clip=false,
scale only axis,
xmin=1,
xmax=25,
xtick={1,2,3,4,5,6,7,8,9,10,11,12,13,14,15,16,17,18,19,20,21,22,23,24},
xticklabels={\empty},
xmajorgrids,
ymin=3.14917181669125,
ymax=15.4644661247791,
ylabel={SSKLD},
ymajorgrids,
axis x line*=bottom,
axis y line*=left,
ytick = {4 ,6 ,8 ,10,12,14},ylabel style={font=\tiny},legend style={font=\tiny},
]
\addplot [color=red,solid,line width=1.0pt,mark=asterisk,mark options={solid},forget plot]
  table[row sep=crcr]{1	4.7955666543788\\
2	7.9031804119595\\
3	7.50891719310472\\
4	8.63967638506837\\
5	8.80820701101136\\
6	9.02773686063952\\
7	10.6586655312439\\
8	9.43927545727088\\
9	9.04469630178643\\
10	10.9338966591234\\
11	9.85108347384432\\
12	10.3903010735329\\
13	10.7166126418322\\
14	10.6849380314611\\
15	11.918458869932\\
16	10.9166436792639\\
17	14.058605567981\\
18	12.5180571630602\\
19	11.5118415881494\\
20	11.8711852682721\\
21	12.5498545940881\\
22	11.9011698254175\\
23	12.7813754486822\\
24	12.2670962650917\\
};
\addplot [color=green,solid,line width=1.0pt,mark=+,mark options={solid},forget plot]
  table[row sep=crcr]{1	4.22903981651645\\
2	5.07061603650343\\
3	5.49597148334031\\
4	6.43542476909599\\
5	6.41629778588346\\
6	6.76756048739665\\
7	6.64564173329537\\
8	7.52652765553298\\
9	7.3288825253959\\
10	6.59791886104852\\
11	7.62679262321959\\
12	7.15030780135467\\
13	7.31447373154175\\
14	7.20269771714399\\
15	7.48901498699291\\
16	7.78322437415876\\
17	6.19069235636966\\
18	7.8456046544194\\
19	7.66571617452221\\
20	8.17267451589261\\
21	7.43227132599003\\
22	8.60977091685082\\
23	8.26237037237476\\
24	9.29931760213267\\
};
\addplot [color=blue,solid,line width=1.0pt,mark=o,mark options={solid},forget plot]
  table[row sep=crcr]{1	3.49907979632361\\
2	3.59188196156327\\
3	5.01734563728504\\
4	4.42479058169823\\
5	4.5851915005762\\
6	4.3875658916219\\
7	3.60338305046379\\
8	3.95670688495948\\
9	4.70591679336241\\
10	4.26918610888258\\
11	4.39818635472619\\
12	4.57971962803624\\
13	4.78787066571877\\
14	4.94426486007415\\
15	4.1044204232736\\
16	5.24618161834775\\
17	4.68036651285448\\
18	4.58141398660023\\
19	5.89832303669427\\
20	5.20041762045444\\
21	5.5237548573871\\
22	6.24478521773039\\
23	6.62687639593218\\
24	7.37567690239118\\
};
\addplot [color=black,dashed,line width=1.0pt,forget plot]
  table[row sep=crcr]{1	4.17456208907295\\
2	5.52189280334207\\
3	6.00741143791002\\
4	6.4999639119542\\
5	6.603232099157\\
6	6.72762107988603\\
7	6.96923010500101\\
8	6.97416999925445\\
9	7.02649854018158\\
10	7.26700054301817\\
11	7.29202081726337\\
12	7.37344283430795\\
13	7.6063190130309\\
14	7.6106335362264\\
15	7.83729809339951\\
16	7.98201655725679\\
17	8.30988814573504\\
18	8.31502526802661\\
19	8.35862693312197\\
20	8.41475913487306\\
21	8.50196025915506\\
22	8.91857531999957\\
23	9.22354073899638\\
24	9.64736358987186\\
};
\node[left, inner sep=0mm, rotate=90, text=black]
at (rel axis cs:-0.00202020202020202,-0.0298507462686567,0) {ITTI2};
\node[left, inner sep=0mm, rotate=90, text=black]
at (rel axis cs:0.0383838383838384,-0.0298507462686567,0) {SDSRG};
\node[left, inner sep=0mm, rotate=90, text=black]
at (rel axis cs:0.0808080808080808,-0.0298507462686567,0) {FTS};
\node[left, inner sep=0mm, rotate=90, text=black]
at (rel axis cs:0.121212121212121,-0.0298507462686567,0) {GR};
\node[left, inner sep=0mm, rotate=90, text=black]
at (rel axis cs:0.163636363636364,-0.0298507462686567,0) {HouNIPS};
\node[left, inner sep=0mm, rotate=90, text=black]
at (rel axis cs:0.204040404040404,-0.0298507462686567,0) {SIMFine};
\node[left, inner sep=0mm, rotate=90, text=black]
at (rel axis cs:0.246464646464646,-0.0298507462686567,0) {SDSRL};
\node[left, inner sep=0mm, rotate=90, text=black]
at (rel axis cs:0.288888888888889,-0.0298507462686567,0) {Context};
\node[left, inner sep=0mm, rotate=90, text=black]
at (rel axis cs:0.329292929292929,-0.0298507462686567,0) {SIMCoarse};
\node[left, inner sep=0mm, rotate=90, text=black]
at (rel axis cs:0.371717171717172,-0.0298507462686567,0) {SigSal};
\node[left, inner sep=0mm, rotate=90, text=black]
at (rel axis cs:0.412121212121212,-0.0298507462686567,0) {SpectRes};
\node[left, inner sep=0mm, rotate=90, text=black]
at (rel axis cs:0.454545454545455,-0.0298507462686567,0) {AWS};
\node[left, inner sep=0mm, rotate=90, text=black]
at (rel axis cs:0.496969696969697,-0.0298507462686567,0) {GAFFE};
\node[left, inner sep=0mm, rotate=90, text=black]
at (rel axis cs:0.537373737373737,-0.0298507462686567,0) {Torralba};
\node[left, inner sep=0mm, rotate=90, text=black]
at (rel axis cs:0.57979797979798,-0.0298507462686567,0) {GBVSNoCB};
\node[left, inner sep=0mm, rotate=90, text=black]
at (rel axis cs:0.62020202020202,-0.0298507462686567,0) {AIM};
\node[left, inner sep=0mm, rotate=90, text=black]
at (rel axis cs:0.662626262626263,-0.0298507462686567,0) {ITTI};
\node[left, inner sep=0mm, rotate=90, text=black]
at (rel axis cs:0.703030303030303,-0.0298507462686567,0) {GBVS};
\node[left, inner sep=0mm, rotate=90, text=black]
at (rel axis cs:0.745454545454545,-0.0298507462686567,0) {RandomCS};
\node[left, inner sep=0mm, rotate=90, text=black]
at (rel axis cs:0.787878787878788,-0.0298507462686567,0) {Judd};
\node[left, inner sep=0mm, rotate=90, text=black]
at (rel axis cs:0.828282828282828,-0.0298507462686567,0) {SUN};
\node[left, inner sep=0mm, rotate=90, text=black]
at (rel axis cs:0.870707070707071,-0.0298507462686567,0) {SCIA};
\node[left, inner sep=0mm, rotate=90, text=black]
at (rel axis cs:0.911111111111111,-0.0298507462686567,0) {BMS};
\node[left, inner sep=0mm, rotate=90, text=black]
at (rel axis cs:0.953535353535354,-0.0298507462686567,0) {CovSal};
\end{axis}
\end{tikzpicture}%
 & 
%
\begin{tikzpicture}

\begin{axis}[%
width=\figurewidth,
height=\figureheight,
clip=false,
scale only axis,
xmin=1,
xmax=25,
xtick={1,2,3,4,5,6,7,8,9,10,11,12,13,14,15,16,17,18,19,20,21,22,23,24},
xticklabels={\empty},
xmajorgrids,
ymin=3.14917181669125,
ymax=15.4644661247791,
ylabel={SSKLD},
ymajorgrids,
axis x line*=bottom,
axis y line*=left,
ytick = {4 ,6 ,8 ,10,12,14},ylabel style={font=\tiny},legend style={font=\tiny},
]
\addplot [color=red,solid,line width=1.0pt,mark=asterisk,mark options={solid},forget plot]
  table[row sep=crcr]{1	4.45046587515662\\
2	5.41800554989892\\
3	5.61403618906751\\
4	5.9465008232201\\
5	6.73787830062365\\
6	6.90834526771004\\
7	6.70059359094521\\
8	5.7086350686218\\
9	6.62295164968068\\
10	7.28389474694011\\
11	6.34618444467061\\
12	6.71382685661079\\
13	6.7800888278071\\
14	7.51639328668164\\
15	7.63201967658074\\
16	6.92169759563266\\
17	7.80315030845856\\
18	7.51752092374137\\
19	6.87509727858992\\
20	7.76715369252515\\
21	7.39047315946438\\
22	8.22621263224905\\
23	8.76075839519248\\
24	9.65462473671774\\
};
\addplot [color=green,solid,line width=1.0pt,mark=+,mark options={solid},forget plot]
  table[row sep=crcr]{1	4.16990101494347\\
2	5.52162307798907\\
3	5.79748186163534\\
4	6.42174381181353\\
5	6.66958385876378\\
6	6.43304294481032\\
7	7.35341950835909\\
8	7.40907568795733\\
9	7.07984717967299\\
10	6.34628784864444\\
11	7.5531841154503\\
12	7.54119984155741\\
13	7.5660123138182\\
14	7.57497977162134\\
15	7.24317047058987\\
16	7.75791325273666\\
17	7.47507552050446\\
18	7.82869657182323\\
19	8.32572167656973\\
20	8.59272054972701\\
21	8.32457077234067\\
22	9.06387058856472\\
23	9.22735824810525\\
24	8.75202989248567\\
};
\addplot [color=blue,solid,line width=1.0pt,mark=o,mark options={solid},forget plot]
  table[row sep=crcr]{1	3.89931543304502\\
2	4.20559921879938\\
3	5.03142332742173\\
4	4.74715515105656\\
5	4.09450742522739\\
6	4.73294991906114\\
7	4.5285589921669\\
8	5.49363525150049\\
9	5.44678111108673\\
10	5.52823535029632\\
11	5.39210086448994\\
12	5.10527340187054\\
13	5.46782111396335\\
14	5.24172514905203\\
15	5.5643798034062\\
16	5.85907508371358\\
17	5.29161685229688\\
18	5.24721161008136\\
19	6.55317027155774\\
20	6.20291910857448\\
21	6.99150813738699\\
22	6.68164580093575\\
23	6.75465136116161\\
24	8.04063248828796\\
};
\addplot [color=black,dashed,line width=1.0pt,forget plot]
  table[row sep=crcr]{1	4.17322744104837\\
2	5.04840928222912\\
3	5.48098045937486\\
4	5.70513326203006\\
5	5.83398986153827\\
6	6.02477937719383\\
7	6.19419069715706\\
8	6.20378200269321\\
9	6.38319331348013\\
10	6.38613931529362\\
11	6.43048980820361\\
12	6.45343336667958\\
13	6.60464075186289\\
14	6.77769940245167\\
15	6.8131899835256\\
16	6.84622864402763\\
17	6.85661422708663\\
18	6.86447636854865\\
19	7.25132974223913\\
20	7.52093111694221\\
21	7.56885068973068\\
22	7.99057634058318\\
23	8.24758933481978\\
24	8.81576237249712\\
};
\node[left, inner sep=0mm, rotate=90, text=black]
at (rel axis cs:-0.00202020202020202,-0.0298507462686567,0) {ITTI2};
\node[left, inner sep=0mm, rotate=90, text=black]
at (rel axis cs:0.0383838383838384,-0.0298507462686567,0) {SDSRG};
\node[left, inner sep=0mm, rotate=90, text=black]
at (rel axis cs:0.0808080808080808,-0.0298507462686567,0) {FTS};
\node[left, inner sep=0mm, rotate=90, text=black]
at (rel axis cs:0.121212121212121,-0.0298507462686567,0) {SigSal};
\node[left, inner sep=0mm, rotate=90, text=black]
at (rel axis cs:0.163636363636364,-0.0298507462686567,0) {HouNIPS};
\node[left, inner sep=0mm, rotate=90, text=black]
at (rel axis cs:0.204040404040404,-0.0298507462686567,0) {GR};
\node[left, inner sep=0mm, rotate=90, text=black]
at (rel axis cs:0.246464646464646,-0.0298507462686567,0) {AIM};
\node[left, inner sep=0mm, rotate=90, text=black]
at (rel axis cs:0.288888888888889,-0.0298507462686567,0) {Context};
\node[left, inner sep=0mm, rotate=90, text=black]
at (rel axis cs:0.329292929292929,-0.0298507462686567,0) {SDSRL};
\node[left, inner sep=0mm, rotate=90, text=black]
at (rel axis cs:0.371717171717172,-0.0298507462686567,0) {ITTI};
\node[left, inner sep=0mm, rotate=90, text=black]
at (rel axis cs:0.412121212121212,-0.0298507462686567,0) {SIMFine};
\node[left, inner sep=0mm, rotate=90, text=black]
at (rel axis cs:0.454545454545455,-0.0298507462686567,0) {SIMCoarse};
\node[left, inner sep=0mm, rotate=90, text=black]
at (rel axis cs:0.496969696969697,-0.0298507462686567,0) {SpectRes};
\node[left, inner sep=0mm, rotate=90, text=black]
at (rel axis cs:0.537373737373737,-0.0298507462686567,0) {AWS};
\node[left, inner sep=0mm, rotate=90, text=black]
at (rel axis cs:0.57979797979798,-0.0298507462686567,0) {GAFFE};
\node[left, inner sep=0mm, rotate=90, text=black]
at (rel axis cs:0.62020202020202,-0.0298507462686567,0) {Torralba};
\node[left, inner sep=0mm, rotate=90, text=black]
at (rel axis cs:0.662626262626263,-0.0298507462686567,0) {GBVS};
\node[left, inner sep=0mm, rotate=90, text=black]
at (rel axis cs:0.703030303030303,-0.0298507462686567,0) {GBVSNoCB};
\node[left, inner sep=0mm, rotate=90, text=black]
at (rel axis cs:0.745454545454545,-0.0298507462686567,0) {Judd};
\node[left, inner sep=0mm, rotate=90, text=black]
at (rel axis cs:0.787878787878788,-0.0298507462686567,0) {SUN};
\node[left, inner sep=0mm, rotate=90, text=black]
at (rel axis cs:0.828282828282828,-0.0298507462686567,0) {RandomCS};
\node[left, inner sep=0mm, rotate=90, text=black]
at (rel axis cs:0.870707070707071,-0.0298507462686567,0) {SCIA};
\node[left, inner sep=0mm, rotate=90, text=black]
at (rel axis cs:0.911111111111111,-0.0298507462686567,0) {BMS};
\node[left, inner sep=0mm, rotate=90, text=black]
at (rel axis cs:0.953535353535354,-0.0298507462686567,0) {CovSal};
\end{axis}
\end{tikzpicture}%
 & 
%
\begin{tikzpicture}

\begin{axis}[%
width=\figurewidth,
height=\figureheight,
clip=false,
scale only axis,
xmin=1,
xmax=25,
xtick={1,2,3,4,5,6,7,8,9,10,11,12,13,14,15,16,17,18,19,20,21,22,23,24},
xticklabels={\empty},
xmajorgrids,
ymin=3.14917181669125,
ymax=15.4644661247791,
ylabel={SSKLD},
ymajorgrids,
axis x line*=bottom,
axis y line*=left,
ytick = {4 ,6 ,8 ,10,12,14},ylabel style={font=\tiny},legend style={font=\tiny},
]
\addplot [color=red,solid,line width=1.0pt,mark=asterisk,mark options={solid},forget plot]
  table[row sep=crcr]{1	5.06882281543917\\
2	7.71188244651898\\
3	7.86917898569702\\
4	7.67567780370538\\
5	8.68242333760368\\
6	9.61125113394657\\
7	9.08273741013145\\
8	9.400185379597\\
9	9.22037324954008\\
10	9.9190744739758\\
11	8.57978923823648\\
12	9.57806726002459\\
13	10.8525493334628\\
14	10.2107466988349\\
15	9.85882661019403\\
16	9.58264764321294\\
17	10.28072540772\\
18	11.3634260928203\\
19	10.3490643038738\\
20	11.0344698850391\\
21	11.6729557473902\\
22	11.7981117352754\\
23	12.4225994854107\\
24	12.0073513479353\\
};
\addplot [color=green,solid,line width=1.0pt,mark=+,mark options={solid},forget plot]
  table[row sep=crcr]{1	4.58743811997194\\
2	6.08634281295712\\
3	6.21064588229137\\
4	6.6148189520782\\
5	7.09028354371498\\
6	7.09662226326991\\
7	7.66380091078786\\
8	7.52875812854082\\
9	7.83931454792842\\
10	6.95514746477236\\
11	8.3907761652816\\
12	8.16775867176094\\
13	6.72533954690172\\
14	7.6908103401027\\
15	9.12594209477694\\
16	8.09749385046661\\
17	8.01473452917235\\
18	8.65653415911047\\
19	8.53108945433372\\
20	8.71385474360557\\
21	8.6271060858365\\
22	9.37646934138211\\
23	9.67784205473546\\
24	9.78281012883217\\
};
\addplot [color=blue,solid,line width=1.0pt,mark=o,mark options={solid},forget plot]
  table[row sep=crcr]{1	4.22850764410635\\
2	4.14673358991132\\
3	4.72786356529003\\
4	4.52514487168081\\
5	4.92678936211123\\
6	5.14036343217485\\
7	5.18698220387888\\
8	5.07168341334836\\
9	5.01627333380038\\
10	5.29740672348104\\
11	5.36602301512185\\
12	5.29316871445616\\
13	5.61587638483083\\
14	5.45931463459754\\
15	4.69596392294226\\
16	6.18821357463797\\
17	6.03507416132723\\
18	4.81656154428001\\
19	6.61224335795029\\
20	7.11555734970527\\
21	7.16265381859079\\
22	6.30101421254137\\
23	6.78091182364127\\
24	8.73981006293262\\
};
\addplot [color=black,dashed,line width=1.0pt,forget plot]
  table[row sep=crcr]{1	4.62825619317249\\
2	5.98165294979581\\
3	6.26922947775947\\
4	6.27188054248813\\
5	6.8998320811433\\
6	7.28274560979711\\
7	7.31117350826606\\
8	7.33354230716206\\
9	7.35865371042296\\
10	7.39054288740974\\
11	7.44552947287998\\
12	7.67966488208056\\
13	7.73125508839844\\
14	7.78695722451173\\
15	7.89357754263774\\
16	7.95611835610584\\
17	8.11017803273985\\
18	8.27884059873693\\
19	8.49746570538592\\
20	8.95462732611665\\
21	9.15423855060582\\
22	9.15853176306629\\
23	9.62711778792914\\
24	10.1766571799\\
};
\node[left, inner sep=0mm, rotate=90, text=black]
at (rel axis cs:-0.00202020202020202,-0.0298507462686567,0) {ITTI2};
\node[left, inner sep=0mm, rotate=90, text=black]
at (rel axis cs:0.0383838383838384,-0.0298507462686567,0) {SDSRG};
\node[left, inner sep=0mm, rotate=90, text=black]
at (rel axis cs:0.0808080808080808,-0.0298507462686567,0) {FTS};
\node[left, inner sep=0mm, rotate=90, text=black]
at (rel axis cs:0.121212121212121,-0.0298507462686567,0) {HouNIPS};
\node[left, inner sep=0mm, rotate=90, text=black]
at (rel axis cs:0.163636363636364,-0.0298507462686567,0) {SigSal};
\node[left, inner sep=0mm, rotate=90, text=black]
at (rel axis cs:0.204040404040404,-0.0298507462686567,0) {SIMFine};
\node[left, inner sep=0mm, rotate=90, text=black]
at (rel axis cs:0.246464646464646,-0.0298507462686567,0) {GR};
\node[left, inner sep=0mm, rotate=90, text=black]
at (rel axis cs:0.288888888888889,-0.0298507462686567,0) {ITTI};
\node[left, inner sep=0mm, rotate=90, text=black]
at (rel axis cs:0.329292929292929,-0.0298507462686567,0) {SpectRes};
\node[left, inner sep=0mm, rotate=90, text=black]
at (rel axis cs:0.371717171717172,-0.0298507462686567,0) {SDSRL};
\node[left, inner sep=0mm, rotate=90, text=black]
at (rel axis cs:0.412121212121212,-0.0298507462686567,0) {GAFFE};
\node[left, inner sep=0mm, rotate=90, text=black]
at (rel axis cs:0.454545454545455,-0.0298507462686567,0) {GBVSNoCB};
\node[left, inner sep=0mm, rotate=90, text=black]
at (rel axis cs:0.496969696969697,-0.0298507462686567,0) {SIMCoarse};
\node[left, inner sep=0mm, rotate=90, text=black]
at (rel axis cs:0.537373737373737,-0.0298507462686567,0) {AWS};
\node[left, inner sep=0mm, rotate=90, text=black]
at (rel axis cs:0.57979797979798,-0.0298507462686567,0) {Context};
\node[left, inner sep=0mm, rotate=90, text=black]
at (rel axis cs:0.62020202020202,-0.0298507462686567,0) {AIM};
\node[left, inner sep=0mm, rotate=90, text=black]
at (rel axis cs:0.662626262626263,-0.0298507462686567,0) {SUN};
\node[left, inner sep=0mm, rotate=90, text=black]
at (rel axis cs:0.703030303030303,-0.0298507462686567,0) {GBVS};
\node[left, inner sep=0mm, rotate=90, text=black]
at (rel axis cs:0.745454545454545,-0.0298507462686567,0) {Torralba};
\node[left, inner sep=0mm, rotate=90, text=black]
at (rel axis cs:0.787878787878788,-0.0298507462686567,0) {Judd};
\node[left, inner sep=0mm, rotate=90, text=black]
at (rel axis cs:0.828282828282828,-0.0298507462686567,0) {SCIA};
\node[left, inner sep=0mm, rotate=90, text=black]
at (rel axis cs:0.870707070707071,-0.0298507462686567,0) {RandomCS};
\node[left, inner sep=0mm, rotate=90, text=black]
at (rel axis cs:0.911111111111111,-0.0298507462686567,0) {BMS};
\node[left, inner sep=0mm, rotate=90, text=black]
at (rel axis cs:0.953535353535354,-0.0298507462686567,0) {CovSal};
\end{axis}
\end{tikzpicture}%
 \\
%
\begin{tikzpicture}

\begin{axis}[%
width=\figurewidth,
height=\figureheight,
clip=false,
scale only axis,
xmin=1,
xmax=25,
xtick={1,2,3,4,5,6,7,8,9,10,11,12,13,14,15,16,17,18,19,20,21,22,23,24},
xticklabels={\empty},
xmajorgrids,
ymin=0.327075558329168,
ymax=0.927774250217951,
ylabel={SJSD},
ymajorgrids,
axis x line*=bottom,
axis y line*=left,
ytick = {0.4,0.5,0.6,0.7,0.8,0.9},ylabel style={font=\tiny},legend style={font=\tiny},
]
\addplot [color=red,solid,line width=1.0pt,mark=asterisk,mark options={solid},forget plot]
  table[row sep=crcr]{1	0.38371941572191\\
2	0.646352889147952\\
3	0.609978377350859\\
4	0.697236325488402\\
5	0.720069476018997\\
6	0.743518430479386\\
7	0.732650820966518\\
8	0.74582618951883\\
9	0.706837797764207\\
10	0.712614149612995\\
11	0.771054518428663\\
12	0.765571680111795\\
13	0.772287841591573\\
14	0.72084008964893\\
15	0.755671790984319\\
16	0.776138355380847\\
17	0.787539303335283\\
18	0.781431237862109\\
19	0.79322567118101\\
20	0.7988917953791\\
21	0.836494004205441\\
22	0.810730442934726\\
23	0.843431136561773\\
24	0.81245027017672\\
};
\addplot [color=green,solid,line width=1.0pt,mark=+,mark options={solid},forget plot]
  table[row sep=crcr]{1	0.379287372653925\\
2	0.544935331896225\\
3	0.57620014848099\\
4	0.652810294037094\\
5	0.657409108735807\\
6	0.676234978184702\\
7	0.678398176489439\\
8	0.680865483310899\\
9	0.711100430279514\\
10	0.715934096634149\\
11	0.697630226952496\\
12	0.705563747294337\\
13	0.713522371334441\\
14	0.736942498019066\\
15	0.717401851727192\\
16	0.718545278433246\\
17	0.723584203977837\\
18	0.735061068199546\\
19	0.732281060140378\\
20	0.750755562993101\\
21	0.746213000295509\\
22	0.7572920806113\\
23	0.743807540808213\\
24	0.785796322474511\\
};
\addplot [color=blue,solid,line width=1.0pt,mark=o,mark options={solid},forget plot]
  table[row sep=crcr]{1	0.368344239384292\\
2	0.503037899674071\\
3	0.533556252003352\\
4	0.603724330449766\\
5	0.626770600574126\\
6	0.636120816878292\\
7	0.652421191711768\\
8	0.641537941132242\\
9	0.676145189685812\\
10	0.679820136377702\\
11	0.658248155592428\\
12	0.657532886041336\\
13	0.648928009468492\\
14	0.705528527704486\\
15	0.695576798154882\\
16	0.697531974218986\\
17	0.686265235837965\\
18	0.695302564803785\\
19	0.691704240736122\\
20	0.705146590685282\\
21	0.690924215441407\\
22	0.716159777384161\\
23	0.712979771888552\\
24	0.751023503774109\\
};
\addplot [color=black,dashed,line width=1.0pt,forget plot]
  table[row sep=crcr]{1	0.377117009253376\\
2	0.564775373572749\\
3	0.573244925945067\\
4	0.651256983325087\\
5	0.66808306177631\\
6	0.685291408514127\\
7	0.687823396389242\\
8	0.689409871320657\\
9	0.698027805909844\\
10	0.702789460874949\\
11	0.708977633657862\\
12	0.709556104482489\\
13	0.711579407464836\\
14	0.721103705124161\\
15	0.722883480288798\\
16	0.730738536011026\\
17	0.732462914383695\\
18	0.737264956955146\\
19	0.73907032401917\\
20	0.751597983019161\\
21	0.757877073314119\\
22	0.761394100310062\\
23	0.766739483086179\\
24	0.78309003214178\\
};
\node[left, inner sep=0mm, rotate=90, text=black]
at (rel axis cs:-0.00202020202020202,-0.0298507462686567,0) {ITTI2};
\node[left, inner sep=0mm, rotate=90, text=black]
at (rel axis cs:0.0383838383838384,-0.0298507462686567,0) {GR};
\node[left, inner sep=0mm, rotate=90, text=black]
at (rel axis cs:0.0808080808080808,-0.0298507462686567,0) {FTS};
\node[left, inner sep=0mm, rotate=90, text=black]
at (rel axis cs:0.121212121212121,-0.0298507462686567,0) {HouNIPS};
\node[left, inner sep=0mm, rotate=90, text=black]
at (rel axis cs:0.163636363636364,-0.0298507462686567,0) {SigSal};
\node[left, inner sep=0mm, rotate=90, text=black]
at (rel axis cs:0.204040404040404,-0.0298507462686567,0) {BMS};
\node[left, inner sep=0mm, rotate=90, text=black]
at (rel axis cs:0.246464646464646,-0.0298507462686567,0) {SpectRes};
\node[left, inner sep=0mm, rotate=90, text=black]
at (rel axis cs:0.288888888888889,-0.0298507462686567,0) {SUN};
\node[left, inner sep=0mm, rotate=90, text=black]
at (rel axis cs:0.329292929292929,-0.0298507462686567,0) {SIMFine};
\node[left, inner sep=0mm, rotate=90, text=black]
at (rel axis cs:0.371717171717172,-0.0298507462686567,0) {SIMCoarse};
\node[left, inner sep=0mm, rotate=90, text=black]
at (rel axis cs:0.412121212121212,-0.0298507462686567,0) {Torralba};
\node[left, inner sep=0mm, rotate=90, text=black]
at (rel axis cs:0.454545454545455,-0.0298507462686567,0) {RandomCS};
\node[left, inner sep=0mm, rotate=90, text=black]
at (rel axis cs:0.496969696969697,-0.0298507462686567,0) {SCIA};
\node[left, inner sep=0mm, rotate=90, text=black]
at (rel axis cs:0.537373737373737,-0.0298507462686567,0) {SDSRG};
\node[left, inner sep=0mm, rotate=90, text=black]
at (rel axis cs:0.57979797979798,-0.0298507462686567,0) {Context};
\node[left, inner sep=0mm, rotate=90, text=black]
at (rel axis cs:0.62020202020202,-0.0298507462686567,0) {AWS};
\node[left, inner sep=0mm, rotate=90, text=black]
at (rel axis cs:0.662626262626263,-0.0298507462686567,0) {AIM};
\node[left, inner sep=0mm, rotate=90, text=black]
at (rel axis cs:0.703030303030303,-0.0298507462686567,0) {GBVSNoCB};
\node[left, inner sep=0mm, rotate=90, text=black]
at (rel axis cs:0.745454545454545,-0.0298507462686567,0) {GAFFE};
\node[left, inner sep=0mm, rotate=90, text=black]
at (rel axis cs:0.787878787878788,-0.0298507462686567,0) {GBVS};
\node[left, inner sep=0mm, rotate=90, text=black]
at (rel axis cs:0.828282828282828,-0.0298507462686567,0) {CovSal};
\node[left, inner sep=0mm, rotate=90, text=black]
at (rel axis cs:0.870707070707071,-0.0298507462686567,0) {Judd};
\node[left, inner sep=0mm, rotate=90, text=black]
at (rel axis cs:0.911111111111111,-0.0298507462686567,0) {ITTI};
\node[left, inner sep=0mm, rotate=90, text=black]
at (rel axis cs:0.953535353535354,-0.0298507462686567,0) {SDSRL};
\end{axis}
\end{tikzpicture}%
 & 
%
\begin{tikzpicture}

\begin{axis}[%
width=\figurewidth,
height=\figureheight,
clip=false,
scale only axis,
xmin=1,
xmax=25,
xtick={1,2,3,4,5,6,7,8,9,10,11,12,13,14,15,16,17,18,19,20,21,22,23,24},
xticklabels={\empty},
xmajorgrids,
ymin=0.327075558329168,
ymax=0.927774250217951,
ylabel={SJSD},
ymajorgrids,
axis x line*=bottom,
axis y line*=left,
ytick = {0.4,0.5,0.6,0.7,0.8,0.9},ylabel style={font=\tiny},legend style={font=\tiny},
]
\addplot [color=red,solid,line width=1.0pt,mark=asterisk,mark options={solid},forget plot]
  table[row sep=crcr]{1	0.38420369154246\\
2	0.563184314792809\\
3	0.567898837775131\\
4	0.653243871773243\\
5	0.709889070933157\\
6	0.665583503286201\\
7	0.679501349183836\\
8	0.7098447513346\\
9	0.711335388869275\\
10	0.674202389146699\\
11	0.698862018373251\\
12	0.699379287253973\\
13	0.703791904933037\\
14	0.692269992935232\\
15	0.720751252083766\\
16	0.732509043825598\\
17	0.736282957022221\\
18	0.743727639667251\\
19	0.733212258374312\\
20	0.731630464506051\\
21	0.744515179045534\\
22	0.744235364643956\\
23	0.753970243365222\\
24	0.73158134855485\\
};
\addplot [color=green,solid,line width=1.0pt,mark=+,mark options={solid},forget plot]
  table[row sep=crcr]{1	0.363417287032409\\
2	0.546865006685813\\
3	0.575100242704647\\
4	0.661497696744742\\
5	0.653793632864666\\
6	0.69065817117336\\
7	0.68562284278151\\
8	0.712909989322614\\
9	0.705585563399698\\
10	0.737120105171739\\
11	0.714638476007913\\
12	0.719355927730284\\
13	0.709391927319609\\
14	0.746736822220999\\
15	0.732184306799335\\
16	0.730250870639429\\
17	0.745563075689142\\
18	0.745423411155178\\
19	0.737344675547898\\
20	0.736858224674013\\
21	0.759930146226442\\
22	0.786824067199878\\
23	0.766093043887539\\
24	0.78593348720688\\
};
\addplot [color=blue,solid,line width=1.0pt,mark=o,mark options={solid},forget plot]
  table[row sep=crcr]{1	0.370725999385883\\
2	0.530019733975792\\
3	0.557161536533197\\
4	0.642247620464241\\
5	0.624127491801972\\
6	0.647515651168325\\
7	0.659934055277715\\
8	0.666593483436366\\
9	0.673456554020299\\
10	0.680855498928997\\
11	0.686734869407634\\
12	0.681716384694884\\
13	0.691487226239347\\
14	0.692403323467307\\
15	0.689999274726602\\
16	0.709122721272403\\
17	0.70043355564091\\
18	0.694162464369351\\
19	0.716837698115054\\
20	0.721610103925411\\
21	0.726721988529266\\
22	0.715264382597438\\
23	0.727075223101175\\
24	0.750549734148353\\
};
\addplot [color=black,dashed,line width=1.0pt,forget plot]
  table[row sep=crcr]{1	0.372782325986917\\
2	0.546689685151471\\
3	0.566720205670992\\
4	0.652329729660742\\
5	0.662603398533265\\
6	0.667919108542629\\
7	0.675019415747687\\
8	0.696449408031193\\
9	0.696792502096424\\
10	0.697392664415812\\
11	0.700078454596266\\
12	0.70015053322638\\
13	0.701557019497331\\
14	0.710470046207846\\
15	0.714311611203234\\
16	0.723960878579143\\
17	0.727426529450758\\
18	0.727771171730593\\
19	0.729131544012421\\
20	0.730032931035158\\
21	0.743722437933747\\
22	0.748774604813757\\
23	0.749046170117979\\
24	0.756021523303361\\
};
\node[left, inner sep=0mm, rotate=90, text=black]
at (rel axis cs:-0.00202020202020202,-0.0298507462686567,0) {ITTI2};
\node[left, inner sep=0mm, rotate=90, text=black]
at (rel axis cs:0.0383838383838384,-0.0298507462686567,0) {GR};
\node[left, inner sep=0mm, rotate=90, text=black]
at (rel axis cs:0.0808080808080808,-0.0298507462686567,0) {FTS};
\node[left, inner sep=0mm, rotate=90, text=black]
at (rel axis cs:0.121212121212121,-0.0298507462686567,0) {SigSal};
\node[left, inner sep=0mm, rotate=90, text=black]
at (rel axis cs:0.163636363636364,-0.0298507462686567,0) {HouNIPS};
\node[left, inner sep=0mm, rotate=90, text=black]
at (rel axis cs:0.204040404040404,-0.0298507462686567,0) {BMS};
\node[left, inner sep=0mm, rotate=90, text=black]
at (rel axis cs:0.246464646464646,-0.0298507462686567,0) {SpectRes};
\node[left, inner sep=0mm, rotate=90, text=black]
at (rel axis cs:0.288888888888889,-0.0298507462686567,0) {SUN};
\node[left, inner sep=0mm, rotate=90, text=black]
at (rel axis cs:0.329292929292929,-0.0298507462686567,0) {SCIA};
\node[left, inner sep=0mm, rotate=90, text=black]
at (rel axis cs:0.371717171717172,-0.0298507462686567,0) {SIMFine};
\node[left, inner sep=0mm, rotate=90, text=black]
at (rel axis cs:0.412121212121212,-0.0298507462686567,0) {RandomCS};
\node[left, inner sep=0mm, rotate=90, text=black]
at (rel axis cs:0.454545454545455,-0.0298507462686567,0) {AIM};
\node[left, inner sep=0mm, rotate=90, text=black]
at (rel axis cs:0.496969696969697,-0.0298507462686567,0) {Torralba};
\node[left, inner sep=0mm, rotate=90, text=black]
at (rel axis cs:0.537373737373737,-0.0298507462686567,0) {SIMCoarse};
\node[left, inner sep=0mm, rotate=90, text=black]
at (rel axis cs:0.57979797979798,-0.0298507462686567,0) {GAFFE};
\node[left, inner sep=0mm, rotate=90, text=black]
at (rel axis cs:0.62020202020202,-0.0298507462686567,0) {AWS};
\node[left, inner sep=0mm, rotate=90, text=black]
at (rel axis cs:0.662626262626263,-0.0298507462686567,0) {CovSal};
\node[left, inner sep=0mm, rotate=90, text=black]
at (rel axis cs:0.703030303030303,-0.0298507462686567,0) {Context};
\node[left, inner sep=0mm, rotate=90, text=black]
at (rel axis cs:0.745454545454545,-0.0298507462686567,0) {ITTI};
\node[left, inner sep=0mm, rotate=90, text=black]
at (rel axis cs:0.787878787878788,-0.0298507462686567,0) {GBVSNoCB};
\node[left, inner sep=0mm, rotate=90, text=black]
at (rel axis cs:0.828282828282828,-0.0298507462686567,0) {GBVS};
\node[left, inner sep=0mm, rotate=90, text=black]
at (rel axis cs:0.870707070707071,-0.0298507462686567,0) {SDSRG};
\node[left, inner sep=0mm, rotate=90, text=black]
at (rel axis cs:0.911111111111111,-0.0298507462686567,0) {Judd};
\node[left, inner sep=0mm, rotate=90, text=black]
at (rel axis cs:0.953535353535354,-0.0298507462686567,0) {SDSRL};
\end{axis}
\end{tikzpicture}%
 & 
%
\begin{tikzpicture}

\begin{axis}[%
width=\figurewidth,
height=\figureheight,
clip=false,
scale only axis,
xmin=1,
xmax=25,
xtick={1,2,3,4,5,6,7,8,9,10,11,12,13,14,15,16,17,18,19,20,21,22,23,24},
xticklabels={\empty},
xmajorgrids,
ymin=0.327075558329168,
ymax=0.927774250217951,
ylabel={SJSD},
ymajorgrids,
axis x line*=bottom,
axis y line*=left,
ytick = {0.4,0.5,0.6,0.7,0.8,0.9},ylabel style={font=\tiny},legend style={font=\tiny},
]
\addplot [color=red,solid,line width=1.0pt,mark=asterisk,mark options={solid},forget plot]
  table[row sep=crcr]{1	0.380219536197804\\
2	0.602590566072943\\
3	0.656481504412851\\
4	0.704794721106125\\
5	0.734487396639805\\
6	0.701756206311169\\
7	0.732621220285201\\
8	0.729071308946578\\
9	0.732058943281336\\
10	0.720676853685891\\
11	0.769042138817782\\
12	0.745424206462087\\
13	0.776321836809051\\
14	0.763415198256079\\
15	0.797459366314052\\
16	0.779892667184799\\
17	0.809495376784925\\
18	0.787318430106676\\
19	0.783060947387237\\
20	0.801389399079313\\
21	0.81495435310113\\
22	0.814202357050972\\
23	0.821947032176655\\
24	0.826535895873313\\
};
\addplot [color=green,solid,line width=1.0pt,mark=+,mark options={solid},forget plot]
  table[row sep=crcr]{1	0.391901942446711\\
2	0.556588559168544\\
3	0.615728653888183\\
4	0.663738532198325\\
5	0.657690016040184\\
6	0.684515260182109\\
7	0.702388975081449\\
8	0.716657630856712\\
9	0.721375431882016\\
10	0.733824829324206\\
11	0.726594349896555\\
12	0.738850847526972\\
13	0.735859332282166\\
14	0.74000257009105\\
15	0.745414201652307\\
16	0.751269406910014\\
17	0.750360729357597\\
18	0.753393777230981\\
19	0.758622374952886\\
20	0.775973970293895\\
21	0.772201496207014\\
22	0.779435923329838\\
23	0.787134201174659\\
24	0.788808546202355\\
};
\addplot [color=blue,solid,line width=1.0pt,mark=o,mark options={solid},forget plot]
  table[row sep=crcr]{1	0.379946505694825\\
2	0.513411535281212\\
3	0.568781357201747\\
4	0.612754071752852\\
5	0.623975226116284\\
6	0.637118577229285\\
7	0.649618368762094\\
8	0.657218344409308\\
9	0.686028005676843\\
10	0.695305542326868\\
11	0.670741533115616\\
12	0.693428722644811\\
13	0.686305173323854\\
14	0.703614087936362\\
15	0.697479951539479\\
16	0.709292179964004\\
17	0.686691385536574\\
18	0.715559322305902\\
19	0.721224141111657\\
20	0.699033658381788\\
21	0.726578229697386\\
22	0.721650784128561\\
23	0.745510751358581\\
24	0.757838750732443\\
};
\addplot [color=black,dashed,line width=1.0pt,forget plot]
  table[row sep=crcr]{1	0.384022661446447\\
2	0.557530220174233\\
3	0.613663838500927\\
4	0.660429108352434\\
5	0.672050879598758\\
6	0.674463347907521\\
7	0.694876188042915\\
8	0.700982428070866\\
9	0.713154126946732\\
10	0.716602408445655\\
11	0.722126007276651\\
12	0.725901258877957\\
13	0.732828780805024\\
14	0.73567728542783\\
15	0.746784506501946\\
16	0.746818084686272\\
17	0.748849163893032\\
18	0.752090509881186\\
19	0.75430248781726\\
20	0.758799009251665\\
21	0.771244693001843\\
22	0.771763021503124\\
23	0.784863994903298\\
24	0.79106106426937\\
};
\node[left, inner sep=0mm, rotate=90, text=black]
at (rel axis cs:-0.00202020202020202,-0.0298507462686567,0) {ITTI2};
\node[left, inner sep=0mm, rotate=90, text=black]
at (rel axis cs:0.0383838383838384,-0.0298507462686567,0) {GR};
\node[left, inner sep=0mm, rotate=90, text=black]
at (rel axis cs:0.0808080808080808,-0.0298507462686567,0) {FTS};
\node[left, inner sep=0mm, rotate=90, text=black]
at (rel axis cs:0.121212121212121,-0.0298507462686567,0) {HouNIPS};
\node[left, inner sep=0mm, rotate=90, text=black]
at (rel axis cs:0.163636363636364,-0.0298507462686567,0) {SUN};
\node[left, inner sep=0mm, rotate=90, text=black]
at (rel axis cs:0.204040404040404,-0.0298507462686567,0) {SigSal};
\node[left, inner sep=0mm, rotate=90, text=black]
at (rel axis cs:0.246464646464646,-0.0298507462686567,0) {SpectRes};
\node[left, inner sep=0mm, rotate=90, text=black]
at (rel axis cs:0.288888888888889,-0.0298507462686567,0) {BMS};
\node[left, inner sep=0mm, rotate=90, text=black]
at (rel axis cs:0.329292929292929,-0.0298507462686567,0) {Torralba};
\node[left, inner sep=0mm, rotate=90, text=black]
at (rel axis cs:0.371717171717172,-0.0298507462686567,0) {AIM};
\node[left, inner sep=0mm, rotate=90, text=black]
at (rel axis cs:0.412121212121212,-0.0298507462686567,0) {RandomCS};
\node[left, inner sep=0mm, rotate=90, text=black]
at (rel axis cs:0.454545454545455,-0.0298507462686567,0) {GAFFE};
\node[left, inner sep=0mm, rotate=90, text=black]
at (rel axis cs:0.496969696969697,-0.0298507462686567,0) {SIMFine};
\node[left, inner sep=0mm, rotate=90, text=black]
at (rel axis cs:0.537373737373737,-0.0298507462686567,0) {SDSRG};
\node[left, inner sep=0mm, rotate=90, text=black]
at (rel axis cs:0.57979797979798,-0.0298507462686567,0) {SIMCoarse};
\node[left, inner sep=0mm, rotate=90, text=black]
at (rel axis cs:0.62020202020202,-0.0298507462686567,0) {AWS};
\node[left, inner sep=0mm, rotate=90, text=black]
at (rel axis cs:0.662626262626263,-0.0298507462686567,0) {SCIA};
\node[left, inner sep=0mm, rotate=90, text=black]
at (rel axis cs:0.703030303030303,-0.0298507462686567,0) {GBVSNoCB};
\node[left, inner sep=0mm, rotate=90, text=black]
at (rel axis cs:0.745454545454545,-0.0298507462686567,0) {ITTI};
\node[left, inner sep=0mm, rotate=90, text=black]
at (rel axis cs:0.787878787878788,-0.0298507462686567,0) {Context};
\node[left, inner sep=0mm, rotate=90, text=black]
at (rel axis cs:0.828282828282828,-0.0298507462686567,0) {GBVS};
\node[left, inner sep=0mm, rotate=90, text=black]
at (rel axis cs:0.870707070707071,-0.0298507462686567,0) {CovSal};
\node[left, inner sep=0mm, rotate=90, text=black]
at (rel axis cs:0.911111111111111,-0.0298507462686567,0) {Judd};
\node[left, inner sep=0mm, rotate=90, text=black]
at (rel axis cs:0.953535353535354,-0.0298507462686567,0) {SDSRL};
\end{axis}
\end{tikzpicture}%
\\
%
\begin{tikzpicture}

\begin{axis}[%
width=\figurewidth,
height=\figureheight,
clip=false,
scale only axis,
xmin=1,
xmax=25,
xtick={1,2,3,4,5,6,7,8,9,10,11,12,13,14,15,16,17,18,19,20,21,22,23,24},
xticklabels={\empty},
xmajorgrids,
ymin=0.351610051490797,
ymax=2.00081181494875,
ylabel={SEMD},
ymajorgrids,
axis x line*=bottom,
axis y line*=left,
ytick = {0.4,0.6,0.8,1  ,1.2,1.4,1.6,1.8,2  },ylabel style={font=\tiny},legend style={font=\tiny},
]
\addplot [color=red,solid,line width=1.0pt,mark=asterisk,mark options={solid},forget plot]
  table[row sep=crcr]{1	0.477809283958785\\
2	1.12913313588038\\
3	1.28741262393294\\
4	1.37559171476285\\
5	1.45201585288249\\
6	1.44205737865799\\
7	1.50056449132165\\
8	1.50330683037169\\
9	1.40495484256232\\
10	1.5182417166574\\
11	1.43781478158049\\
12	1.58718754435316\\
13	1.57729580626849\\
14	1.55104588852293\\
15	1.58948892613483\\
16	1.46026909989774\\
17	1.59722283031228\\
18	1.66701064333587\\
19	1.64025186042163\\
20	1.65424080913704\\
21	1.68951833492352\\
22	1.80912298627958\\
23	1.81891983177159\\
24	1.69809379513986\\
};
\addplot [color=green,solid,line width=1.0pt,mark=+,mark options={solid},forget plot]
  table[row sep=crcr]{1	0.49533419335825\\
2	1.0419318053821\\
3	1.00855867338801\\
4	1.26752701745955\\
5	1.28832313305951\\
6	1.28511786321903\\
7	1.28303351383983\\
8	1.33252104823345\\
9	1.38887945104556\\
10	1.35697050862891\\
11	1.41398479973358\\
12	1.38464749033068\\
13	1.38833850552553\\
14	1.42874461419524\\
15	1.39512761416508\\
16	1.50272781946087\\
17	1.40851511093605\\
18	1.45207551343127\\
19	1.45186338585882\\
20	1.46982808157648\\
21	1.4746616047784\\
22	1.46636255667636\\
23	1.52308398915727\\
24	1.57487356218668\\
};
\addplot [color=blue,solid,line width=1.0pt,mark=o,mark options={solid},forget plot]
  table[row sep=crcr]{1	0.42393090308795\\
2	0.9258227025865\\
3	0.92832458180279\\
4	1.10970329422296\\
5	1.16269532928602\\
6	1.22177723746007\\
7	1.17176007255998\\
8	1.20579471251419\\
9	1.28857547712511\\
10	1.211212687127\\
11	1.31322809060758\\
12	1.26141221214107\\
13	1.28265108111493\\
14	1.29536676328721\\
15	1.29235590298693\\
16	1.34448958286888\\
17	1.31698075465085\\
18	1.24443629206778\\
19	1.28628171857383\\
20	1.34350172035207\\
21	1.31052339193409\\
22	1.32441491514479\\
23	1.3552636566521\\
24	1.43578209781326\\
};
\addplot [color=black,dashed,line width=1.0pt,forget plot]
  table[row sep=crcr]{1	0.465691460134995\\
2	1.03229588128299\\
3	1.07476529304125\\
4	1.25094067548179\\
5	1.30101143840934\\
6	1.31631749311236\\
7	1.31845269257382\\
8	1.34720753037311\\
9	1.360803256911\\
10	1.3621416374711\\
11	1.38834255730722\\
12	1.41108241560831\\
13	1.41609513096965\\
14	1.42505242200179\\
15	1.42565748109561\\
16	1.43582883407583\\
17	1.44090623196639\\
18	1.45450748294497\\
19	1.45946565495143\\
20	1.48919020368853\\
21	1.491567777212\\
22	1.53330015270024\\
23	1.56575582586032\\
24	1.56958315171327\\
};
\node[left, inner sep=0mm, rotate=90, text=black]
at (rel axis cs:-0.00202020202020202,-0.0298507462686567,0) {ITTI2};
\node[left, inner sep=0mm, rotate=90, text=black]
at (rel axis cs:0.0383838383838384,-0.0298507462686567,0) {FTS};
\node[left, inner sep=0mm, rotate=90, text=black]
at (rel axis cs:0.0808080808080808,-0.0298507462686567,0) {GR};
\node[left, inner sep=0mm, rotate=90, text=black]
at (rel axis cs:0.121212121212121,-0.0298507462686567,0) {HouNIPS};
\node[left, inner sep=0mm, rotate=90, text=black]
at (rel axis cs:0.163636363636364,-0.0298507462686567,0) {SigSal};
\node[left, inner sep=0mm, rotate=90, text=black]
at (rel axis cs:0.204040404040404,-0.0298507462686567,0) {SpectRes};
\node[left, inner sep=0mm, rotate=90, text=black]
at (rel axis cs:0.246464646464646,-0.0298507462686567,0) {SUN};
\node[left, inner sep=0mm, rotate=90, text=black]
at (rel axis cs:0.288888888888889,-0.0298507462686567,0) {Torralba};
\node[left, inner sep=0mm, rotate=90, text=black]
at (rel axis cs:0.329292929292929,-0.0298507462686567,0) {SIMFine};
\node[left, inner sep=0mm, rotate=90, text=black]
at (rel axis cs:0.371717171717172,-0.0298507462686567,0) {RandomCS};
\node[left, inner sep=0mm, rotate=90, text=black]
at (rel axis cs:0.412121212121212,-0.0298507462686567,0) {SIMCoarse};
\node[left, inner sep=0mm, rotate=90, text=black]
at (rel axis cs:0.454545454545455,-0.0298507462686567,0) {BMS};
\node[left, inner sep=0mm, rotate=90, text=black]
at (rel axis cs:0.496969696969697,-0.0298507462686567,0) {AIM};
\node[left, inner sep=0mm, rotate=90, text=black]
at (rel axis cs:0.537373737373737,-0.0298507462686567,0) {Context};
\node[left, inner sep=0mm, rotate=90, text=black]
at (rel axis cs:0.57979797979798,-0.0298507462686567,0) {GAFFE};
\node[left, inner sep=0mm, rotate=90, text=black]
at (rel axis cs:0.62020202020202,-0.0298507462686567,0) {SDSRG};
\node[left, inner sep=0mm, rotate=90, text=black]
at (rel axis cs:0.662626262626263,-0.0298507462686567,0) {AWS};
\node[left, inner sep=0mm, rotate=90, text=black]
at (rel axis cs:0.703030303030303,-0.0298507462686567,0) {SCIA};
\node[left, inner sep=0mm, rotate=90, text=black]
at (rel axis cs:0.745454545454545,-0.0298507462686567,0) {GBVSNoCB};
\node[left, inner sep=0mm, rotate=90, text=black]
at (rel axis cs:0.787878787878788,-0.0298507462686567,0) {Judd};
\node[left, inner sep=0mm, rotate=90, text=black]
at (rel axis cs:0.828282828282828,-0.0298507462686567,0) {GBVS};
\node[left, inner sep=0mm, rotate=90, text=black]
at (rel axis cs:0.870707070707071,-0.0298507462686567,0) {ITTI};
\node[left, inner sep=0mm, rotate=90, text=black]
at (rel axis cs:0.911111111111111,-0.0298507462686567,0) {CovSal};
\node[left, inner sep=0mm, rotate=90, text=black]
at (rel axis cs:0.953535353535354,-0.0298507462686567,0) {SDSRL};
\end{axis}
\end{tikzpicture}%
 & 
%
\begin{tikzpicture}

\begin{axis}[%
width=\figurewidth,
height=\figureheight,
clip=false,
scale only axis,
xmin=1,
xmax=25,
xtick={1,2,3,4,5,6,7,8,9,10,11,12,13,14,15,16,17,18,19,20,21,22,23,24},
xticklabels={\empty},
xmajorgrids,
ymin=0.351610051490797,
ymax=2.00081181494875,
ylabel={SEMD},
ymajorgrids,
axis x line*=bottom,
axis y line*=left,
ytick = {0.4,0.6,0.8,1  ,1.2,1.4,1.6,1.8,2  },ylabel style={font=\tiny},legend style={font=\tiny},
]
\addplot [color=red,solid,line width=1.0pt,mark=asterisk,mark options={solid},forget plot]
  table[row sep=crcr]{1	0.46604758922067\\
2	1.04567210794111\\
3	1.09112264197544\\
4	1.22600170656024\\
5	1.36964457758663\\
6	1.27080467381166\\
7	1.32126969928453\\
8	1.31752071741813\\
9	1.36849814642008\\
10	1.34115570118713\\
11	1.35519459740074\\
12	1.37909990989385\\
13	1.32033145827114\\
14	1.42765925678367\\
15	1.42409867827255\\
16	1.34704707526869\\
17	1.45062180552785\\
18	1.40142486135344\\
19	1.45035936527297\\
20	1.45179492778475\\
21	1.46586803793374\\
22	1.45223986360197\\
23	1.45248165906892\\
24	1.39731879140171\\
};
\addplot [color=green,solid,line width=1.0pt,mark=+,mark options={solid},forget plot]
  table[row sep=crcr]{1	0.390677834989775\\
2	1.05015004059495\\
3	1.0190899633065\\
4	1.24249962017083\\
5	1.24802791742145\\
6	1.30433283089695\\
7	1.35565651520591\\
8	1.35267333878074\\
9	1.37363341958559\\
10	1.37024941638849\\
11	1.39712501143926\\
12	1.40528876666987\\
13	1.45066146294304\\
14	1.39216328456738\\
15	1.41445263306511\\
16	1.49613086358259\\
17	1.46717839704079\\
18	1.44024204309283\\
19	1.42381498284959\\
20	1.47825617451005\\
21	1.47177987719978\\
22	1.49367309319794\\
23	1.56180996500241\\
24	1.54598177358979\\
};
\addplot [color=blue,solid,line width=1.0pt,mark=o,mark options={solid},forget plot]
  table[row sep=crcr]{1	0.442484891907085\\
2	0.97906841437426\\
3	0.992082533256445\\
4	1.20293058706792\\
5	1.15477767545722\\
6	1.21037820281175\\
7	1.26691372440079\\
8	1.29516644144131\\
9	1.24969918752921\\
10	1.29537510659122\\
11	1.29659817411952\\
12	1.29131611924801\\
13	1.313646506749\\
14	1.32155479604893\\
15	1.34153786662239\\
16	1.34028032927246\\
17	1.30162542242469\\
18	1.3784046277985\\
19	1.37652582896196\\
20	1.35767814642718\\
21	1.3751114684467\\
22	1.40638301066167\\
23	1.38541264180916\\
24	1.46146694845461\\
};
\addplot [color=black,dashed,line width=1.0pt,forget plot]
  table[row sep=crcr]{1	0.43307010537251\\
2	1.02496352097011\\
3	1.03409837951279\\
4	1.22381063793299\\
5	1.2574833901551\\
6	1.26183856917345\\
7	1.31461331296374\\
8	1.32178683254673\\
9	1.33061025117829\\
10	1.33559340805561\\
11	1.34963926098651\\
12	1.35856826527057\\
13	1.36154647598773\\
14	1.38045911246666\\
15	1.39336305932002\\
16	1.39448608937458\\
17	1.40647520833111\\
18	1.40669051074825\\
19	1.41690005902817\\
20	1.42924308290733\\
21	1.43758646119341\\
22	1.45076532248719\\
23	1.46656808862683\\
24	1.46825583781537\\
};
\node[left, inner sep=0mm, rotate=90, text=black]
at (rel axis cs:-0.00202020202020202,-0.0298507462686567,0) {ITTI2};
\node[left, inner sep=0mm, rotate=90, text=black]
at (rel axis cs:0.0383838383838384,-0.0298507462686567,0) {FTS};
\node[left, inner sep=0mm, rotate=90, text=black]
at (rel axis cs:0.0808080808080808,-0.0298507462686567,0) {GR};
\node[left, inner sep=0mm, rotate=90, text=black]
at (rel axis cs:0.121212121212121,-0.0298507462686567,0) {SigSal};
\node[left, inner sep=0mm, rotate=90, text=black]
at (rel axis cs:0.163636363636364,-0.0298507462686567,0) {HouNIPS};
\node[left, inner sep=0mm, rotate=90, text=black]
at (rel axis cs:0.204040404040404,-0.0298507462686567,0) {SpectRes};
\node[left, inner sep=0mm, rotate=90, text=black]
at (rel axis cs:0.246464646464646,-0.0298507462686567,0) {AIM};
\node[left, inner sep=0mm, rotate=90, text=black]
at (rel axis cs:0.288888888888889,-0.0298507462686567,0) {RandomCS};
\node[left, inner sep=0mm, rotate=90, text=black]
at (rel axis cs:0.329292929292929,-0.0298507462686567,0) {SUN};
\node[left, inner sep=0mm, rotate=90, text=black]
at (rel axis cs:0.371717171717172,-0.0298507462686567,0) {Torralba};
\node[left, inner sep=0mm, rotate=90, text=black]
at (rel axis cs:0.412121212121212,-0.0298507462686567,0) {BMS};
\node[left, inner sep=0mm, rotate=90, text=black]
at (rel axis cs:0.454545454545455,-0.0298507462686567,0) {GAFFE};
\node[left, inner sep=0mm, rotate=90, text=black]
at (rel axis cs:0.496969696969697,-0.0298507462686567,0) {SIMFine};
\node[left, inner sep=0mm, rotate=90, text=black]
at (rel axis cs:0.537373737373737,-0.0298507462686567,0) {SCIA};
\node[left, inner sep=0mm, rotate=90, text=black]
at (rel axis cs:0.57979797979798,-0.0298507462686567,0) {AWS};
\node[left, inner sep=0mm, rotate=90, text=black]
at (rel axis cs:0.62020202020202,-0.0298507462686567,0) {SIMCoarse};
\node[left, inner sep=0mm, rotate=90, text=black]
at (rel axis cs:0.662626262626263,-0.0298507462686567,0) {Context};
\node[left, inner sep=0mm, rotate=90, text=black]
at (rel axis cs:0.703030303030303,-0.0298507462686567,0) {GBVSNoCB};
\node[left, inner sep=0mm, rotate=90, text=black]
at (rel axis cs:0.745454545454545,-0.0298507462686567,0) {ITTI};
\node[left, inner sep=0mm, rotate=90, text=black]
at (rel axis cs:0.787878787878788,-0.0298507462686567,0) {Judd};
\node[left, inner sep=0mm, rotate=90, text=black]
at (rel axis cs:0.828282828282828,-0.0298507462686567,0) {CovSal};
\node[left, inner sep=0mm, rotate=90, text=black]
at (rel axis cs:0.870707070707071,-0.0298507462686567,0) {GBVS};
\node[left, inner sep=0mm, rotate=90, text=black]
at (rel axis cs:0.911111111111111,-0.0298507462686567,0) {SDSRG};
\node[left, inner sep=0mm, rotate=90, text=black]
at (rel axis cs:0.953535353535354,-0.0298507462686567,0) {SDSRL};
\end{axis}
\end{tikzpicture}%
 & 
%
\begin{tikzpicture}

\begin{axis}[%
width=\figurewidth,
height=\figureheight,
clip=false,
scale only axis,
xmin=1,
xmax=25,
xtick={1,2,3,4,5,6,7,8,9,10,11,12,13,14,15,16,17,18,19,20,21,22,23,24},
xticklabels={\empty},
xmajorgrids,
ymin=0.351610051490797,
ymax=2.00081181494875,
ylabel={SEMD},
ymajorgrids,
axis x line*=bottom,
axis y line*=left,
ytick = {0.4,0.6,0.8,1  ,1.2,1.4,1.6,1.8,2  },ylabel style={font=\tiny},legend style={font=\tiny},
]
\addplot [color=red,solid,line width=1.0pt,mark=asterisk,mark options={solid},forget plot]
  table[row sep=crcr]{1	0.45952172873043\\
2	1.05907209030147\\
3	1.19695450071441\\
4	1.34445116000635\\
5	1.34937522970861\\
6	1.37623292549822\\
7	1.43320524520057\\
8	1.41228228764744\\
9	1.37068097627701\\
10	1.41600772939867\\
11	1.52408281980772\\
12	1.53163464977852\\
13	1.58120964098568\\
14	1.53773508401739\\
15	1.58221093780103\\
16	1.54980423424638\\
17	1.60220181038085\\
18	1.63072655505587\\
19	1.64901074186189\\
20	1.72029477252722\\
21	1.64178482588506\\
22	1.69432902836269\\
23	1.69540799203473\\
24	1.6781031246248\\
};
\addplot [color=green,solid,line width=1.0pt,mark=+,mark options={solid},forget plot]
  table[row sep=crcr]{1	0.503381250043495\\
2	1.03831827293268\\
3	1.0878950844009\\
4	1.19164478451041\\
5	1.29443245380962\\
6	1.29743956737672\\
7	1.32924995141587\\
8	1.35487173859096\\
9	1.40204265403644\\
10	1.42909403902696\\
11	1.39087871840996\\
12	1.4678755585889\\
13	1.44675802039094\\
14	1.48253161387459\\
15	1.47801283951377\\
16	1.48924940392638\\
17	1.46713653043918\\
18	1.47585463664667\\
19	1.57597018691977\\
20	1.52706221468357\\
21	1.53294460940435\\
22	1.51965251224881\\
23	1.59809591117894\\
24	1.58092536977461\\
};
\addplot [color=blue,solid,line width=1.0pt,mark=o,mark options={solid},forget plot]
  table[row sep=crcr]{1	0.454373018866845\\
2	0.9763784512965\\
3	1.00195592520643\\
4	1.13199317657729\\
5	1.13178856055258\\
6	1.18078458010046\\
7	1.18046669188066\\
8	1.26544794846413\\
9	1.29730301602189\\
10	1.29263300963755\\
11	1.26688211219434\\
12	1.33500252816723\\
13	1.31039618960197\\
14	1.32313474360103\\
15	1.35127365581173\\
16	1.37398062077424\\
17	1.35357242139409\\
18	1.34852728068528\\
19	1.3114836683076\\
20	1.30943943016711\\
21	1.39809274916595\\
22	1.39187909435849\\
23	1.43002662009366\\
24	1.47235740806178\\
};
\addplot [color=black,dashed,line width=1.0pt,forget plot]
  table[row sep=crcr]{1	0.472425332546923\\
2	1.02458960484355\\
3	1.09560183677391\\
4	1.22269637369802\\
5	1.25853208135694\\
6	1.28481902432513\\
7	1.3143072961657\\
8	1.34420065823418\\
9	1.35667554877845\\
10	1.37924492602106\\
11	1.39394788347067\\
12	1.44483757884488\\
13	1.44612128365953\\
14	1.44780048049767\\
15	1.47049914437551\\
16	1.471011419649\\
17	1.47430358740471\\
18	1.48503615746261\\
19	1.51215486569642\\
20	1.51893213912597\\
21	1.52427406148512\\
22	1.53528687832333\\
23	1.57451017443578\\
24	1.57712863415373\\
};
\node[left, inner sep=0mm, rotate=90, text=black]
at (rel axis cs:-0.00202020202020202,-0.0298507462686567,0) {ITTI2};
\node[left, inner sep=0mm, rotate=90, text=black]
at (rel axis cs:0.0383838383838384,-0.0298507462686567,0) {GR};
\node[left, inner sep=0mm, rotate=90, text=black]
at (rel axis cs:0.0808080808080808,-0.0298507462686567,0) {FTS};
\node[left, inner sep=0mm, rotate=90, text=black]
at (rel axis cs:0.121212121212121,-0.0298507462686567,0) {SUN};
\node[left, inner sep=0mm, rotate=90, text=black]
at (rel axis cs:0.163636363636364,-0.0298507462686567,0) {HouNIPS};
\node[left, inner sep=0mm, rotate=90, text=black]
at (rel axis cs:0.204040404040404,-0.0298507462686567,0) {SigSal};
\node[left, inner sep=0mm, rotate=90, text=black]
at (rel axis cs:0.246464646464646,-0.0298507462686567,0) {SpectRes};
\node[left, inner sep=0mm, rotate=90, text=black]
at (rel axis cs:0.288888888888889,-0.0298507462686567,0) {Torralba};
\node[left, inner sep=0mm, rotate=90, text=black]
at (rel axis cs:0.329292929292929,-0.0298507462686567,0) {AIM};
\node[left, inner sep=0mm, rotate=90, text=black]
at (rel axis cs:0.371717171717172,-0.0298507462686567,0) {GAFFE};
\node[left, inner sep=0mm, rotate=90, text=black]
at (rel axis cs:0.412121212121212,-0.0298507462686567,0) {RandomCS};
\node[left, inner sep=0mm, rotate=90, text=black]
at (rel axis cs:0.454545454545455,-0.0298507462686567,0) {SDSRG};
\node[left, inner sep=0mm, rotate=90, text=black]
at (rel axis cs:0.496969696969697,-0.0298507462686567,0) {SIMFine};
\node[left, inner sep=0mm, rotate=90, text=black]
at (rel axis cs:0.537373737373737,-0.0298507462686567,0) {BMS};
\node[left, inner sep=0mm, rotate=90, text=black]
at (rel axis cs:0.57979797979798,-0.0298507462686567,0) {AWS};
\node[left, inner sep=0mm, rotate=90, text=black]
at (rel axis cs:0.62020202020202,-0.0298507462686567,0) {ITTI};
\node[left, inner sep=0mm, rotate=90, text=black]
at (rel axis cs:0.662626262626263,-0.0298507462686567,0) {GBVSNoCB};
\node[left, inner sep=0mm, rotate=90, text=black]
at (rel axis cs:0.703030303030303,-0.0298507462686567,0) {SIMCoarse};
\node[left, inner sep=0mm, rotate=90, text=black]
at (rel axis cs:0.745454545454545,-0.0298507462686567,0) {Context};
\node[left, inner sep=0mm, rotate=90, text=black]
at (rel axis cs:0.787878787878788,-0.0298507462686567,0) {SCIA};
\node[left, inner sep=0mm, rotate=90, text=black]
at (rel axis cs:0.828282828282828,-0.0298507462686567,0) {Judd};
\node[left, inner sep=0mm, rotate=90, text=black]
at (rel axis cs:0.870707070707071,-0.0298507462686567,0) {GBVS};
\node[left, inner sep=0mm, rotate=90, text=black]
at (rel axis cs:0.911111111111111,-0.0298507462686567,0) {CovSal};
\node[left, inner sep=0mm, rotate=90, text=black]
at (rel axis cs:0.953535353535354,-0.0298507462686567,0) {SDSRL};
\end{axis}
\end{tikzpicture}%
 \\
     (a)Blur & (b)JPEG & (c)Noise     
\end{tabular}
\caption{SAUC, SNSS, SSKLD, SJSD, and SEMD scores for the TUD database for different distortion types:(a) Blur, (b) JPEG and (c) Noise. Blue, green and red lines correspond to low, medium and high levels of distortion, respectively, and the black dotted line represents the average over all distortion levels according to which the models have been sorted.}
\label{fig:MetricScores_TUD}
\end{figure*}

\begin{table}[t]
   \tiny
	\centering
	\caption{ Top 5 Models over 3 or more metrics for the TUD database, for different levels and types of distortions.}
	\tabcolsep=0.02cm
	\resizebox{\columnwidth}{!}{%
	\begin{tabular}{|ccc|ccc|ccc|}
		\hline
		& Blur  &       &       & JPEG  &       &       & Noise &  \\
		\hline
		High  & Medium & Low   & High  & Medium & Low   & High  & Medium & Low \\
		\hline
		ITTI & SDSRL & SDSRG & RandomCS & SimCoarse & SDSRG  & SCIA & Judd & Judd \\
		SCIA &       &       & GR & SDSRG  & SDSRL      & SDSRL & RandomCS &  \\
		Judd &  &  & SDSRG &  &   & CovSal & SCIA & \\
		CovSal &  &  & Context &  &   &  & CovSal & \\		
		\hline
	\end{tabular}%
}
\label{tab:top5overmetrics}%
\end{table}%

Figure~\ref{fig:MetricScores_TUD} shows the performance of each model for the different distortion levels for a particular distortion type in terms of the SAUC, SNSS, SSKLD, SJSD and SEMD metrics, respectively. In each case the models are sorted such that the average peformance over different levels of distortions increases from left to right. 
Table~\ref{tab:top5overmetrics} shows the models that occur in the top five for three or more metrics.  Tables~\ref{tab:stdevs_levels} and \ref{tab:stdevs_dists} show the average standard deviation of the  metric scores for all the models over different levels and types of distortion, respectively. To account for the different ranges of scores for each of the metrics, the metric scores obtained for every model are normalized by the maximum value of the metric over all distortion types for each distortion level (Table~\ref{tab:stdevs_levels}) or over all distortion levels for each distortion type (Table~\ref{tab:stdevs_dists}) before computing the standard deviation. We can draw the following conclusions from Figure~\ref{fig:MetricScores_TUD} and Tables~\ref{tab:top5overmetrics} to~\ref{tab:stdevs_dists}:\\
- The performance for each model changes with a change in distortion type as the ranking of the models in terms of average performance over different levels of distortions changes with the type of distortion for a given metric. Also, the performance for each model varies  with a change in distortion levels. It is found that  increasing the level of distortion improves performance to a great extent for the blur type of distortion.\\ 
- The performance scores for the different metrics show that performance is higher for the high and medium levels of distortions than low levels of distortions (near distortion-free) for most of the models over all metrics as well as all distortion types. This result suggests that addition of certain levels of distortion helps the VA models predict the salient areas in the distorted image. As the eye-tracking results for this database have been recorded for a quality assessment task, it is highly probable that subjects fixated upon distortions which the saliency models capture thereby helping with metric performance. \\
- If we consider the performance of models over different metrics, as shown in Table~\ref{tab:top5overmetrics}, SDSR~\cite{SDSR} performs the best overall by ranking in the top five for three or more of the five metrics for the medium and low distortion levels for the blur distortion type, the high, medium and low distortion levels for the JPEG distortion type, and the high distortion level for the noise distortion type.\\
 \begin{table}[t]
  	\centering
  	\caption{Average standard deviation across models for all metrics for different levels of distortion.}
  	\begin{tabular}{rrrrrr}
  		\toprule
  		& SAUC  & SNSS  & SEMD  & SJSD  & SSKLD \\
  		\midrule
  		High & 0.0510 & 0.33  & 0.0681 & \bf 0.0353 & 0.2601 \\
  		Medium & 0.0257 & 0.1705 & 0.0190 & \bf 0.0144 & 0.2259 \\
  		Low & 0.0281 & 0.2191 & 0.0115 & \bf 0.0069 & 0.2588 \\
  		\bottomrule
  	\end{tabular}%
  	\label{tab:stdevs_levels}%
  \end{table}%
  \begin{table}[t]
  	\centering
  	\caption{Average standard deviation across models for all metrics for different types of distortion.}
  	\begin{tabular}{rrrrrr}
  		\toprule
  		& SAUC  & SNSS  & SEMD  & SJSD  & SSKLD \\
  		\midrule
  		Blur & 0.0768 & 0.4841 & 0.0810 & \bf 0.0409 & 0.3945 \\
  		JPEG & 0.0253 & 0.1811 & 0.0323 & \bf 0.0197 & 0.2291 \\
  		Noise & 0.0492 & 0.3128 & 0.0558 & \bf 0.0314 & 0.2698\\  		
  		\bottomrule
  	\end{tabular}%
  	\label{tab:stdevs_dists}%
  \end{table}%
- From Tables~\ref{tab:stdevs_levels} and \ref{tab:stdevs_dists}, it
can be seen that the proposed SJSD metric has the lowest average standard deviation across different distortion types and levels. This shows that the proposed SJSD metric is less sensitive (and thus more resilient) to the changes in the amount and type of distortion in the sense that it exhibits a more consistent and stable performance across different distortion types and levels as compared to the existing SAUC metric.


\subsubsection{Variation in performance with varying distortion levels for the Judd Low Resolution Database}
\label{subsubsec:JLRResults}
\begin{figure*}[]
	\tiny
	\centering
	\setlength{\figureheight}{0.08\textwidth}
	\setlength{\figurewidth}{0.25\textwidth}
	\begin{tabular}{ccc}    
%
\begin{tikzpicture}

\begin{axis}[%
width=\figurewidth,
height=\figureheight,
clip=false,
scale only axis,
xmin=1,
xmax=25,
xtick={1,2,3,4,5,6,7,8,9,10,11,12,13,14,15,16,17,18,19,20,21,22,23,24},
xticklabels={\empty},
xmajorgrids,
ymin=0.486506839599354,
ymax=0.761292015021609,
ylabel={SAUC},
ymajorgrids,
axis x line*=bottom,
axis y line*=left,
ytick = {0.5 ,0.55,0.6 ,0.65,0.7 ,0.75},ylabel style={font=\tiny},legend style={font=\tiny},
]
\addplot [color=red,solid,line width=1.0pt,mark=asterisk,mark options={solid},forget plot]
  table[row sep=crcr]{1	0.600343363260164\\
2	0.573700606543573\\
3	0.600509423431826\\
4	0.611403236080552\\
5	0.62926073717903\\
6	0.625499111056838\\
7	0.640468517169259\\
8	0.620504866288171\\
9	0.651290475622999\\
10	0.628561010855628\\
11	0.633516581406601\\
12	0.642893317361881\\
13	0.646591031265774\\
14	0.643231438889561\\
15	0.653596656523588\\
16	0.648788230040744\\
17	0.651594798093416\\
18	0.657562062021564\\
19	0.652174749430569\\
20	0.646044190023771\\
21	0.648208058852246\\
22	0.660330452045374\\
23	0.670157234286132\\
24	0.675253127456803\\
};
\addplot [color=green,solid,line width=1.0pt,mark=+,mark options={solid},forget plot]
  table[row sep=crcr]{1	0.56536114860422\\
2	0.581077086110559\\
3	0.582831884540001\\
4	0.588971354319581\\
5	0.611343908377628\\
6	0.606821334047494\\
7	0.614606405112524\\
8	0.631655225161743\\
9	0.630167896981902\\
10	0.641752563772442\\
11	0.639804346623191\\
12	0.639826762309935\\
13	0.643242514398843\\
14	0.641754192995814\\
15	0.646953786964866\\
16	0.647039100760327\\
17	0.65302310437774\\
18	0.649591855826273\\
19	0.659426123657184\\
20	0.657552164266958\\
21	0.661612800313686\\
22	0.668271248563288\\
23	0.678403794329137\\
24	0.679048542897522\\
};
\addplot [color=blue,solid,line width=1.0pt,mark=o,mark options={solid},forget plot]
  table[row sep=crcr]{1	0.545965726165799\\
2	0.58858237542926\\
3	0.574992263592668\\
4	0.581044972641125\\
5	0.60022956599256\\
6	0.608965162147341\\
7	0.625979582863915\\
8	0.647916726786333\\
9	0.627541456456102\\
10	0.652110595337362\\
11	0.651441247945831\\
12	0.645103585400443\\
13	0.64520607642508\\
14	0.65096312392148\\
15	0.652573744324299\\
16	0.659137967054745\\
17	0.656823363779205\\
18	0.658788381076129\\
19	0.662183544783226\\
20	0.672163468869239\\
21	0.675126904354479\\
22	0.680686279386186\\
23	0.692083650019645\\
24	0.690072909286375\\
};
\addplot [color=black,dashed,line width=1.0pt,forget plot]
  table[row sep=crcr]{1	0.570556746010061\\
2	0.581120022694464\\
3	0.586111190521499\\
4	0.593806521013753\\
5	0.613611403849739\\
6	0.613761869083891\\
7	0.627018168381899\\
8	0.633358939412083\\
9	0.636333276353668\\
10	0.640808056655144\\
11	0.641587391991875\\
12	0.64260788835742\\
13	0.645013207363232\\
14	0.645316251935618\\
15	0.651041395937584\\
16	0.651655099285272\\
17	0.653813755416787\\
18	0.655314099641322\\
19	0.657928139290326\\
20	0.658586607719989\\
21	0.661649254506803\\
22	0.669762659998283\\
23	0.680214892878305\\
24	0.681458193213567\\
};
\node[left, inner sep=0mm, rotate=90, text=black]
at (rel axis cs:-0.00202020202020202,-0.0298507462686567,0) {CovSal};
\node[left, inner sep=0mm, rotate=90, text=black]
at (rel axis cs:0.0383838383838384,-0.0298507462686567,0) {FTS};
\node[left, inner sep=0mm, rotate=90, text=black]
at (rel axis cs:0.0808080808080808,-0.0298507462686567,0) {ITTI2};
\node[left, inner sep=0mm, rotate=90, text=black]
at (rel axis cs:0.121212121212121,-0.0298507462686567,0) {RandomCS};
\node[left, inner sep=0mm, rotate=90, text=black]
at (rel axis cs:0.163636363636364,-0.0298507462686567,0) {SCIA};
\node[left, inner sep=0mm, rotate=90, text=black]
at (rel axis cs:0.204040404040404,-0.0298507462686567,0) {GAFFE};
\node[left, inner sep=0mm, rotate=90, text=black]
at (rel axis cs:0.246464646464646,-0.0298507462686567,0) {ITTI};
\node[left, inner sep=0mm, rotate=90, text=black]
at (rel axis cs:0.288888888888889,-0.0298507462686567,0) {SUN};
\node[left, inner sep=0mm, rotate=90, text=black]
at (rel axis cs:0.329292929292929,-0.0298507462686567,0) {Judd};
\node[left, inner sep=0mm, rotate=90, text=black]
at (rel axis cs:0.371717171717172,-0.0298507462686567,0) {SpectRes};
\node[left, inner sep=0mm, rotate=90, text=black]
at (rel axis cs:0.412121212121212,-0.0298507462686567,0) {SigSal};
\node[left, inner sep=0mm, rotate=90, text=black]
at (rel axis cs:0.454545454545455,-0.0298507462686567,0) {GBVSNoCB};
\node[left, inner sep=0mm, rotate=90, text=black]
at (rel axis cs:0.496969696969697,-0.0298507462686567,0) {GBVS};
\node[left, inner sep=0mm, rotate=90, text=black]
at (rel axis cs:0.537373737373737,-0.0298507462686567,0) {HouNIPS};
\node[left, inner sep=0mm, rotate=90, text=black]
at (rel axis cs:0.57979797979798,-0.0298507462686567,0) {SDSRL};
\node[left, inner sep=0mm, rotate=90, text=black]
at (rel axis cs:0.62020202020202,-0.0298507462686567,0) {SDSRG};
\node[left, inner sep=0mm, rotate=90, text=black]
at (rel axis cs:0.662626262626263,-0.0298507462686567,0) {SIMCoarse};
\node[left, inner sep=0mm, rotate=90, text=black]
at (rel axis cs:0.703030303030303,-0.0298507462686567,0) {Context};
\node[left, inner sep=0mm, rotate=90, text=black]
at (rel axis cs:0.745454545454545,-0.0298507462686567,0) {SIMFine};
\node[left, inner sep=0mm, rotate=90, text=black]
at (rel axis cs:0.787878787878788,-0.0298507462686567,0) {Torralba};
\node[left, inner sep=0mm, rotate=90, text=black]
at (rel axis cs:0.828282828282828,-0.0298507462686567,0) {GR};
\node[left, inner sep=0mm, rotate=90, text=black]
at (rel axis cs:0.870707070707071,-0.0298507462686567,0) {AIM};
\node[left, inner sep=0mm, rotate=90, text=black]
at (rel axis cs:0.911111111111111,-0.0298507462686567,0) {AWS};
\node[left, inner sep=0mm, rotate=90, text=black]
at (rel axis cs:0.953535353535354,-0.0298507462686567,0) {BMS};
\end{axis}
\end{tikzpicture}%
 & 
%
\begin{tikzpicture}

\begin{axis}[%
width=\figurewidth,
height=\figureheight,
clip=false,
scale only axis,
xmin=1,
xmax=25,
xtick={1,2,3,4,5,6,7,8,9,10,11,12,13,14,15,16,17,18,19,20,21,22,23,24},
xticklabels={\empty},
xmajorgrids,
ymin=0.486506839599354,
ymax=0.761292015021609,
ylabel={SAUC},
ymajorgrids,
axis x line*=bottom,
axis y line*=left,
legend style={at={(0.5,1.03)},anchor=south,legend columns=4,draw=black,fill=white,legend cell align=left},
ytick = {0.5 ,0.55,0.6 ,0.65,0.7 ,0.75},ylabel style={font=\tiny},legend style={font=\tiny},
]
\addplot [color=red,solid,line width=1.0pt,mark=asterisk,mark options={solid}]
  table[row sep=crcr]{1	0.556993371504284\\
2	0.612377239075912\\
3	0.593365773693809\\
4	0.62486367632264\\
5	0.622551926075345\\
6	0.653561653194826\\
7	0.627718369699239\\
8	0.645219434851685\\
9	0.655018540204132\\
10	0.657493782593261\\
11	0.653699236345795\\
12	0.631519139187647\\
13	0.643847637058115\\
14	0.663324859661319\\
15	0.652176287261719\\
16	0.6546400237234\\
17	0.663966311608687\\
18	0.66224534737028\\
19	0.655774473835227\\
20	0.651418121647925\\
21	0.658261527735506\\
22	0.651610584277936\\
23	0.6652603235934\\
24	0.669621552135164\\
};
\addlegendentry{064};

\addplot [color=green,solid,line width=1.0pt,mark=+,mark options={solid}]
  table[row sep=crcr]{1	0.564882763208183\\
2	0.576833330740136\\
3	0.585052246228901\\
4	0.588870883020944\\
5	0.593170702817947\\
6	0.619450964036069\\
7	0.634157986744609\\
8	0.626558922148133\\
9	0.627756646400501\\
10	0.632736484538294\\
11	0.635591615858398\\
12	0.640168172223967\\
13	0.639886018496848\\
14	0.645862716327021\\
15	0.648698787900203\\
16	0.650390991377376\\
17	0.648471536239891\\
18	0.649274857886337\\
19	0.651842132192287\\
20	0.655658673515295\\
21	0.657834908545454\\
22	0.661303886444516\\
23	0.660540880044522\\
24	0.66513380963784\\
};
\addlegendentry{128};

\addplot [color=blue,solid,line width=1.0pt,mark=o,mark options={solid}]
  table[row sep=crcr]{1	0.557750839922075\\
2	0.554272150885695\\
3	0.587667590291538\\
4	0.574870897421138\\
5	0.576558190887762\\
6	0.603169908314116\\
7	0.632760521872912\\
8	0.624582326866291\\
9	0.626933889923896\\
10	0.621735151480762\\
11	0.628348772229125\\
12	0.646299642333099\\
13	0.647322966746245\\
14	0.643684825012082\\
15	0.653459216595978\\
16	0.653920137266629\\
17	0.646982728963966\\
18	0.651487398784209\\
19	0.657121372005185\\
20	0.659813144852475\\
21	0.661355392637503\\
22	0.666940499784777\\
23	0.66505607195383\\
24	0.668129688623306\\
};
\addlegendentry{256};

\addplot [color=black,dashed,line width=1.0pt]
  table[row sep=crcr]{1	0.559875658211514\\
2	0.581160906900581\\
3	0.588695203404749\\
4	0.596201818921574\\
5	0.597426939927018\\
6	0.62539417518167\\
7	0.631545626105587\\
8	0.63212022795537\\
9	0.636569692176176\\
10	0.637321806204106\\
11	0.639213208144439\\
12	0.639328984581571\\
13	0.64368554076707\\
14	0.650957467000141\\
15	0.6514447639193\\
16	0.652983717455801\\
17	0.653140192270848\\
18	0.654335868013609\\
19	0.654912659344233\\
20	0.655629980005231\\
21	0.659150609639488\\
22	0.659951656835743\\
23	0.663619091863917\\
24	0.667628350132103\\
};
\addlegendentry{Average};

\node[left, inner sep=0mm, rotate=90, text=black]
at (rel axis cs:-0.00202020202020202,-0.0298507462686567,0) {FTS};
\node[left, inner sep=0mm, rotate=90, text=black]
at (rel axis cs:0.0383838383838384,-0.0298507462686567,0) {CovSal};
\node[left, inner sep=0mm, rotate=90, text=black]
at (rel axis cs:0.0808080808080808,-0.0298507462686567,0) {ITTI2};
\node[left, inner sep=0mm, rotate=90, text=black]
at (rel axis cs:0.121212121212121,-0.0298507462686567,0) {RandomCS};
\node[left, inner sep=0mm, rotate=90, text=black]
at (rel axis cs:0.163636363636364,-0.0298507462686567,0) {GAFFE};
\node[left, inner sep=0mm, rotate=90, text=black]
at (rel axis cs:0.204040404040404,-0.0298507462686567,0) {SCIA};
\node[left, inner sep=0mm, rotate=90, text=black]
at (rel axis cs:0.246464646464646,-0.0298507462686567,0) {SpectRes};
\node[left, inner sep=0mm, rotate=90, text=black]
at (rel axis cs:0.288888888888889,-0.0298507462686567,0) {GBVSNoCB};
\node[left, inner sep=0mm, rotate=90, text=black]
at (rel axis cs:0.329292929292929,-0.0298507462686567,0) {ITTI};
\node[left, inner sep=0mm, rotate=90, text=black]
at (rel axis cs:0.371717171717172,-0.0298507462686567,0) {Judd};
\node[left, inner sep=0mm, rotate=90, text=black]
at (rel axis cs:0.412121212121212,-0.0298507462686567,0) {GBVS};
\node[left, inner sep=0mm, rotate=90, text=black]
at (rel axis cs:0.454545454545455,-0.0298507462686567,0) {SUN};
\node[left, inner sep=0mm, rotate=90, text=black]
at (rel axis cs:0.496969696969697,-0.0298507462686567,0) {Torralba};
\node[left, inner sep=0mm, rotate=90, text=black]
at (rel axis cs:0.537373737373737,-0.0298507462686567,0) {SIMCoarse};
\node[left, inner sep=0mm, rotate=90, text=black]
at (rel axis cs:0.57979797979798,-0.0298507462686567,0) {SDSRG};
\node[left, inner sep=0mm, rotate=90, text=black]
at (rel axis cs:0.62020202020202,-0.0298507462686567,0) {HouNIPS};
\node[left, inner sep=0mm, rotate=90, text=black]
at (rel axis cs:0.662626262626263,-0.0298507462686567,0) {SIMFine};
\node[left, inner sep=0mm, rotate=90, text=black]
at (rel axis cs:0.703030303030303,-0.0298507462686567,0) {SDSRL};
\node[left, inner sep=0mm, rotate=90, text=black]
at (rel axis cs:0.745454545454545,-0.0298507462686567,0) {SigSal};
\node[left, inner sep=0mm, rotate=90, text=black]
at (rel axis cs:0.787878787878788,-0.0298507462686567,0) {AWS};
\node[left, inner sep=0mm, rotate=90, text=black]
at (rel axis cs:0.828282828282828,-0.0298507462686567,0) {AIM};
\node[left, inner sep=0mm, rotate=90, text=black]
at (rel axis cs:0.870707070707071,-0.0298507462686567,0) {GR};
\node[left, inner sep=0mm, rotate=90, text=black]
at (rel axis cs:0.911111111111111,-0.0298507462686567,0) {Context};
\node[left, inner sep=0mm, rotate=90, text=black]
at (rel axis cs:0.953535353535354,-0.0298507462686567,0) {BMS};
\end{axis}
\end{tikzpicture}%
 & 
%
\begin{tikzpicture}

\begin{axis}[%
width=\figurewidth,
height=\figureheight,
clip=false,
scale only axis,
xmin=1,
xmax=25,
xtick={1,2,3,4,5,6,7,8,9,10,11,12,13,14,15,16,17,18,19,20,21,22,23,24},
xticklabels={\empty},
xmajorgrids,
ymin=0.486506839599354,
ymax=0.761292015021609,
ylabel={SAUC},
ymajorgrids,
axis x line*=bottom,
axis y line*=left,
ytick = {0.5 ,0.55,0.6 ,0.65,0.7 ,0.75},ylabel style={font=\tiny},legend style={font=\tiny},
]
\addplot [color=red,solid,line width=1.0pt,mark=asterisk,mark options={solid},forget plot]
  table[row sep=crcr]{1	0.540563155110394\\
2	0.590117909713788\\
3	0.606950470467397\\
4	0.609540647717472\\
5	0.622871173689007\\
6	0.624612205016383\\
7	0.618318452067083\\
8	0.609606688047822\\
9	0.633878144214882\\
10	0.620605182555001\\
11	0.631042933100473\\
12	0.632831448388451\\
13	0.643474274425516\\
14	0.631452875530189\\
15	0.634335722424864\\
16	0.639418103297261\\
17	0.651254514830507\\
18	0.637417449036663\\
19	0.629921590734767\\
20	0.639631643774973\\
21	0.647854811283823\\
22	0.649030032584987\\
23	0.647661663020868\\
24	0.64477976575072\\
};
\addplot [color=green,solid,line width=1.0pt,mark=+,mark options={solid},forget plot]
  table[row sep=crcr]{1	0.548210843008843\\
2	0.569218005577004\\
3	0.57102349741197\\
4	0.581233520988294\\
5	0.593123788920459\\
6	0.595645290977123\\
7	0.604911967521517\\
8	0.615421614198373\\
9	0.613324846271541\\
10	0.620292799748768\\
11	0.610472983635369\\
12	0.619398586034789\\
13	0.621641377051884\\
14	0.62148731590232\\
15	0.622702812269208\\
16	0.622456066620269\\
17	0.619892277260206\\
18	0.625727912171541\\
19	0.634060850346161\\
20	0.632632305565297\\
21	0.62955748990368\\
22	0.630907491684569\\
23	0.634540623086103\\
24	0.643586695837317\\
};
\addplot [color=blue,solid,line width=1.0pt,mark=o,mark options={solid},forget plot]
  table[row sep=crcr]{1	0.552480955807661\\
2	0.567200156152433\\
3	0.54982115200018\\
4	0.569545263491527\\
5	0.579201290232598\\
6	0.585212232059157\\
7	0.608888805329182\\
8	0.618420945906664\\
9	0.601406518823896\\
10	0.611953217291274\\
11	0.612956097329758\\
12	0.611704123429924\\
13	0.606997148829216\\
14	0.621502228365102\\
15	0.619553477737208\\
16	0.618060961873863\\
17	0.608889343677873\\
18	0.624068955715344\\
19	0.628107336361718\\
20	0.628431245797059\\
21	0.627763513537533\\
22	0.633513276115779\\
23	0.634237121927623\\
24	0.639224783503995\\
};
\addplot [color=black,dashed,line width=1.0pt,forget plot]
  table[row sep=crcr]{1	0.547084984642299\\
2	0.575512023814408\\
3	0.575931706626516\\
4	0.586773144065764\\
5	0.598398750947355\\
6	0.601823242684221\\
7	0.610706408305927\\
8	0.61448308271762\\
9	0.616203169770106\\
10	0.617617066531681\\
11	0.618157338021867\\
12	0.621311385951055\\
13	0.624037600102205\\
14	0.624814139932537\\
15	0.625530670810427\\
16	0.626645043930464\\
17	0.626678711922862\\
18	0.629071438974516\\
19	0.630696592480882\\
20	0.633565065045776\\
21	0.635058604908345\\
22	0.637816933461778\\
23	0.638813136011531\\
24	0.642530415030677\\
};
\node[left, inner sep=0mm, rotate=90, text=black]
at (rel axis cs:-0.00202020202020202,-0.0298507462686567,0) {FTS};
\node[left, inner sep=0mm, rotate=90, text=black]
at (rel axis cs:0.0383838383838384,-0.0298507462686567,0) {ITTI2};
\node[left, inner sep=0mm, rotate=90, text=black]
at (rel axis cs:0.0808080808080808,-0.0298507462686567,0) {CovSal};
\node[left, inner sep=0mm, rotate=90, text=black]
at (rel axis cs:0.121212121212121,-0.0298507462686567,0) {RandomCS};
\node[left, inner sep=0mm, rotate=90, text=black]
at (rel axis cs:0.163636363636364,-0.0298507462686567,0) {SCIA};
\node[left, inner sep=0mm, rotate=90, text=black]
at (rel axis cs:0.204040404040404,-0.0298507462686567,0) {GAFFE};
\node[left, inner sep=0mm, rotate=90, text=black]
at (rel axis cs:0.246464646464646,-0.0298507462686567,0) {SDSRG};
\node[left, inner sep=0mm, rotate=90, text=black]
at (rel axis cs:0.288888888888889,-0.0298507462686567,0) {SUN};
\node[left, inner sep=0mm, rotate=90, text=black]
at (rel axis cs:0.329292929292929,-0.0298507462686567,0) {GBVSNoCB};
\node[left, inner sep=0mm, rotate=90, text=black]
at (rel axis cs:0.371717171717172,-0.0298507462686567,0) {SpectRes};
\node[left, inner sep=0mm, rotate=90, text=black]
at (rel axis cs:0.412121212121212,-0.0298507462686567,0) {SDSRL};
\node[left, inner sep=0mm, rotate=90, text=black]
at (rel axis cs:0.454545454545455,-0.0298507462686567,0) {SigSal};
\node[left, inner sep=0mm, rotate=90, text=black]
at (rel axis cs:0.496969696969697,-0.0298507462686567,0) {GBVS};
\node[left, inner sep=0mm, rotate=90, text=black]
at (rel axis cs:0.537373737373737,-0.0298507462686567,0) {SIMCoarse};
\node[left, inner sep=0mm, rotate=90, text=black]
at (rel axis cs:0.57979797979798,-0.0298507462686567,0) {Context};
\node[left, inner sep=0mm, rotate=90, text=black]
at (rel axis cs:0.62020202020202,-0.0298507462686567,0) {HouNIPS};
\node[left, inner sep=0mm, rotate=90, text=black]
at (rel axis cs:0.662626262626263,-0.0298507462686567,0) {Judd};
\node[left, inner sep=0mm, rotate=90, text=black]
at (rel axis cs:0.703030303030303,-0.0298507462686567,0) {SIMFine};
\node[left, inner sep=0mm, rotate=90, text=black]
at (rel axis cs:0.745454545454545,-0.0298507462686567,0) {GR};
\node[left, inner sep=0mm, rotate=90, text=black]
at (rel axis cs:0.787878787878788,-0.0298507462686567,0) {Torralba};
\node[left, inner sep=0mm, rotate=90, text=black]
at (rel axis cs:0.828282828282828,-0.0298507462686567,0) {ITTI};
\node[left, inner sep=0mm, rotate=90, text=black]
at (rel axis cs:0.870707070707071,-0.0298507462686567,0) {BMS};
\node[left, inner sep=0mm, rotate=90, text=black]
at (rel axis cs:0.911111111111111,-0.0298507462686567,0) {AIM};
\node[left, inner sep=0mm, rotate=90, text=black]
at (rel axis cs:0.953535353535354,-0.0298507462686567,0) {AWS};
\end{axis}
\end{tikzpicture}%
 \\
%
\begin{tikzpicture}

\begin{axis}[%
width=\figurewidth,
height=\figureheight,
clip=false,
scale only axis,
xmin=1,
xmax=25,
xtick={1,2,3,4,5,6,7,8,9,10,11,12,13,14,15,16,17,18,19,20,21,22,23,24},
xticklabels={\empty},
xmajorgrids,
ymin=0.0908419745906792,
ymax=1.04860616005264,
ylabel={SNSS},
ymajorgrids,
axis x line*=bottom,
axis y line*=left,
ytick = {0.1,0.2,0.3,0.4,0.5,0.6,0.7,0.8,0.9,1  },ylabel style={font=\tiny},legend style={font=\tiny},
]
\addplot [color=red,solid,line width=1.0pt,mark=asterisk,mark options={solid},forget plot]
  table[row sep=crcr]{1	0.291065622658898\\
2	0.383992201998516\\
3	0.405562494052137\\
4	0.342303569728509\\
5	0.410403195843156\\
6	0.511348005493424\\
7	0.648873346901438\\
8	0.602066880938912\\
9	0.484334052475267\\
10	0.52191575585763\\
11	0.55450215246066\\
12	0.536443100948817\\
13	0.577347124280805\\
14	0.532702506132751\\
15	0.537673393265573\\
16	0.5462592558896\\
17	0.542595218198461\\
18	0.600747234500128\\
19	0.621472887945167\\
20	0.725239496468464\\
21	0.673649860510194\\
22	0.684852314730807\\
23	0.733031348266167\\
24	0.753490107456795\\
};
\addplot [color=green,solid,line width=1.0pt,mark=+,mark options={solid},forget plot]
  table[row sep=crcr]{1	0.344370604984829\\
2	0.35511999525927\\
3	0.371429236519423\\
4	0.415546167641315\\
5	0.435503287330217\\
6	0.451295845887552\\
7	0.456377123998515\\
8	0.503018352061288\\
9	0.557023709876449\\
10	0.556564885434648\\
11	0.572131269042933\\
12	0.586971096591937\\
13	0.588045535392793\\
14	0.594807346156481\\
15	0.621451581997962\\
16	0.63382046996142\\
17	0.63847212170418\\
18	0.625444943451553\\
19	0.632502264483445\\
20	0.685099806050784\\
21	0.717836094147976\\
22	0.729240625382016\\
23	0.809709812996388\\
24	0.806121967882454\\
};
\addplot [color=blue,solid,line width=1.0pt,mark=o,mark options={solid},forget plot]
  table[row sep=crcr]{1	0.404299834669438\\
2	0.355482357209139\\
3	0.37676348403885\\
4	0.508932750802668\\
5	0.496346653719133\\
6	0.465751362456179\\
7	0.364135018639006\\
8	0.537828100334645\\
9	0.632440569307086\\
10	0.604591541395884\\
11	0.610902200224878\\
12	0.636820629219352\\
13	0.612714579035765\\
14	0.65544016707579\\
15	0.68144609107988\\
16	0.686053335618857\\
17	0.720429966985005\\
18	0.684325849367991\\
19	0.680342150871266\\
20	0.634575944609382\\
21	0.772552674102261\\
22	0.76940389997614\\
23	0.882836775604607\\
24	0.878265783853534\\
};
\addplot [color=black,dashed,line width=1.0pt,forget plot]
  table[row sep=crcr]{1	0.346578687437722\\
2	0.364864851488975\\
3	0.384585071536803\\
4	0.42226082939083\\
5	0.447417712297502\\
6	0.476131737945718\\
7	0.489795163179653\\
8	0.547637777778282\\
9	0.557932777219601\\
10	0.561024060896054\\
11	0.579178540576157\\
12	0.586744942253369\\
13	0.592702412903121\\
14	0.594316673121674\\
15	0.613523688781138\\
16	0.622044353823292\\
17	0.633832435629215\\
18	0.636839342439891\\
19	0.644772434433293\\
20	0.681638415709543\\
21	0.72134620958681\\
22	0.727832280029654\\
23	0.808525978955721\\
24	0.812625953064261\\
};
\node[left, inner sep=0mm, rotate=90, text=black]
at (rel axis cs:-0.00202020202020202,-0.0298507462686567,0) {FTS};
\node[left, inner sep=0mm, rotate=90, text=black]
at (rel axis cs:0.0383838383838384,-0.0298507462686567,0) {GAFFE};
\node[left, inner sep=0mm, rotate=90, text=black]
at (rel axis cs:0.0808080808080808,-0.0298507462686567,0) {RandomCS};
\node[left, inner sep=0mm, rotate=90, text=black]
at (rel axis cs:0.121212121212121,-0.0298507462686567,0) {SUN};
\node[left, inner sep=0mm, rotate=90, text=black]
at (rel axis cs:0.163636363636364,-0.0298507462686567,0) {ITTI2};
\node[left, inner sep=0mm, rotate=90, text=black]
at (rel axis cs:0.204040404040404,-0.0298507462686567,0) {Judd};
\node[left, inner sep=0mm, rotate=90, text=black]
at (rel axis cs:0.246464646464646,-0.0298507462686567,0) {CovSal};
\node[left, inner sep=0mm, rotate=90, text=black]
at (rel axis cs:0.288888888888889,-0.0298507462686567,0) {ITTI};
\node[left, inner sep=0mm, rotate=90, text=black]
at (rel axis cs:0.329292929292929,-0.0298507462686567,0) {Torralba};
\node[left, inner sep=0mm, rotate=90, text=black]
at (rel axis cs:0.371717171717172,-0.0298507462686567,0) {GBVSNoCB};
\node[left, inner sep=0mm, rotate=90, text=black]
at (rel axis cs:0.412121212121212,-0.0298507462686567,0) {SDSRL};
\node[left, inner sep=0mm, rotate=90, text=black]
at (rel axis cs:0.454545454545455,-0.0298507462686567,0) {AIM};
\node[left, inner sep=0mm, rotate=90, text=black]
at (rel axis cs:0.496969696969697,-0.0298507462686567,0) {GBVS};
\node[left, inner sep=0mm, rotate=90, text=black]
at (rel axis cs:0.537373737373737,-0.0298507462686567,0) {SpectRes};
\node[left, inner sep=0mm, rotate=90, text=black]
at (rel axis cs:0.57979797979798,-0.0298507462686567,0) {GR};
\node[left, inner sep=0mm, rotate=90, text=black]
at (rel axis cs:0.62020202020202,-0.0298507462686567,0) {SigSal};
\node[left, inner sep=0mm, rotate=90, text=black]
at (rel axis cs:0.662626262626263,-0.0298507462686567,0) {HouNIPS};
\node[left, inner sep=0mm, rotate=90, text=black]
at (rel axis cs:0.703030303030303,-0.0298507462686567,0) {SDSRG};
\node[left, inner sep=0mm, rotate=90, text=black]
at (rel axis cs:0.745454545454545,-0.0298507462686567,0) {Context};
\node[left, inner sep=0mm, rotate=90, text=black]
at (rel axis cs:0.787878787878788,-0.0298507462686567,0) {SCIA};
\node[left, inner sep=0mm, rotate=90, text=black]
at (rel axis cs:0.828282828282828,-0.0298507462686567,0) {SIMCoarse};
\node[left, inner sep=0mm, rotate=90, text=black]
at (rel axis cs:0.870707070707071,-0.0298507462686567,0) {SIMFine};
\node[left, inner sep=0mm, rotate=90, text=black]
at (rel axis cs:0.911111111111111,-0.0298507462686567,0) {AWS};
\node[left, inner sep=0mm, rotate=90, text=black]
at (rel axis cs:0.953535353535354,-0.0298507462686567,0) {BMS};
\end{axis}
\end{tikzpicture}%
 & 
%
\begin{tikzpicture}

\begin{axis}[%
width=\figurewidth,
height=\figureheight,
clip=false,
scale only axis,
xmin=1,
xmax=25,
xtick={1,2,3,4,5,6,7,8,9,10,11,12,13,14,15,16,17,18,19,20,21,22,23,24},
xticklabels={\empty},
xmajorgrids,
ymin=0.0908419745906792,
ymax=1.04860616005264,
ylabel={SNSS},
ymajorgrids,
axis x line*=bottom,
axis y line*=left,
ytick = {0.1,0.2,0.3,0.4,0.5,0.6,0.7,0.8,0.9,1  },ylabel style={font=\tiny},legend style={font=\tiny},
]
\addplot [color=red,solid,line width=1.0pt,mark=asterisk,mark options={solid},forget plot]
  table[row sep=crcr]{1	0.224741908684429\\
2	0.338053102680557\\
3	0.428180514253534\\
4	0.376036240614244\\
5	0.316406588709446\\
6	0.448672162333023\\
7	0.529682689835877\\
8	0.443244464885741\\
9	0.522491555275111\\
10	0.491268068948395\\
11	0.472185846170976\\
12	0.547864986446603\\
13	0.724145745135597\\
14	0.477444008059652\\
15	0.663001390179128\\
16	0.522323169931894\\
17	0.593475770945103\\
18	0.605400943767617\\
19	0.581609554523084\\
20	0.661823710556228\\
21	0.595295630908884\\
22	0.674095775143692\\
23	0.700933416090352\\
24	0.953278327320582\\
};
\addplot [color=green,solid,line width=1.0pt,mark=+,mark options={solid},forget plot]
  table[row sep=crcr]{1	0.253948336881989\\
2	0.255597658723314\\
3	0.329313293464481\\
4	0.395757710926084\\
5	0.409361106341381\\
6	0.457290267336811\\
7	0.46055757115467\\
8	0.477921753116721\\
9	0.497035892022375\\
10	0.522236395880974\\
11	0.53801276758442\\
12	0.535382965050478\\
13	0.519423645504209\\
14	0.570325304560491\\
15	0.555879263537879\\
16	0.601823876693847\\
17	0.605345072149778\\
18	0.618489981575488\\
19	0.620760826338204\\
20	0.613859017768058\\
21	0.648532340481411\\
22	0.622486808686758\\
23	0.705217744172923\\
24	0.766138952242439\\
};
\addplot [color=blue,solid,line width=1.0pt,mark=o,mark options={solid},forget plot]
  table[row sep=crcr]{1	0.232784662177814\\
2	0.230472564013033\\
3	0.287523543467785\\
4	0.386879976045875\\
5	0.464936092244231\\
6	0.459073575106826\\
7	0.422539990404502\\
8	0.5038919597939\\
9	0.481521623923771\\
10	0.530391517396481\\
11	0.542938043797404\\
12	0.541495323016915\\
13	0.399565999479537\\
14	0.611959086637633\\
15	0.487627155522832\\
16	0.637729017758682\\
17	0.609935312688922\\
18	0.610264306661202\\
19	0.64569321556703\\
20	0.602871931139738\\
21	0.655262835031121\\
22	0.619033737542453\\
23	0.720682101852687\\
24	0.665508025441858\\
};
\addplot [color=black,dashed,line width=1.0pt,forget plot]
  table[row sep=crcr]{1	0.237158302581411\\
2	0.274707775138968\\
3	0.348339117061933\\
4	0.386224642528734\\
5	0.396901262431686\\
6	0.45501200159222\\
7	0.470926750465016\\
8	0.475019392598788\\
9	0.500349690407086\\
10	0.514631994075283\\
11	0.517712219184267\\
12	0.541581091504665\\
13	0.547711796706448\\
14	0.553242799752592\\
15	0.568835936413279\\
16	0.587292021461474\\
17	0.602918718594601\\
18	0.611385077334769\\
19	0.616021198809439\\
20	0.626184886488008\\
21	0.633030268807139\\
22	0.638538773790967\\
23	0.708944420705321\\
24	0.794975101668293\\
};
\node[left, inner sep=0mm, rotate=90, text=black]
at (rel axis cs:-0.00202020202020202,-0.0298507462686567,0) {FTS};
\node[left, inner sep=0mm, rotate=90, text=black]
at (rel axis cs:0.0383838383838384,-0.0298507462686567,0) {GAFFE};
\node[left, inner sep=0mm, rotate=90, text=black]
at (rel axis cs:0.0808080808080808,-0.0298507462686567,0) {RandomCS};
\node[left, inner sep=0mm, rotate=90, text=black]
at (rel axis cs:0.121212121212121,-0.0298507462686567,0) {ITTI2};
\node[left, inner sep=0mm, rotate=90, text=black]
at (rel axis cs:0.163636363636364,-0.0298507462686567,0) {SUN};
\node[left, inner sep=0mm, rotate=90, text=black]
at (rel axis cs:0.204040404040404,-0.0298507462686567,0) {GBVSNoCB};
\node[left, inner sep=0mm, rotate=90, text=black]
at (rel axis cs:0.246464646464646,-0.0298507462686567,0) {Judd};
\node[left, inner sep=0mm, rotate=90, text=black]
at (rel axis cs:0.288888888888889,-0.0298507462686567,0) {Torralba};
\node[left, inner sep=0mm, rotate=90, text=black]
at (rel axis cs:0.329292929292929,-0.0298507462686567,0) {GBVS};
\node[left, inner sep=0mm, rotate=90, text=black]
at (rel axis cs:0.371717171717172,-0.0298507462686567,0) {AIM};
\node[left, inner sep=0mm, rotate=90, text=black]
at (rel axis cs:0.412121212121212,-0.0298507462686567,0) {SpectRes};
\node[left, inner sep=0mm, rotate=90, text=black]
at (rel axis cs:0.454545454545455,-0.0298507462686567,0) {SDSRL};
\node[left, inner sep=0mm, rotate=90, text=black]
at (rel axis cs:0.496969696969697,-0.0298507462686567,0) {CovSal};
\node[left, inner sep=0mm, rotate=90, text=black]
at (rel axis cs:0.537373737373737,-0.0298507462686567,0) {GR};
\node[left, inner sep=0mm, rotate=90, text=black]
at (rel axis cs:0.57979797979798,-0.0298507462686567,0) {ITTI};
\node[left, inner sep=0mm, rotate=90, text=black]
at (rel axis cs:0.62020202020202,-0.0298507462686567,0) {HouNIPS};
\node[left, inner sep=0mm, rotate=90, text=black]
at (rel axis cs:0.662626262626263,-0.0298507462686567,0) {SDSRG};
\node[left, inner sep=0mm, rotate=90, text=black]
at (rel axis cs:0.703030303030303,-0.0298507462686567,0) {Context};
\node[left, inner sep=0mm, rotate=90, text=black]
at (rel axis cs:0.745454545454545,-0.0298507462686567,0) {SigSal};
\node[left, inner sep=0mm, rotate=90, text=black]
at (rel axis cs:0.787878787878788,-0.0298507462686567,0) {SIMFine};
\node[left, inner sep=0mm, rotate=90, text=black]
at (rel axis cs:0.828282828282828,-0.0298507462686567,0) {AWS};
\node[left, inner sep=0mm, rotate=90, text=black]
at (rel axis cs:0.870707070707071,-0.0298507462686567,0) {SIMCoarse};
\node[left, inner sep=0mm, rotate=90, text=black]
at (rel axis cs:0.911111111111111,-0.0298507462686567,0) {BMS};
\node[left, inner sep=0mm, rotate=90, text=black]
at (rel axis cs:0.953535353535354,-0.0298507462686567,0) {SCIA};
\end{axis}
\end{tikzpicture}%
 & 
%
\begin{tikzpicture}

\begin{axis}[%
width=\figurewidth,
height=\figureheight,
clip=false,
scale only axis,
xmin=1,
xmax=25,
xtick={1,2,3,4,5,6,7,8,9,10,11,12,13,14,15,16,17,18,19,20,21,22,23,24},
xticklabels={\empty},
xmajorgrids,
ymin=0.0908419745906792,
ymax=1.04860616005264,
ylabel={SNSS},
ymajorgrids,
axis x line*=bottom,
axis y line*=left,
ytick = {0.1,0.2,0.3,0.4,0.5,0.6,0.7,0.8,0.9,1  },ylabel style={font=\tiny},legend style={font=\tiny},
]
\addplot [color=red,solid,line width=1.0pt,mark=asterisk,mark options={solid},forget plot]
  table[row sep=crcr]{1	0.100935527322977\\
2	0.325766618268681\\
3	0.351663260675071\\
4	0.286057473862568\\
5	0.375376630800766\\
6	0.404529426949558\\
7	0.424988903031141\\
8	0.488698196105562\\
9	0.401627682031494\\
10	0.402970117770848\\
11	0.459431264603598\\
12	0.485834178300961\\
13	0.457987759180247\\
14	0.444717691185442\\
15	0.45611315802629\\
16	0.476996849500087\\
17	0.496707089585392\\
18	0.664739741754262\\
19	0.626213248666617\\
20	0.553276749045355\\
21	0.578448440637867\\
22	0.618300817939913\\
23	0.599085649077089\\
24	0.791947689156181\\
};
\addplot [color=green,solid,line width=1.0pt,mark=+,mark options={solid},forget plot]
  table[row sep=crcr]{1	0.143837120643224\\
2	0.231665234523652\\
3	0.263115722690922\\
4	0.338006544418997\\
5	0.333535344465083\\
6	0.375462080123012\\
7	0.374486975922792\\
8	0.397400908510287\\
9	0.432358344456378\\
10	0.418957136033557\\
11	0.403554392895386\\
12	0.422134020650114\\
13	0.443814198709943\\
14	0.439212855873516\\
15	0.480248828032656\\
16	0.479047004408972\\
17	0.470608574559619\\
18	0.462028817476154\\
19	0.53576752051927\\
20	0.535479866715438\\
21	0.545713548957318\\
22	0.537741827424515\\
23	0.600095429534133\\
24	0.608653404909264\\
};
\addplot [color=blue,solid,line width=1.0pt,mark=o,mark options={solid},forget plot]
  table[row sep=crcr]{1	0.173062911868437\\
2	0.2160455018876\\
3	0.242282440078372\\
4	0.391987896278209\\
5	0.328927920417408\\
6	0.377231937445281\\
7	0.400687379489679\\
8	0.365220425560995\\
9	0.423140777658908\\
10	0.445985737997771\\
11	0.447067770367344\\
12	0.404674307238494\\
13	0.442806918231283\\
14	0.464071968610155\\
15	0.494530863593043\\
16	0.489149659813967\\
17	0.484447906815746\\
18	0.35843742539845\\
19	0.494894147785387\\
20	0.56873937042059\\
21	0.573855039816058\\
22	0.572410693370828\\
23	0.608015384297126\\
24	0.525732444333297\\
};
\addplot [color=black,dashed,line width=1.0pt,forget plot]
  table[row sep=crcr]{1	0.139278519944879\\
2	0.257825784893311\\
3	0.285687141148122\\
4	0.338683971519925\\
5	0.345946631894419\\
6	0.385741148172617\\
7	0.400054419481204\\
8	0.417106510058948\\
9	0.419042268048927\\
10	0.422637663934058\\
11	0.436684475955443\\
12	0.43754750206319\\
13	0.448202958707158\\
14	0.449334171889704\\
15	0.47696428321733\\
16	0.481731171241009\\
17	0.483921190320252\\
18	0.495068661542955\\
19	0.552291638990425\\
20	0.552498662060461\\
21	0.566005676470414\\
22	0.576151112911752\\
23	0.602398820969449\\
24	0.642111179466248\\
};
\node[left, inner sep=0mm, rotate=90, text=black]
at (rel axis cs:-0.00202020202020202,-0.0298507462686567,0) {FTS};
\node[left, inner sep=0mm, rotate=90, text=black]
at (rel axis cs:0.0383838383838384,-0.0298507462686567,0) {GAFFE};
\node[left, inner sep=0mm, rotate=90, text=black]
at (rel axis cs:0.0808080808080808,-0.0298507462686567,0) {RandomCS};
\node[left, inner sep=0mm, rotate=90, text=black]
at (rel axis cs:0.121212121212121,-0.0298507462686567,0) {SUN};
\node[left, inner sep=0mm, rotate=90, text=black]
at (rel axis cs:0.163636363636364,-0.0298507462686567,0) {ITTI2};
\node[left, inner sep=0mm, rotate=90, text=black]
at (rel axis cs:0.204040404040404,-0.0298507462686567,0) {GBVSNoCB};
\node[left, inner sep=0mm, rotate=90, text=black]
at (rel axis cs:0.246464646464646,-0.0298507462686567,0) {SDSRL};
\node[left, inner sep=0mm, rotate=90, text=black]
at (rel axis cs:0.288888888888889,-0.0298507462686567,0) {Judd};
\node[left, inner sep=0mm, rotate=90, text=black]
at (rel axis cs:0.329292929292929,-0.0298507462686567,0) {SpectRes};
\node[left, inner sep=0mm, rotate=90, text=black]
at (rel axis cs:0.371717171717172,-0.0298507462686567,0) {Torralba};
\node[left, inner sep=0mm, rotate=90, text=black]
at (rel axis cs:0.412121212121212,-0.0298507462686567,0) {SDSRG};
\node[left, inner sep=0mm, rotate=90, text=black]
at (rel axis cs:0.454545454545455,-0.0298507462686567,0) {GBVS};
\node[left, inner sep=0mm, rotate=90, text=black]
at (rel axis cs:0.496969696969697,-0.0298507462686567,0) {SigSal};
\node[left, inner sep=0mm, rotate=90, text=black]
at (rel axis cs:0.537373737373737,-0.0298507462686567,0) {AIM};
\node[left, inner sep=0mm, rotate=90, text=black]
at (rel axis cs:0.57979797979798,-0.0298507462686567,0) {GR};
\node[left, inner sep=0mm, rotate=90, text=black]
at (rel axis cs:0.62020202020202,-0.0298507462686567,0) {HouNIPS};
\node[left, inner sep=0mm, rotate=90, text=black]
at (rel axis cs:0.662626262626263,-0.0298507462686567,0) {Context};
\node[left, inner sep=0mm, rotate=90, text=black]
at (rel axis cs:0.703030303030303,-0.0298507462686567,0) {CovSal};
\node[left, inner sep=0mm, rotate=90, text=black]
at (rel axis cs:0.745454545454545,-0.0298507462686567,0) {ITTI};
\node[left, inner sep=0mm, rotate=90, text=black]
at (rel axis cs:0.787878787878788,-0.0298507462686567,0) {SIMCoarse};
\node[left, inner sep=0mm, rotate=90, text=black]
at (rel axis cs:0.828282828282828,-0.0298507462686567,0) {SIMFine};
\node[left, inner sep=0mm, rotate=90, text=black]
at (rel axis cs:0.870707070707071,-0.0298507462686567,0) {BMS};
\node[left, inner sep=0mm, rotate=90, text=black]
at (rel axis cs:0.911111111111111,-0.0298507462686567,0) {AWS};
\node[left, inner sep=0mm, rotate=90, text=black]
at (rel axis cs:0.953535353535354,-0.0298507462686567,0) {SCIA};
\end{axis}
\end{tikzpicture}%
 \\
%
\begin{tikzpicture}

\begin{axis}[%
width=\figurewidth,
height=\figureheight,
clip=false,
scale only axis,
xmin=1,
xmax=25,
xtick={1,2,3,4,5,6,7,8,9,10,11,12,13,14,15,16,17,18,19,20,21,22,23,24},
xticklabels={\empty},
xmajorgrids,
ymin=8.25638568744857,
ymax=17.5139742896068,
ylabel={SSKLD},
ymajorgrids,
axis x line*=bottom,
axis y line*=left,
ytick = {9 ,10,11,12,13,14,15,16,17},ylabel style={font=\tiny},legend style={font=\tiny},
]
\addplot [color=red,solid,line width=1.0pt,mark=asterisk,mark options={solid},forget plot]
  table[row sep=crcr]{1	12.4385916388399\\
2	12.4353004316766\\
3	12.2761332243285\\
4	11.7714809359986\\
5	12.8622488483966\\
6	13.0000772515529\\
7	13.0513753874511\\
8	13.3571857978181\\
9	13.6198069515865\\
10	13.0925074635624\\
11	13.3224054541766\\
12	13.7415980252564\\
13	13.4922842701024\\
14	13.3305917566046\\
15	14.0377082976447\\
16	13.952004436047\\
17	13.8392130801452\\
18	14.2122458432079\\
19	14.0712658413133\\
20	14.9223776183423\\
21	15.3162283597782\\
22	14.9767393405726\\
23	15.4673792506625\\
24	15.8904356183843\\
};
\addplot [color=green,solid,line width=1.0pt,mark=+,mark options={solid},forget plot]
  table[row sep=crcr]{1	11.6731123025661\\
2	11.8904762883994\\
3	11.6510869906635\\
4	11.912242655224\\
5	11.9807689617156\\
6	12.415303468711\\
7	12.5906000929368\\
8	12.5422787138044\\
9	12.5721877679423\\
10	12.7102482798183\\
11	13.0767614966745\\
12	13.0435286274222\\
13	13.0937072545033\\
14	13.0584869414108\\
15	12.9484647460352\\
16	13.0163624099418\\
17	13.2389942681304\\
18	13.3035608458169\\
19	13.2833713288402\\
20	14.0472626130176\\
21	14.3588588955422\\
22	14.3759135583784\\
23	14.5642156532687\\
24	14.6195200508718\\
};
\addplot [color=blue,solid,line width=1.0pt,mark=o,mark options={solid},forget plot]
  table[row sep=crcr]{1	11.1934982735878\\
2	11.1522003772066\\
3	11.9635134175096\\
4	12.2101768347536\\
5	11.12740971199\\
6	12.2889982626885\\
7	12.5328424021039\\
8	12.7050035936816\\
9	12.4790152160197\\
10	13.1944302143615\\
11	12.6125132900003\\
12	12.6732229069575\\
13	12.8919523090335\\
14	13.2239898775285\\
15	12.7637244014924\\
16	12.8060967308519\\
17	12.9275062031507\\
18	13.0395596083763\\
19	13.3159507216144\\
20	13.7833483455373\\
21	13.3698559288456\\
22	14.3576656040115\\
23	14.0526515558267\\
24	14.2666152869149\\
};
\addplot [color=black,dashed,line width=1.0pt,forget plot]
  table[row sep=crcr]{1	11.7684007383313\\
2	11.8259923657609\\
3	11.9635778775005\\
4	11.9646334753254\\
5	11.9901425073674\\
6	12.5681263276508\\
7	12.7249392941639\\
8	12.8681560351014\\
9	12.8903366451829\\
10	12.9990619859141\\
11	13.0038934136171\\
12	13.1527831865454\\
13	13.1593146112131\\
14	13.2043561918479\\
15	13.2499658150574\\
16	13.2581545256136\\
17	13.3352378504754\\
18	13.518455432467\\
19	13.5568626305893\\
20	14.2509961922991\\
21	14.348314394722\\
22	14.5701061676541\\
23	14.6947488199193\\
24	14.925523652057\\
};
\node[left, inner sep=0mm, rotate=90, text=black]
at (rel axis cs:-0.00202020202020202,-0.0328358208955224,0) {GAFFE};
\node[left, inner sep=0mm, rotate=90, text=black]
at (rel axis cs:0.0383838383838384,-0.0328358208955224,0) {ITTI2};
\node[left, inner sep=0mm, rotate=90, text=black]
at (rel axis cs:0.0808080808080808,-0.0328358208955224,0) {FTS};
\node[left, inner sep=0mm, rotate=90, text=black]
at (rel axis cs:0.121212121212121,-0.0328358208955224,0) {SUN};
\node[left, inner sep=0mm, rotate=90, text=black]
at (rel axis cs:0.163636363636364,-0.0328358208955224,0) {Judd};
\node[left, inner sep=0mm, rotate=90, text=black]
at (rel axis cs:0.204040404040404,-0.0328358208955224,0) {SigSal};
\node[left, inner sep=0mm, rotate=90, text=black]
at (rel axis cs:0.246464646464646,-0.0328358208955224,0) {Torralba};
\node[left, inner sep=0mm, rotate=90, text=black]
at (rel axis cs:0.288888888888889,-0.0328358208955224,0) {AIM};
\node[left, inner sep=0mm, rotate=90, text=black]
at (rel axis cs:0.329292929292929,-0.0328358208955224,0) {SDSRL};
\node[left, inner sep=0mm, rotate=90, text=black]
at (rel axis cs:0.371717171717172,-0.0328358208955224,0) {GR};
\node[left, inner sep=0mm, rotate=90, text=black]
at (rel axis cs:0.412121212121212,-0.0328358208955224,0) {RandomCS};
\node[left, inner sep=0mm, rotate=90, text=black]
at (rel axis cs:0.454545454545455,-0.0328358208955224,0) {SIMFine};
\node[left, inner sep=0mm, rotate=90, text=black]
at (rel axis cs:0.496969696969697,-0.0328358208955224,0) {GBVSNoCB};
\node[left, inner sep=0mm, rotate=90, text=black]
at (rel axis cs:0.537373737373737,-0.0328358208955224,0) {HouNIPS};
\node[left, inner sep=0mm, rotate=90, text=black]
at (rel axis cs:0.57979797979798,-0.0328358208955224,0) {SDSRG};
\node[left, inner sep=0mm, rotate=90, text=black]
at (rel axis cs:0.62020202020202,-0.0328358208955224,0) {SIMCoarse};
\node[left, inner sep=0mm, rotate=90, text=black]
at (rel axis cs:0.662626262626263,-0.0328358208955224,0) {GBVS};
\node[left, inner sep=0mm, rotate=90, text=black]
at (rel axis cs:0.703030303030303,-0.0328358208955224,0) {Context};
\node[left, inner sep=0mm, rotate=90, text=black]
at (rel axis cs:0.745454545454545,-0.0328358208955224,0) {SpectRes};
\node[left, inner sep=0mm, rotate=90, text=black]
at (rel axis cs:0.787878787878788,-0.0328358208955224,0) {BMS};
\node[left, inner sep=0mm, rotate=90, text=black]
at (rel axis cs:0.828282828282828,-0.0328358208955224,0) {SCIA};
\node[left, inner sep=0mm, rotate=90, text=black]
at (rel axis cs:0.870707070707071,-0.0328358208955224,0) {AWS};
\node[left, inner sep=0mm, rotate=90, text=black]
at (rel axis cs:0.911111111111111,-0.0328358208955224,0) {ITTI};
\node[left, inner sep=0mm, rotate=90, text=black]
at (rel axis cs:0.953535353535354,-0.0328358208955224,0) {CovSal};
\end{axis}
\end{tikzpicture}%
 & 
%
\begin{tikzpicture}

\begin{axis}[%
width=\figurewidth,
height=\figureheight,
clip=false,
scale only axis,
xmin=1,
xmax=25,
xtick={1,2,3,4,5,6,7,8,9,10,11,12,13,14,15,16,17,18,19,20,21,22,23,24},
xticklabels={\empty},
xmajorgrids,
ymin=8.25638568744857,
ymax=17.5139742896068,
ylabel={SSKLD},
ymajorgrids,
axis x line*=bottom,
axis y line*=left,
ytick = {9 ,10,11,12,13,14,15,16,17},ylabel style={font=\tiny},legend style={font=\tiny},
]
\addplot [color=red,solid,line width=1.0pt,mark=asterisk,mark options={solid},forget plot]
  table[row sep=crcr]{1	11.5134760250242\\
2	11.6106553409031\\
3	11.394791332893\\
4	10.8437376974708\\
5	12.400867634458\\
6	13.1979593297271\\
7	12.7603890136611\\
8	12.4091147395616\\
9	13.0085006329529\\
10	12.804173848026\\
11	13.938550183324\\
12	13.4102016151622\\
13	13.4654187467044\\
14	13.7468042804732\\
15	13.9011570074678\\
16	13.5753620423626\\
17	13.3282128872478\\
18	13.9050387509\\
19	14.4925807312094\\
20	14.3538975044602\\
21	15.0649235350781\\
22	15.0240993813813\\
23	15.4617425662568\\
24	15.9217948087334\\
};
\addplot [color=green,solid,line width=1.0pt,mark=+,mark options={solid},forget plot]
  table[row sep=crcr]{1	10.723302095494\\
2	10.9147421907124\\
3	11.1591944018836\\
4	11.4094808718916\\
5	11.5403558742938\\
6	11.8407884709129\\
7	11.9239770208974\\
8	12.1679964339168\\
9	12.3606592150596\\
10	12.8745812516954\\
11	12.5276740863266\\
12	12.8171736347015\\
13	12.9182305470634\\
14	13.0280046176294\\
15	13.0749207458085\\
16	13.2047325398555\\
17	13.453313346294\\
18	13.4154447473089\\
19	13.5820937488694\\
20	13.8431063231663\\
21	13.3744697652573\\
22	13.8719690141007\\
23	14.4202257612372\\
24	14.7732864066983\\
};
\addplot [color=blue,solid,line width=1.0pt,mark=o,mark options={solid},forget plot]
  table[row sep=crcr]{1	9.87419387798387\\
2	10.2568759181155\\
3	10.7179269737153\\
4	11.0890372553165\\
5	11.2817159939415\\
6	10.8915962393796\\
7	11.4743519526905\\
8	11.9976708535758\\
9	11.7820483199003\\
10	12.6218185718952\\
11	11.8678529737658\\
12	12.1548111123956\\
13	12.1721204866002\\
14	12.3237084533404\\
15	12.2110782291097\\
16	12.7066805605048\\
17	12.8616600227906\\
18	12.9639852966738\\
19	12.6760596684512\\
20	13.0807001290291\\
21	12.9335059917918\\
22	12.8270879267374\\
23	13.2737433728524\\
24	13.0666646887804\\
};
\addplot [color=black,dashed,line width=1.0pt,forget plot]
  table[row sep=crcr]{1	10.703657332834\\
2	10.9274244832437\\
3	11.0906375694973\\
4	11.114085274893\\
5	11.7409798342311\\
6	11.9767813466732\\
7	12.0529059957497\\
8	12.1915940090181\\
9	12.3837360559709\\
10	12.7668578905389\\
11	12.7780257478055\\
12	12.7940621207531\\
13	12.8519232601227\\
14	13.0328391171477\\
15	13.062385327462\\
16	13.1622583809076\\
17	13.2143954187775\\
18	13.4281562649609\\
19	13.58357804951\\
20	13.7592346522185\\
21	13.7909664307091\\
22	13.9077187740731\\
23	14.3852372334488\\
24	14.5872486347374\\
};
\node[left, inner sep=0mm, rotate=90, text=black]
at (rel axis cs:-0.00202020202020202,-0.0328358208955224,0) {GAFFE};
\node[left, inner sep=0mm, rotate=90, text=black]
at (rel axis cs:0.0383838383838384,-0.0328358208955224,0) {ITTI2};
\node[left, inner sep=0mm, rotate=90, text=black]
at (rel axis cs:0.0808080808080808,-0.0328358208955224,0) {FTS};
\node[left, inner sep=0mm, rotate=90, text=black]
at (rel axis cs:0.121212121212121,-0.0328358208955224,0) {SUN};
\node[left, inner sep=0mm, rotate=90, text=black]
at (rel axis cs:0.163636363636364,-0.0328358208955224,0) {AIM};
\node[left, inner sep=0mm, rotate=90, text=black]
at (rel axis cs:0.204040404040404,-0.0328358208955224,0) {Judd};
\node[left, inner sep=0mm, rotate=90, text=black]
at (rel axis cs:0.246464646464646,-0.0328358208955224,0) {RandomCS};
\node[left, inner sep=0mm, rotate=90, text=black]
at (rel axis cs:0.288888888888889,-0.0328358208955224,0) {Torralba};
\node[left, inner sep=0mm, rotate=90, text=black]
at (rel axis cs:0.329292929292929,-0.0328358208955224,0) {GBVSNoCB};
\node[left, inner sep=0mm, rotate=90, text=black]
at (rel axis cs:0.371717171717172,-0.0328358208955224,0) {GR};
\node[left, inner sep=0mm, rotate=90, text=black]
at (rel axis cs:0.412121212121212,-0.0328358208955224,0) {SDSRL};
\node[left, inner sep=0mm, rotate=90, text=black]
at (rel axis cs:0.454545454545455,-0.0328358208955224,0) {GBVS};
\node[left, inner sep=0mm, rotate=90, text=black]
at (rel axis cs:0.496969696969697,-0.0328358208955224,0) {SDSRG};
\node[left, inner sep=0mm, rotate=90, text=black]
at (rel axis cs:0.537373737373737,-0.0328358208955224,0) {SIMFine};
\node[left, inner sep=0mm, rotate=90, text=black]
at (rel axis cs:0.57979797979798,-0.0328358208955224,0) {SIMCoarse};
\node[left, inner sep=0mm, rotate=90, text=black]
at (rel axis cs:0.62020202020202,-0.0328358208955224,0) {SigSal};
\node[left, inner sep=0mm, rotate=90, text=black]
at (rel axis cs:0.662626262626263,-0.0328358208955224,0) {SpectRes};
\node[left, inner sep=0mm, rotate=90, text=black]
at (rel axis cs:0.703030303030303,-0.0328358208955224,0) {HouNIPS};
\node[left, inner sep=0mm, rotate=90, text=black]
at (rel axis cs:0.745454545454545,-0.0328358208955224,0) {Context};
\node[left, inner sep=0mm, rotate=90, text=black]
at (rel axis cs:0.787878787878788,-0.0328358208955224,0) {AWS};
\node[left, inner sep=0mm, rotate=90, text=black]
at (rel axis cs:0.828282828282828,-0.0328358208955224,0) {BMS};
\node[left, inner sep=0mm, rotate=90, text=black]
at (rel axis cs:0.870707070707071,-0.0328358208955224,0) {SCIA};
\node[left, inner sep=0mm, rotate=90, text=black]
at (rel axis cs:0.911111111111111,-0.0328358208955224,0) {CovSal};
\node[left, inner sep=0mm, rotate=90, text=black]
at (rel axis cs:0.953535353535354,-0.0328358208955224,0) {ITTI};
\end{axis}
\end{tikzpicture}%
 & 
%
\begin{tikzpicture}

\begin{axis}[%
width=\figurewidth,
height=\figureheight,
clip=false,
scale only axis,
xmin=1,
xmax=25,
xtick={1,2,3,4,5,6,7,8,9,10,11,12,13,14,15,16,17,18,19,20,21,22,23,24},
xticklabels={\empty},
xmajorgrids,
ymin=8.25638568744857,
ymax=17.5139742896068,
ylabel={SSKLD},
ymajorgrids,
axis x line*=bottom,
axis y line*=left,
ytick = {9 ,10,11,12,13,14,15,16,17},ylabel style={font=\tiny},legend style={font=\tiny},
]
\addplot [color=red,solid,line width=1.0pt,mark=asterisk,mark options={solid},forget plot]
  table[row sep=crcr]{1	9.96076425203888\\
2	10.1061907608868\\
3	10.8780005548597\\
4	9.86048508126126\\
5	10.7909028036818\\
6	11.2500383821504\\
7	11.2326546339937\\
8	11.2702953974123\\
9	11.2352505139052\\
10	10.9388685107018\\
11	10.8803392597182\\
12	10.9863035030491\\
13	11.1770121761683\\
14	11.1536708286264\\
15	11.4879022830048\\
16	11.8756770416465\\
17	11.2260570934796\\
18	12.0895392592533\\
19	11.9536358712805\\
20	12.8699323699027\\
21	12.579551312813\\
22	13.3660472328927\\
23	14.1961831685313\\
24	14.1151875299591\\
};
\addplot [color=green,solid,line width=1.0pt,mark=+,mark options={solid},forget plot]
  table[row sep=crcr]{1	9.17376187494285\\
2	9.36690544138562\\
3	9.81792376503233\\
4	10.1551105370817\\
5	10.446569097609\\
6	10.3701476401949\\
7	10.2707288788325\\
8	10.335324442087\\
9	10.2534582610523\\
10	10.5008258752614\\
11	10.5532544619243\\
12	10.7901425833577\\
13	11.2207062778121\\
14	11.1225439763136\\
15	10.9665812529322\\
16	11.1012342412302\\
17	11.1741563929784\\
18	11.1089840800917\\
19	11.3246013594438\\
20	11.5607729110152\\
21	11.9072185686219\\
22	12.0785016397781\\
23	12.5658953981352\\
24	12.8118745294771\\
};
\addplot [color=blue,solid,line width=1.0pt,mark=o,mark options={solid},forget plot]
  table[row sep=crcr]{1	9.28619009279219\\
2	9.28079069844232\\
3	9.87226132801642\\
4	10.7231579552645\\
5	10.2191098931426\\
6	9.87922715027952\\
7	10.3345628499651\\
8	10.4159220479212\\
9	10.5859436895401\\
10	10.6349884862279\\
11	10.6915238821131\\
12	10.7230970424457\\
13	10.8957039135019\\
14	11.2735365498842\\
15	11.1750879573128\\
16	10.8255914866979\\
17	11.5090693729167\\
18	10.7765065834456\\
19	11.1992105922219\\
20	11.5856965428287\\
21	11.9432414245639\\
22	12.1591355246628\\
23	11.5991497754063\\
24	12.8868382452995\\
};
\addplot [color=black,dashed,line width=1.0pt,forget plot]
  table[row sep=crcr]{1	9.47357207325798\\
2	9.58462896690491\\
3	10.1893952159695\\
4	10.2462511912025\\
5	10.4855272648111\\
6	10.4998043908749\\
7	10.6126487875971\\
8	10.6738472958068\\
9	10.6915508214992\\
10	10.691560957397\\
11	10.7083725345852\\
12	10.8331810429508\\
13	11.0978074558274\\
14	11.1832504516081\\
15	11.2098571644166\\
16	11.2675009231915\\
17	11.3030942864582\\
18	11.3250099742636\\
19	11.4924826076487\\
20	12.0054672745822\\
21	12.1433371019996\\
22	12.5345614657779\\
23	12.7870761140243\\
24	13.2713001015786\\
};
\node[left, inner sep=0mm, rotate=90, text=black]
at (rel axis cs:-0.00202020202020202,-0.0328358208955224,0) {FTS};
\node[left, inner sep=0mm, rotate=90, text=black]
at (rel axis cs:0.0383838383838384,-0.0328358208955224,0) {GAFFE};
\node[left, inner sep=0mm, rotate=90, text=black]
at (rel axis cs:0.0808080808080808,-0.0328358208955224,0) {ITTI2};
\node[left, inner sep=0mm, rotate=90, text=black]
at (rel axis cs:0.121212121212121,-0.0328358208955224,0) {SUN};
\node[left, inner sep=0mm, rotate=90, text=black]
at (rel axis cs:0.163636363636364,-0.0328358208955224,0) {SigSal};
\node[left, inner sep=0mm, rotate=90, text=black]
at (rel axis cs:0.204040404040404,-0.0328358208955224,0) {Judd};
\node[left, inner sep=0mm, rotate=90, text=black]
at (rel axis cs:0.246464646464646,-0.0328358208955224,0) {SDSRG};
\node[left, inner sep=0mm, rotate=90, text=black]
at (rel axis cs:0.288888888888889,-0.0328358208955224,0) {SDSRL};
\node[left, inner sep=0mm, rotate=90, text=black]
at (rel axis cs:0.329292929292929,-0.0328358208955224,0) {GBVSNoCB};
\node[left, inner sep=0mm, rotate=90, text=black]
at (rel axis cs:0.371717171717172,-0.0328358208955224,0) {AIM};
\node[left, inner sep=0mm, rotate=90, text=black]
at (rel axis cs:0.412121212121212,-0.0328358208955224,0) {Torralba};
\node[left, inner sep=0mm, rotate=90, text=black]
at (rel axis cs:0.454545454545455,-0.0328358208955224,0) {SpectRes};
\node[left, inner sep=0mm, rotate=90, text=black]
at (rel axis cs:0.496969696969697,-0.0328358208955224,0) {HouNIPS};
\node[left, inner sep=0mm, rotate=90, text=black]
at (rel axis cs:0.537373737373737,-0.0328358208955224,0) {SIMFine};
\node[left, inner sep=0mm, rotate=90, text=black]
at (rel axis cs:0.57979797979798,-0.0328358208955224,0) {GR};
\node[left, inner sep=0mm, rotate=90, text=black]
at (rel axis cs:0.62020202020202,-0.0328358208955224,0) {RandomCS};
\node[left, inner sep=0mm, rotate=90, text=black]
at (rel axis cs:0.662626262626263,-0.0328358208955224,0) {SIMCoarse};
\node[left, inner sep=0mm, rotate=90, text=black]
at (rel axis cs:0.703030303030303,-0.0328358208955224,0) {GBVS};
\node[left, inner sep=0mm, rotate=90, text=black]
at (rel axis cs:0.745454545454545,-0.0328358208955224,0) {Context};
\node[left, inner sep=0mm, rotate=90, text=black]
at (rel axis cs:0.787878787878788,-0.0328358208955224,0) {BMS};
\node[left, inner sep=0mm, rotate=90, text=black]
at (rel axis cs:0.828282828282828,-0.0328358208955224,0) {AWS};
\node[left, inner sep=0mm, rotate=90, text=black]
at (rel axis cs:0.870707070707071,-0.0328358208955224,0) {SCIA};
\node[left, inner sep=0mm, rotate=90, text=black]
at (rel axis cs:0.911111111111111,-0.0328358208955224,0) {ITTI};
\node[left, inner sep=0mm, rotate=90, text=black]
at (rel axis cs:0.953535353535354,-0.0328358208955224,0) {CovSal};
\end{axis}
\end{tikzpicture}%
 \\
%
\begin{tikzpicture}

\begin{axis}[%
width=\figurewidth,
height=\figureheight,
clip=false,
scale only axis,
xmin=1,
xmax=25,
xtick={1,2,3,4,5,6,7,8,9,10,11,12,13,14,15,16,17,18,19,20,21,22,23,24},
xticklabels={\empty},
xmajorgrids,
ymin=0.322303702521356,
ymax=0.943909334486764,
ylabel={SJSD},
ymajorgrids,
axis x line*=bottom,
axis y line*=left,
ytick = {0.4,0.5,0.6,0.7,0.8,0.9},ylabel style={font=\tiny},legend style={font=\tiny},
]
\addplot [color=red,solid,line width=1.0pt,mark=asterisk,mark options={solid},forget plot]
  table[row sep=crcr]{1	0.385616826205389\\
2	0.719706280185706\\
3	0.736637533106261\\
4	0.755579538945873\\
5	0.719044879405256\\
6	0.776890839441443\\
7	0.792064695419664\\
8	0.778443506122976\\
9	0.777956197864739\\
10	0.782691112702265\\
11	0.782684997672722\\
12	0.751352190112408\\
13	0.787370853715912\\
14	0.799973975500088\\
15	0.801243911805559\\
16	0.813770939584019\\
17	0.796245032459172\\
18	0.8064437805364\\
19	0.809022964278661\\
20	0.802626302190115\\
21	0.80513007914878\\
22	0.80978178165635\\
23	0.854141122621186\\
24	0.858099394987967\\
};
\addplot [color=green,solid,line width=1.0pt,mark=+,mark options={solid},forget plot]
  table[row sep=crcr]{1	0.391248765385332\\
2	0.721351290231801\\
3	0.729066493797234\\
4	0.75912487515978\\
5	0.784730040135028\\
6	0.769851759940148\\
7	0.77410251738594\\
8	0.780058175306496\\
9	0.781741190871273\\
10	0.780730548008477\\
11	0.78305109025\\
12	0.798354016741648\\
13	0.789342895831511\\
14	0.787521938825396\\
15	0.796229065778416\\
16	0.790191837731694\\
17	0.803466938370446\\
18	0.798179517403975\\
19	0.802634631768602\\
20	0.804137618325697\\
21	0.810420426246022\\
22	0.829037631518006\\
23	0.840366282669879\\
24	0.840659302183748\\
};
\addplot [color=blue,solid,line width=1.0pt,mark=o,mark options={solid},forget plot]
  table[row sep=crcr]{1	0.399315404610949\\
2	0.706517638796335\\
3	0.729730569601925\\
4	0.762262079991027\\
5	0.783821170057604\\
6	0.772387205990265\\
7	0.76598593107405\\
8	0.781625916540628\\
9	0.782557595092314\\
10	0.779224989419426\\
11	0.783063657546014\\
12	0.81021931710165\\
13	0.784223381062733\\
14	0.78256937539208\\
15	0.794215712981988\\
16	0.791208538968848\\
17	0.801385170244037\\
18	0.797724197493586\\
19	0.793669601333749\\
20	0.809289741379983\\
21	0.812756256224249\\
22	0.82952239566027\\
23	0.822447858522889\\
24	0.830906746640165\\
};
\addplot [color=black,dashed,line width=1.0pt,forget plot]
  table[row sep=crcr]{1	0.392060332067224\\
2	0.715858403071281\\
3	0.731811532168473\\
4	0.75898883136556\\
5	0.762532029865963\\
6	0.773043268457285\\
7	0.777384381293218\\
8	0.7800425326567\\
9	0.780751661276109\\
10	0.780882216710056\\
11	0.782933248489579\\
12	0.786641841318569\\
13	0.786979043536719\\
14	0.790021763239188\\
15	0.797229563521988\\
16	0.798390438761521\\
17	0.800365713691218\\
18	0.800782498477987\\
19	0.801775732460337\\
20	0.805351220631932\\
21	0.80943558720635\\
22	0.822780602944875\\
23	0.838985087937985\\
24	0.84322181460396\\
};
\node[left, inner sep=0mm, rotate=90, text=black]
at (rel axis cs:-0.00202020202020202,-0.0298507462686567,0) {ITTI2};
\node[left, inner sep=0mm, rotate=90, text=black]
at (rel axis cs:0.0383838383838384,-0.0298507462686567,0) {GR};
\node[left, inner sep=0mm, rotate=90, text=black]
at (rel axis cs:0.0808080808080808,-0.0298507462686567,0) {FTS};
\node[left, inner sep=0mm, rotate=90, text=black]
at (rel axis cs:0.121212121212121,-0.0298507462686567,0) {SUN};
\node[left, inner sep=0mm, rotate=90, text=black]
at (rel axis cs:0.163636363636364,-0.0298507462686567,0) {SpectRes};
\node[left, inner sep=0mm, rotate=90, text=black]
at (rel axis cs:0.204040404040404,-0.0298507462686567,0) {HouNIPS};
\node[left, inner sep=0mm, rotate=90, text=black]
at (rel axis cs:0.246464646464646,-0.0298507462686567,0) {GAFFE};
\node[left, inner sep=0mm, rotate=90, text=black]
at (rel axis cs:0.288888888888889,-0.0298507462686567,0) {BMS};
\node[left, inner sep=0mm, rotate=90, text=black]
at (rel axis cs:0.329292929292929,-0.0298507462686567,0) {SigSal};
\node[left, inner sep=0mm, rotate=90, text=black]
at (rel axis cs:0.371717171717172,-0.0298507462686567,0) {GBVSNoCB};
\node[left, inner sep=0mm, rotate=90, text=black]
at (rel axis cs:0.412121212121212,-0.0298507462686567,0) {SIMFine};
\node[left, inner sep=0mm, rotate=90, text=black]
at (rel axis cs:0.454545454545455,-0.0298507462686567,0) {SDSRG};
\node[left, inner sep=0mm, rotate=90, text=black]
at (rel axis cs:0.496969696969697,-0.0298507462686567,0) {SCIA};
\node[left, inner sep=0mm, rotate=90, text=black]
at (rel axis cs:0.537373737373737,-0.0298507462686567,0) {RandomCS};
\node[left, inner sep=0mm, rotate=90, text=black]
at (rel axis cs:0.57979797979798,-0.0298507462686567,0) {GBVS};
\node[left, inner sep=0mm, rotate=90, text=black]
at (rel axis cs:0.62020202020202,-0.0298507462686567,0) {Torralba};
\node[left, inner sep=0mm, rotate=90, text=black]
at (rel axis cs:0.662626262626263,-0.0298507462686567,0) {SIMCoarse};
\node[left, inner sep=0mm, rotate=90, text=black]
at (rel axis cs:0.703030303030303,-0.0298507462686567,0) {AIM};
\node[left, inner sep=0mm, rotate=90, text=black]
at (rel axis cs:0.745454545454545,-0.0298507462686567,0) {Judd};
\node[left, inner sep=0mm, rotate=90, text=black]
at (rel axis cs:0.787878787878788,-0.0298507462686567,0) {Context};
\node[left, inner sep=0mm, rotate=90, text=black]
at (rel axis cs:0.828282828282828,-0.0298507462686567,0) {AWS};
\node[left, inner sep=0mm, rotate=90, text=black]
at (rel axis cs:0.870707070707071,-0.0298507462686567,0) {SDSRL};
\node[left, inner sep=0mm, rotate=90, text=black]
at (rel axis cs:0.911111111111111,-0.0298507462686567,0) {ITTI};
\node[left, inner sep=0mm, rotate=90, text=black]
at (rel axis cs:0.953535353535354,-0.0298507462686567,0) {CovSal};
\end{axis}
\end{tikzpicture}%
 & 
%
\begin{tikzpicture}

\begin{axis}[%
width=\figurewidth,
height=\figureheight,
clip=false,
scale only axis,
xmin=1,
xmax=25,
xtick={1,2,3,4,5,6,7,8,9,10,11,12,13,14,15,16,17,18,19,20,21,22,23,24},
xticklabels={\empty},
xmajorgrids,
ymin=0.322303702521356,
ymax=0.943909334486764,
ylabel={SJSD},
ymajorgrids,
axis x line*=bottom,
axis y line*=left,
ytick = {0.4,0.5,0.6,0.7,0.8,0.9},ylabel style={font=\tiny},legend style={font=\tiny},
]
\addplot [color=red,solid,line width=1.0pt,mark=asterisk,mark options={solid},forget plot]
  table[row sep=crcr]{1	0.376567158038809\\
2	0.70777198746527\\
3	0.726086968408237\\
4	0.688778585077025\\
5	0.750136458357169\\
6	0.783049485762873\\
7	0.776894192126436\\
8	0.778946934102719\\
9	0.77444040799315\\
10	0.795242454269678\\
11	0.776699037119066\\
12	0.777842281682321\\
13	0.801232921181621\\
14	0.796863085115438\\
15	0.804086111037029\\
16	0.797033917818979\\
17	0.80711058126572\\
18	0.799203120261879\\
19	0.790590803630612\\
20	0.771952847627658\\
21	0.808472147683178\\
22	0.82245111509228\\
23	0.857741078221474\\
24	0.85723788130642\\
};
\addplot [color=green,solid,line width=1.0pt,mark=+,mark options={solid},forget plot]
  table[row sep=crcr]{1	0.385524540169754\\
2	0.709453099381676\\
3	0.722070257319314\\
4	0.777592355438109\\
5	0.755296587221841\\
6	0.769957123406166\\
7	0.771275439334514\\
8	0.769169657040729\\
9	0.779885064615108\\
10	0.779768774423102\\
11	0.783625494586538\\
12	0.789892063642778\\
13	0.789690411027826\\
14	0.796533238227577\\
15	0.798273790541511\\
16	0.802554736219444\\
17	0.798342682899502\\
18	0.799817810979387\\
19	0.809783608071963\\
20	0.820464158534615\\
21	0.805048176734266\\
22	0.835203679876767\\
23	0.84219420059361\\
24	0.846650336283354\\
};
\addplot [color=blue,solid,line width=1.0pt,mark=o,mark options={solid},forget plot]
  table[row sep=crcr]{1	0.385169216358831\\
2	0.693494854951312\\
3	0.721701475634349\\
4	0.77791957716731\\
5	0.760956642805332\\
6	0.75786709283355\\
7	0.766812458296758\\
8	0.771389710845957\\
9	0.776380628991218\\
10	0.765662596361063\\
11	0.78146828346468\\
12	0.794927659954639\\
13	0.785073750637669\\
14	0.793786701026789\\
15	0.788536418163875\\
16	0.794475732283412\\
17	0.788769830531639\\
18	0.806244843464531\\
19	0.812691570578243\\
20	0.82535414583369\\
21	0.805840301387158\\
22	0.830595268181544\\
23	0.821868827246148\\
24	0.821080051636814\\
};
\addplot [color=black,dashed,line width=1.0pt,forget plot]
  table[row sep=crcr]{1	0.382420304855798\\
2	0.703573313932753\\
3	0.7232862337873\\
4	0.748096839227481\\
5	0.755463229461447\\
6	0.770291234000863\\
7	0.771660696585903\\
8	0.773168767329802\\
9	0.776902033866492\\
10	0.780224608351281\\
11	0.780597605056761\\
12	0.787554001759912\\
13	0.791999027615706\\
14	0.795727674789934\\
15	0.796965439914138\\
16	0.798021462107278\\
17	0.798074364898953\\
18	0.801755258235266\\
19	0.804355327426939\\
20	0.805923717331988\\
21	0.806453541934868\\
22	0.829416687716864\\
23	0.840601368687077\\
24	0.841656089742196\\
};
\node[left, inner sep=0mm, rotate=90, text=black]
at (rel axis cs:-0.00202020202020202,-0.0298507462686567,0) {ITTI2};
\node[left, inner sep=0mm, rotate=90, text=black]
at (rel axis cs:0.0383838383838384,-0.0298507462686567,0) {GR};
\node[left, inner sep=0mm, rotate=90, text=black]
at (rel axis cs:0.0808080808080808,-0.0298507462686567,0) {FTS};
\node[left, inner sep=0mm, rotate=90, text=black]
at (rel axis cs:0.121212121212121,-0.0298507462686567,0) {SpectRes};
\node[left, inner sep=0mm, rotate=90, text=black]
at (rel axis cs:0.163636363636364,-0.0298507462686567,0) {SUN};
\node[left, inner sep=0mm, rotate=90, text=black]
at (rel axis cs:0.204040404040404,-0.0298507462686567,0) {GAFFE};
\node[left, inner sep=0mm, rotate=90, text=black]
at (rel axis cs:0.246464646464646,-0.0298507462686567,0) {BMS};
\node[left, inner sep=0mm, rotate=90, text=black]
at (rel axis cs:0.288888888888889,-0.0298507462686567,0) {HouNIPS};
\node[left, inner sep=0mm, rotate=90, text=black]
at (rel axis cs:0.329292929292929,-0.0298507462686567,0) {GBVSNoCB};
\node[left, inner sep=0mm, rotate=90, text=black]
at (rel axis cs:0.371717171717172,-0.0298507462686567,0) {RandomCS};
\node[left, inner sep=0mm, rotate=90, text=black]
at (rel axis cs:0.412121212121212,-0.0298507462686567,0) {SigSal};
\node[left, inner sep=0mm, rotate=90, text=black]
at (rel axis cs:0.454545454545455,-0.0298507462686567,0) {SIMFine};
\node[left, inner sep=0mm, rotate=90, text=black]
at (rel axis cs:0.496969696969697,-0.0298507462686567,0) {SCIA};
\node[left, inner sep=0mm, rotate=90, text=black]
at (rel axis cs:0.537373737373737,-0.0298507462686567,0) {AIM};
\node[left, inner sep=0mm, rotate=90, text=black]
at (rel axis cs:0.57979797979798,-0.0298507462686567,0) {Judd};
\node[left, inner sep=0mm, rotate=90, text=black]
at (rel axis cs:0.62020202020202,-0.0298507462686567,0) {GBVS};
\node[left, inner sep=0mm, rotate=90, text=black]
at (rel axis cs:0.662626262626263,-0.0298507462686567,0) {Torralba};
\node[left, inner sep=0mm, rotate=90, text=black]
at (rel axis cs:0.703030303030303,-0.0298507462686567,0) {Context};
\node[left, inner sep=0mm, rotate=90, text=black]
at (rel axis cs:0.745454545454545,-0.0298507462686567,0) {SIMCoarse};
\node[left, inner sep=0mm, rotate=90, text=black]
at (rel axis cs:0.787878787878788,-0.0298507462686567,0) {SDSRG};
\node[left, inner sep=0mm, rotate=90, text=black]
at (rel axis cs:0.828282828282828,-0.0298507462686567,0) {AWS};
\node[left, inner sep=0mm, rotate=90, text=black]
at (rel axis cs:0.870707070707071,-0.0298507462686567,0) {SDSRL};
\node[left, inner sep=0mm, rotate=90, text=black]
at (rel axis cs:0.911111111111111,-0.0298507462686567,0) {CovSal};
\node[left, inner sep=0mm, rotate=90, text=black]
at (rel axis cs:0.953535353535354,-0.0298507462686567,0) {ITTI};
\end{axis}
\end{tikzpicture}%
 & 
%
\begin{tikzpicture}

\begin{axis}[%
width=\figurewidth,
height=\figureheight,
clip=false,
scale only axis,
xmin=1,
xmax=25,
xtick={1,2,3,4,5,6,7,8,9,10,11,12,13,14,15,16,17,18,19,20,21,22,23,24},
xticklabels={\empty},
xmajorgrids,
ymin=0.322303702521356,
ymax=0.943909334486764,
ylabel={SJSD},
ymajorgrids,
axis x line*=bottom,
axis y line*=left,
ytick = {0.4,0.5,0.6,0.7,0.8,0.9},ylabel style={font=\tiny},legend style={font=\tiny},
]
\addplot [color=red,solid,line width=1.0pt,mark=asterisk,mark options={solid},forget plot]
  table[row sep=crcr]{1	0.3621926668674\\
2	0.663759382065568\\
3	0.685594893395591\\
4	0.644823709781778\\
5	0.68343535667439\\
6	0.733781413877291\\
7	0.749225789443385\\
8	0.746239108390817\\
9	0.741899030904439\\
10	0.759651103854142\\
11	0.747949072019138\\
12	0.755667950519346\\
13	0.755341360731493\\
14	0.73487435547157\\
15	0.766699963540578\\
16	0.74772461297853\\
17	0.746929386532415\\
18	0.772565666201761\\
19	0.780669031654393\\
20	0.782280700350778\\
21	0.789500674499115\\
22	0.773298241022976\\
23	0.840026351854012\\
24	0.850503613179394\\
};
\addplot [color=green,solid,line width=1.0pt,mark=+,mark options={solid},forget plot]
  table[row sep=crcr]{1	0.358115225023729\\
2	0.654858019654469\\
3	0.683521611469316\\
4	0.752248595824769\\
5	0.748378193900135\\
6	0.726952798591042\\
7	0.735045113901767\\
8	0.742723491999456\\
9	0.748715406662282\\
10	0.745522618446739\\
11	0.746379090124976\\
12	0.751399951561542\\
13	0.752850042574753\\
14	0.767512910306404\\
15	0.759854061856385\\
16	0.76477446660589\\
17	0.772096064564688\\
18	0.772077130661636\\
19	0.777394364288325\\
20	0.776675009978954\\
21	0.783439466698926\\
22	0.796639188563188\\
23	0.822446064937368\\
24	0.83469764861629\\
};
\addplot [color=blue,solid,line width=1.0pt,mark=o,mark options={solid},forget plot]
  table[row sep=crcr]{1	0.36142254164041\\
2	0.665079419376614\\
3	0.686368892812208\\
4	0.75652969952545\\
5	0.770359540618603\\
6	0.751171082033155\\
7	0.738295981319047\\
8	0.743872634149528\\
9	0.754012226508265\\
10	0.742795884579407\\
11	0.754184972869204\\
12	0.748068186895614\\
13	0.759101702130955\\
14	0.782809346648463\\
15	0.75964390351879\\
16	0.776113237991517\\
17	0.793821430107284\\
18	0.784815838335674\\
19	0.774683643872521\\
20	0.783316050482231\\
21	0.778579396793151\\
22	0.803470136321091\\
23	0.799017208549222\\
24	0.827015008892479\\
};
\addplot [color=black,dashed,line width=1.0pt,forget plot]
  table[row sep=crcr]{1	0.36057681117718\\
2	0.661232273698884\\
3	0.685161799225705\\
4	0.717867335043999\\
5	0.734057697064376\\
6	0.737301764833829\\
7	0.7408556282214\\
8	0.744278411513267\\
9	0.748208888024995\\
10	0.74932320229343\\
11	0.749504378337773\\
12	0.751712029658834\\
13	0.755764368479067\\
14	0.761732204142146\\
15	0.762065976305251\\
16	0.762870772525312\\
17	0.770948960401462\\
18	0.776486211733024\\
19	0.77758234660508\\
20	0.780757253603988\\
21	0.783839845997064\\
22	0.791135855302418\\
23	0.820496541780201\\
24	0.837405423562721\\
};
\node[left, inner sep=0mm, rotate=90, text=black]
at (rel axis cs:-0.00202020202020202,-0.0298507462686567,0) {ITTI2};
\node[left, inner sep=0mm, rotate=90, text=black]
at (rel axis cs:0.0383838383838384,-0.0298507462686567,0) {FTS};
\node[left, inner sep=0mm, rotate=90, text=black]
at (rel axis cs:0.0808080808080808,-0.0298507462686567,0) {GR};
\node[left, inner sep=0mm, rotate=90, text=black]
at (rel axis cs:0.121212121212121,-0.0298507462686567,0) {SpectRes};
\node[left, inner sep=0mm, rotate=90, text=black]
at (rel axis cs:0.163636363636364,-0.0298507462686567,0) {SDSRG};
\node[left, inner sep=0mm, rotate=90, text=black]
at (rel axis cs:0.204040404040404,-0.0298507462686567,0) {SUN};
\node[left, inner sep=0mm, rotate=90, text=black]
at (rel axis cs:0.246464646464646,-0.0298507462686567,0) {BMS};
\node[left, inner sep=0mm, rotate=90, text=black]
at (rel axis cs:0.288888888888889,-0.0298507462686567,0) {GBVSNoCB};
\node[left, inner sep=0mm, rotate=90, text=black]
at (rel axis cs:0.329292929292929,-0.0298507462686567,0) {SigSal};
\node[left, inner sep=0mm, rotate=90, text=black]
at (rel axis cs:0.371717171717172,-0.0298507462686567,0) {RandomCS};
\node[left, inner sep=0mm, rotate=90, text=black]
at (rel axis cs:0.412121212121212,-0.0298507462686567,0) {HouNIPS};
\node[left, inner sep=0mm, rotate=90, text=black]
at (rel axis cs:0.454545454545455,-0.0298507462686567,0) {GAFFE};
\node[left, inner sep=0mm, rotate=90, text=black]
at (rel axis cs:0.496969696969697,-0.0298507462686567,0) {SCIA};
\node[left, inner sep=0mm, rotate=90, text=black]
at (rel axis cs:0.537373737373737,-0.0298507462686567,0) {SIMFine};
\node[left, inner sep=0mm, rotate=90, text=black]
at (rel axis cs:0.57979797979798,-0.0298507462686567,0) {GBVS};
\node[left, inner sep=0mm, rotate=90, text=black]
at (rel axis cs:0.62020202020202,-0.0298507462686567,0) {Context};
\node[left, inner sep=0mm, rotate=90, text=black]
at (rel axis cs:0.662626262626263,-0.0298507462686567,0) {SIMCoarse};
\node[left, inner sep=0mm, rotate=90, text=black]
at (rel axis cs:0.703030303030303,-0.0298507462686567,0) {AIM};
\node[left, inner sep=0mm, rotate=90, text=black]
at (rel axis cs:0.745454545454545,-0.0298507462686567,0) {Torralba};
\node[left, inner sep=0mm, rotate=90, text=black]
at (rel axis cs:0.787878787878788,-0.0298507462686567,0) {AWS};
\node[left, inner sep=0mm, rotate=90, text=black]
at (rel axis cs:0.828282828282828,-0.0298507462686567,0) {Judd};
\node[left, inner sep=0mm, rotate=90, text=black]
at (rel axis cs:0.870707070707071,-0.0298507462686567,0) {SDSRL};
\node[left, inner sep=0mm, rotate=90, text=black]
at (rel axis cs:0.911111111111111,-0.0298507462686567,0) {ITTI};
\node[left, inner sep=0mm, rotate=90, text=black]
at (rel axis cs:0.953535353535354,-0.0298507462686567,0) {CovSal};
\end{axis}
\end{tikzpicture}%
\\
%
\begin{tikzpicture}

\begin{axis}[%
width=\figurewidth,
height=\figureheight,
clip=false,
scale only axis,
xmin=1,
xmax=25,
xtick={1,2,3,4,5,6,7,8,9,10,11,12,13,14,15,16,17,18,19,20,21,22,23,24},
xticklabels={\empty},
xmajorgrids,
ymin=0.417914900919072,
ymax=2.08395532448465,
ylabel={SEMD},
ymajorgrids,
axis x line*=bottom,
axis y line*=left,
ytick = {0.5,1  ,1.5,2  },ylabel style={font=\tiny},legend style={font=\tiny},
]
\addplot [color=red,solid,line width=1.0pt,mark=asterisk,mark options={solid},forget plot]
  table[row sep=crcr]{1	0.572537826975192\\
2	1.41509252309181\\
3	1.49125350662292\\
4	1.46569212051597\\
5	1.6600419322325\\
6	1.48913934751206\\
7	1.66057333434726\\
8	1.68961802953877\\
9	1.67703715112857\\
10	1.69752278580971\\
11	1.66383104434123\\
12	1.68845386491442\\
13	1.67440977995182\\
14	1.69244190037776\\
15	1.69066284702332\\
16	1.73537884961727\\
17	1.62938545737113\\
18	1.74593190253601\\
19	1.7548524419598\\
20	1.74445652480357\\
21	1.75343482867115\\
22	1.76534232863602\\
23	1.8945048404406\\
24	1.89242863941264\\
};
\addplot [color=green,solid,line width=1.0pt,mark=+,mark options={solid},forget plot]
  table[row sep=crcr]{1	0.597460626798068\\
2	1.44174125149367\\
3	1.48224527768881\\
4	1.50731920369735\\
5	1.595032882491\\
6	1.69099829020061\\
7	1.63415425734147\\
8	1.63073668409407\\
9	1.6516655998896\\
10	1.67071288041901\\
11	1.65806824797594\\
12	1.68306846043113\\
13	1.68870704783096\\
14	1.70070864275596\\
15	1.71495641500298\\
16	1.72353305874851\\
17	1.7637805975353\\
18	1.75033160839918\\
19	1.74103100508823\\
20	1.75521033696935\\
21	1.81083969701472\\
22	1.796390947846\\
23	1.83677012875166\\
24	1.85409353212684\\
};
\addplot [color=blue,solid,line width=1.0pt,mark=o,mark options={solid},forget plot]
  table[row sep=crcr]{1	0.612644333695124\\
2	1.41392363676148\\
3	1.4838607771707\\
4	1.52510433867445\\
5	1.57273647840231\\
6	1.71321702054082\\
7	1.64729769392979\\
8	1.62951112258715\\
9	1.6275030981691\\
10	1.62367672276979\\
11	1.6809164763463\\
12	1.67856622547503\\
13	1.70375139476279\\
14	1.69742951508949\\
15	1.72810947955838\\
16	1.70875328279713\\
17	1.80471630247582\\
18	1.72838195335685\\
19	1.74969718388997\\
20	1.75774812466116\\
21	1.81189605966342\\
22	1.81707259190425\\
23	1.79394689011724\\
24	1.79966795090163\\
};
\addplot [color=black,dashed,line width=1.0pt,forget plot]
  table[row sep=crcr]{1	0.594214262489461\\
2	1.42358580378232\\
3	1.48578652049415\\
4	1.49937188762926\\
5	1.60927043104193\\
6	1.63111821941783\\
7	1.64734176187284\\
8	1.64995527874\\
9	1.65206861639575\\
10	1.66397079633284\\
11	1.66760525622116\\
12	1.68336285027353\\
13	1.68895607418186\\
14	1.69686001940774\\
15	1.71124291386156\\
16	1.72255506372097\\
17	1.73262745246075\\
18	1.74154848809735\\
19	1.74852687697933\\
20	1.75247166214469\\
21	1.79205686178309\\
22	1.79293528946209\\
23	1.84174061976983\\
24	1.8487300408137\\
};
\node[left, inner sep=0mm, rotate=90, text=black]
at (rel axis cs:-0.00404040404040404,-0.029673590504451,0) {ITTI2};
\node[left, inner sep=0mm, rotate=90, text=black]
at (rel axis cs:0.0383838383838384,-0.029673590504451,0) {GR};
\node[left, inner sep=0mm, rotate=90, text=black]
at (rel axis cs:0.0808080808080808,-0.029673590504451,0) {FTS};
\node[left, inner sep=0mm, rotate=90, text=black]
at (rel axis cs:0.121212121212121,-0.029673590504451,0) {SUN};
\node[left, inner sep=0mm, rotate=90, text=black]
at (rel axis cs:0.163636363636364,-0.029673590504451,0) {GAFFE};
\node[left, inner sep=0mm, rotate=90, text=black]
at (rel axis cs:0.204040404040404,-0.029673590504451,0) {SpectRes};
\node[left, inner sep=0mm, rotate=90, text=black]
at (rel axis cs:0.246464646464646,-0.029673590504451,0) {AIM};
\node[left, inner sep=0mm, rotate=90, text=black]
at (rel axis cs:0.288888888888889,-0.029673590504451,0) {Torralba};
\node[left, inner sep=0mm, rotate=90, text=black]
at (rel axis cs:0.329292929292929,-0.029673590504451,0) {RandomCS};
\node[left, inner sep=0mm, rotate=90, text=black]
at (rel axis cs:0.371717171717172,-0.029673590504451,0) {Judd};
\node[left, inner sep=0mm, rotate=90, text=black]
at (rel axis cs:0.412121212121212,-0.029673590504451,0) {HouNIPS};
\node[left, inner sep=0mm, rotate=90, text=black]
at (rel axis cs:0.454545454545455,-0.029673590504451,0) {GBVSNoCB};
\node[left, inner sep=0mm, rotate=90, text=black]
at (rel axis cs:0.496969696969697,-0.029673590504451,0) {SigSal};
\node[left, inner sep=0mm, rotate=90, text=black]
at (rel axis cs:0.537373737373737,-0.029673590504451,0) {SIMFine};
\node[left, inner sep=0mm, rotate=90, text=black]
at (rel axis cs:0.57979797979798,-0.029673590504451,0) {BMS};
\node[left, inner sep=0mm, rotate=90, text=black]
at (rel axis cs:0.62020202020202,-0.029673590504451,0) {GBVS};
\node[left, inner sep=0mm, rotate=90, text=black]
at (rel axis cs:0.662626262626263,-0.029673590504451,0) {SDSRG};
\node[left, inner sep=0mm, rotate=90, text=black]
at (rel axis cs:0.703030303030303,-0.029673590504451,0) {SCIA};
\node[left, inner sep=0mm, rotate=90, text=black]
at (rel axis cs:0.745454545454545,-0.029673590504451,0) {Context};
\node[left, inner sep=0mm, rotate=90, text=black]
at (rel axis cs:0.787878787878788,-0.029673590504451,0) {SIMCoarse};
\node[left, inner sep=0mm, rotate=90, text=black]
at (rel axis cs:0.828282828282828,-0.029673590504451,0) {SDSRL};
\node[left, inner sep=0mm, rotate=90, text=black]
at (rel axis cs:0.870707070707071,-0.029673590504451,0) {AWS};
\node[left, inner sep=0mm, rotate=90, text=black]
at (rel axis cs:0.911111111111111,-0.029673590504451,0) {CovSal};
\node[left, inner sep=0mm, rotate=90, text=black]
at (rel axis cs:0.953535353535354,-0.029673590504451,0) {ITTI};
\end{axis}
\end{tikzpicture}%
 & 
%
\begin{tikzpicture}

\begin{axis}[%
width=\figurewidth,
height=\figureheight,
clip=false,
scale only axis,
xmin=1,
xmax=25,
xtick={1,2,3,4,5,6,7,8,9,10,11,12,13,14,15,16,17,18,19,20,21,22,23,24},
xticklabels={\empty},
xmajorgrids,
ymin=0.417914900919072,
ymax=2.08395532448465,
ylabel={SEMD},
ymajorgrids,
axis x line*=bottom,
axis y line*=left,
ytick = {0.5,1  ,1.5,2  },ylabel style={font=\tiny},legend style={font=\tiny},
]
\addplot [color=red,solid,line width=1.0pt,mark=asterisk,mark options={solid},forget plot]
  table[row sep=crcr]{1	0.509428540305973\\
2	1.37462848248881\\
3	1.45769987482106\\
4	1.43935450518798\\
5	1.61055647773732\\
6	1.40252801478712\\
7	1.66172936489469\\
8	1.61691349261609\\
9	1.65444079973996\\
10	1.67489304326805\\
11	1.66316348254504\\
12	1.63694681551157\\
13	1.65672512265263\\
14	1.67258873743109\\
15	1.67317336637488\\
16	1.70199020898376\\
17	1.71250951577114\\
18	1.7692746742307\\
19	1.76728629775571\\
20	1.7250899622144\\
21	1.67329907807458\\
22	1.78439058948619\\
23	1.88109949573394\\
24	1.88975969794697\\
};
\addplot [color=green,solid,line width=1.0pt,mark=+,mark options={solid},forget plot]
  table[row sep=crcr]{1	0.544420643253773\\
2	1.38698283847521\\
3	1.44052476180686\\
4	1.48219905524833\\
5	1.56745804604537\\
6	1.65623858706233\\
7	1.60025060350176\\
8	1.62608605620655\\
9	1.62647404739706\\
10	1.63491785706299\\
11	1.62284195670853\\
12	1.64708897277884\\
13	1.67421172891134\\
14	1.67272682982153\\
15	1.69639191386552\\
16	1.71652101578565\\
17	1.70832804526792\\
18	1.72771242911999\\
19	1.75418400790931\\
20	1.76241186731671\\
21	1.81043435506204\\
22	1.80124411408739\\
23	1.82195757246807\\
24	1.84298612649773\\
};
\addplot [color=blue,solid,line width=1.0pt,mark=o,mark options={solid},forget plot]
  table[row sep=crcr]{1	0.533363724520334\\
2	1.36883995938472\\
3	1.45039016781061\\
4	1.49651487135607\\
5	1.53005718540386\\
6	1.66385378983092\\
7	1.56190287633485\\
8	1.62290648773325\\
9	1.61418178739881\\
10	1.59376549927084\\
11	1.6279627375617\\
12	1.6387859261611\\
13	1.67319365910744\\
14	1.6603653669382\\
15	1.72099519015369\\
16	1.69102148518648\\
17	1.72507285352874\\
18	1.68913506780055\\
19	1.74822307761643\\
20	1.78239409182408\\
21	1.81843040307523\\
22	1.78318305102219\\
23	1.74689728489227\\
24	1.76354179665708\\
};
\addplot [color=black,dashed,line width=1.0pt,forget plot]
  table[row sep=crcr]{1	0.529070969360027\\
2	1.37681709344958\\
3	1.44953826814617\\
4	1.47268947726413\\
5	1.56935723639552\\
6	1.57420679722679\\
7	1.60796094824377\\
8	1.62196867885196\\
9	1.63169887817861\\
10	1.63452546653396\\
11	1.63798939227176\\
12	1.64094057148384\\
13	1.66804350355713\\
14	1.66856031139694\\
15	1.69685349013136\\
16	1.7031775699853\\
17	1.7153034715226\\
18	1.72870739038375\\
19	1.75656446109382\\
20	1.75663197378506\\
21	1.76738794540395\\
22	1.78960591819859\\
23	1.81665145103143\\
24	1.83209587370059\\
};
\node[left, inner sep=0mm, rotate=90, text=black]
at (rel axis cs:-0.00404040404040404,-0.029673590504451,0) {ITTI2};
\node[left, inner sep=0mm, rotate=90, text=black]
at (rel axis cs:0.0383838383838384,-0.029673590504451,0) {GR};
\node[left, inner sep=0mm, rotate=90, text=black]
at (rel axis cs:0.0808080808080808,-0.029673590504451,0) {FTS};
\node[left, inner sep=0mm, rotate=90, text=black]
at (rel axis cs:0.121212121212121,-0.029673590504451,0) {SUN};
\node[left, inner sep=0mm, rotate=90, text=black]
at (rel axis cs:0.163636363636364,-0.029673590504451,0) {GAFFE};
\node[left, inner sep=0mm, rotate=90, text=black]
at (rel axis cs:0.204040404040404,-0.029673590504451,0) {SpectRes};
\node[left, inner sep=0mm, rotate=90, text=black]
at (rel axis cs:0.246464646464646,-0.029673590504451,0) {RandomCS};
\node[left, inner sep=0mm, rotate=90, text=black]
at (rel axis cs:0.288888888888889,-0.029673590504451,0) {AIM};
\node[left, inner sep=0mm, rotate=90, text=black]
at (rel axis cs:0.329292929292929,-0.029673590504451,0) {Torralba};
\node[left, inner sep=0mm, rotate=90, text=black]
at (rel axis cs:0.371717171717172,-0.029673590504451,0) {Judd};
\node[left, inner sep=0mm, rotate=90, text=black]
at (rel axis cs:0.412121212121212,-0.029673590504451,0) {HouNIPS};
\node[left, inner sep=0mm, rotate=90, text=black]
at (rel axis cs:0.454545454545455,-0.029673590504451,0) {GBVSNoCB};
\node[left, inner sep=0mm, rotate=90, text=black]
at (rel axis cs:0.496969696969697,-0.029673590504451,0) {SigSal};
\node[left, inner sep=0mm, rotate=90, text=black]
at (rel axis cs:0.537373737373737,-0.029673590504451,0) {BMS};
\node[left, inner sep=0mm, rotate=90, text=black]
at (rel axis cs:0.57979797979798,-0.029673590504451,0) {SIMFine};
\node[left, inner sep=0mm, rotate=90, text=black]
at (rel axis cs:0.62020202020202,-0.029673590504451,0) {GBVS};
\node[left, inner sep=0mm, rotate=90, text=black]
at (rel axis cs:0.662626262626263,-0.029673590504451,0) {Context};
\node[left, inner sep=0mm, rotate=90, text=black]
at (rel axis cs:0.703030303030303,-0.029673590504451,0) {SCIA};
\node[left, inner sep=0mm, rotate=90, text=black]
at (rel axis cs:0.745454545454545,-0.029673590504451,0) {AWS};
\node[left, inner sep=0mm, rotate=90, text=black]
at (rel axis cs:0.787878787878788,-0.029673590504451,0) {SIMCoarse};
\node[left, inner sep=0mm, rotate=90, text=black]
at (rel axis cs:0.828282828282828,-0.029673590504451,0) {SDSRG};
\node[left, inner sep=0mm, rotate=90, text=black]
at (rel axis cs:0.870707070707071,-0.029673590504451,0) {SDSRL};
\node[left, inner sep=0mm, rotate=90, text=black]
at (rel axis cs:0.911111111111111,-0.029673590504451,0) {CovSal};
\node[left, inner sep=0mm, rotate=90, text=black]
at (rel axis cs:0.953535353535354,-0.029673590504451,0) {ITTI};
\end{axis}
\end{tikzpicture}%
 & 
%
\begin{tikzpicture}

\begin{axis}[%
width=\figurewidth,
height=\figureheight,
clip=false,
scale only axis,
xmin=1,
xmax=25,
xtick={1,2,3,4,5,6,7,8,9,10,11,12,13,14,15,16,17,18,19,20,21,22,23,24},
xticklabels={\empty},
xmajorgrids,
ymin=0.417914900919072,
ymax=2.08395532448465,
ylabel={SEMD},
ymajorgrids,
axis x line*=bottom,
axis y line*=left,
ytick = {0.5,1  ,1.5,2  },ylabel style={font=\tiny},legend style={font=\tiny},
]
\addplot [color=red,solid,line width=1.0pt,mark=asterisk,mark options={solid},forget plot]
  table[row sep=crcr]{1	0.485576783577349\\
2	1.28850339642583\\
3	1.31224958379333\\
4	1.38526260430279\\
5	1.28763262560914\\
6	1.54263420416775\\
7	1.52243699837527\\
8	1.55073160480261\\
9	1.52413231826779\\
10	1.52814417676297\\
11	1.54552154305633\\
12	1.58059692905412\\
13	1.56703614338328\\
14	1.43368388037104\\
15	1.54702834223505\\
16	1.6073415002016\\
17	1.599654099194\\
18	1.61835191724595\\
19	1.53280353471266\\
20	1.5724607819275\\
21	1.65404412085171\\
22	1.61259009363619\\
23	1.82734896919375\\
24	1.83791218226832\\
};
\addplot [color=green,solid,line width=1.0pt,mark=+,mark options={solid},forget plot]
  table[row sep=crcr]{1	0.46434988991008\\
2	1.24871359847036\\
3	1.31078887232856\\
4	1.38815713364627\\
5	1.56328972080317\\
6	1.48994415002592\\
7	1.49580971283325\\
8	1.51158130781837\\
9	1.53919173216774\\
10	1.55479515760122\\
11	1.54862317052112\\
12	1.53188951424877\\
13	1.55904021205306\\
14	1.60748863021597\\
15	1.58360538616381\\
16	1.57283240411449\\
17	1.571650928283\\
18	1.58534194615427\\
19	1.59952471807944\\
20	1.63120454364642\\
21	1.63900107307002\\
22	1.67172754007231\\
23	1.75673333122641\\
24	1.7715229997924\\
};
\addplot [color=blue,solid,line width=1.0pt,mark=o,mark options={solid},forget plot]
  table[row sep=crcr]{1	0.485938701078487\\
2	1.27668231820851\\
3	1.3202059669048\\
4	1.46119507828095\\
5	1.57040382758561\\
6	1.46899237469879\\
7	1.49198742928192\\
8	1.52994845049555\\
9	1.57429845519596\\
10	1.55639686419194\\
11	1.5716454036503\\
12	1.56121218534039\\
13	1.55298054979329\\
14	1.64802775772253\\
15	1.61099923744271\\
16	1.56645722042389\\
17	1.58862574844906\\
18	1.56055472919166\\
19	1.65437312216231\\
20	1.69468030725216\\
21	1.64910042193797\\
22	1.69502277577325\\
23	1.67988392341135\\
24	1.74627803799237\\
};
\addplot [color=black,dashed,line width=1.0pt,forget plot]
  table[row sep=crcr]{1	0.478621791521972\\
2	1.2712997710349\\
3	1.31441480767556\\
4	1.41153827207667\\
5	1.47377539133264\\
6	1.50052357629749\\
7	1.50341138016348\\
8	1.53075378770551\\
9	1.54587416854383\\
10	1.54644539951871\\
11	1.55526337240925\\
12	1.5578995428811\\
13	1.55968563507654\\
14	1.56306675610318\\
15	1.58054432194719\\
16	1.58221037491333\\
17	1.58664359197536\\
18	1.58808286419729\\
19	1.5955671249848\\
20	1.6327818776087\\
21	1.64738187195323\\
22	1.65978013649392\\
23	1.75465540794384\\
24	1.7852377400177\\
};
\node[left, inner sep=0mm, rotate=90, text=black]
at (rel axis cs:-0.00404040404040404,-0.029673590504451,0) {ITTI2};
\node[left, inner sep=0mm, rotate=90, text=black]
at (rel axis cs:0.0383838383838384,-0.029673590504451,0) {FTS};
\node[left, inner sep=0mm, rotate=90, text=black]
at (rel axis cs:0.0808080808080808,-0.029673590504451,0) {GR};
\node[left, inner sep=0mm, rotate=90, text=black]
at (rel axis cs:0.121212121212121,-0.029673590504451,0) {SUN};
\node[left, inner sep=0mm, rotate=90, text=black]
at (rel axis cs:0.163636363636364,-0.029673590504451,0) {SpectRes};
\node[left, inner sep=0mm, rotate=90, text=black]
at (rel axis cs:0.204040404040404,-0.029673590504451,0) {RandomCS};
\node[left, inner sep=0mm, rotate=90, text=black]
at (rel axis cs:0.246464646464646,-0.029673590504451,0) {GAFFE};
\node[left, inner sep=0mm, rotate=90, text=black]
at (rel axis cs:0.288888888888889,-0.029673590504451,0) {GBVSNoCB};
\node[left, inner sep=0mm, rotate=90, text=black]
at (rel axis cs:0.329292929292929,-0.029673590504451,0) {AIM};
\node[left, inner sep=0mm, rotate=90, text=black]
at (rel axis cs:0.371717171717172,-0.029673590504451,0) {SigSal};
\node[left, inner sep=0mm, rotate=90, text=black]
at (rel axis cs:0.412121212121212,-0.029673590504451,0) {HouNIPS};
\node[left, inner sep=0mm, rotate=90, text=black]
at (rel axis cs:0.454545454545455,-0.029673590504451,0) {BMS};
\node[left, inner sep=0mm, rotate=90, text=black]
at (rel axis cs:0.496969696969697,-0.029673590504451,0) {Torralba};
\node[left, inner sep=0mm, rotate=90, text=black]
at (rel axis cs:0.537373737373737,-0.029673590504451,0) {SDSRG};
\node[left, inner sep=0mm, rotate=90, text=black]
at (rel axis cs:0.57979797979798,-0.029673590504451,0) {Context};
\node[left, inner sep=0mm, rotate=90, text=black]
at (rel axis cs:0.62020202020202,-0.029673590504451,0) {GBVS};
\node[left, inner sep=0mm, rotate=90, text=black]
at (rel axis cs:0.662626262626263,-0.029673590504451,0) {SCIA};
\node[left, inner sep=0mm, rotate=90, text=black]
at (rel axis cs:0.703030303030303,-0.029673590504451,0) {Judd};
\node[left, inner sep=0mm, rotate=90, text=black]
at (rel axis cs:0.745454545454545,-0.029673590504451,0) {SIMFine};
\node[left, inner sep=0mm, rotate=90, text=black]
at (rel axis cs:0.787878787878788,-0.029673590504451,0) {SIMCoarse};
\node[left, inner sep=0mm, rotate=90, text=black]
at (rel axis cs:0.828282828282828,-0.029673590504451,0) {AWS};
\node[left, inner sep=0mm, rotate=90, text=black]
at (rel axis cs:0.870707070707071,-0.029673590504451,0) {SDSRL};
\node[left, inner sep=0mm, rotate=90, text=black]
at (rel axis cs:0.911111111111111,-0.029673590504451,0) {ITTI};
\node[left, inner sep=0mm, rotate=90, text=black]
at (rel axis cs:0.953535353535354,-0.029673590504451,0) {CovSal};
\end{axis}
\end{tikzpicture}%
 \\
		(a) Easy & (b) Medium & (c) Hard     
	\end{tabular}
	\caption{SAUC, SNSS, SSKLD, SJSD, and SEMD scores for the Judd Low Resolution database for different image complexity categories:(a) Easy, (b) Medium and (c) Hard. Blue, green and red lines correspond to low, medium and high levels of blur, respectively, and the black dotted line represents the average over all distortion levels according to which the models have been sorted.}
	\label{fig:MetricScores_JLR}
\end{figure*}
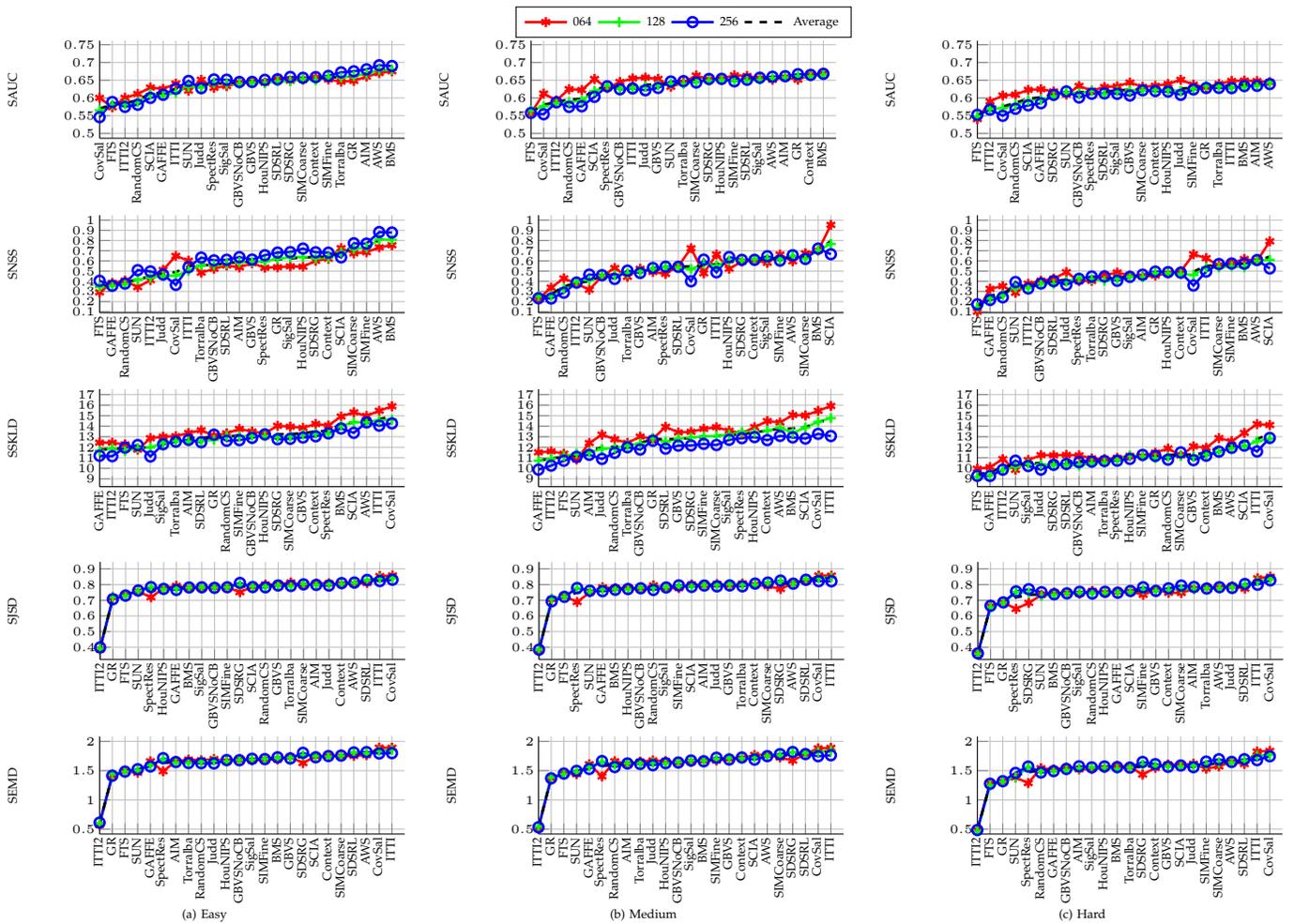

The Judd low resolution database has only one type of distortion, i.e., blur, but it has 8 different levels of distortions that correspond to the 8 different resolutions (512 to 4 in factors of 2). The images are also divided into three categories: easy, medium and hard, based on their complexity.
We report the variation in performance for the three different image categories for only 3 out of the 8 resolutions for ease of presentation. The 3 resolutions chosen are 64, 128 and 256 which correspond to high, medium and low levels of blur, respectively.  
Figure~\ref{fig:MetricScores_JLR} shows the variation in performance in terms of the SAUC, SNSS, SSKLD, SJSD and SEMD metrics, respectively, for the 3 different distortion levels.  From these figures we observe the following: \\
- For the SSKLD and SJSD metrics, most of the models perform better for high and medium levels of distortions than that for the low or near-zero distortion case over all the image categories. For the SAUC metric, the performance at the high and medium levels of  distortions is greater than that for the low level  for most models over the hard and medium image categories. For the SEMD and SNSS metrics, no trend in scores is observed. \\
- For all the metrics, the average scores over all blur levels over all the models decrease with increase in image complexity.  This is seen in the fall of the curve representing average scores (marked by a black dotted line in Figure~\ref{fig:MetricScores_JLR}) as the image complexity goes from easy to hard. This is to be expected as increase in image complexity increases the clutter in the scene thus making it harder for the VA models to distinguish the salient locations from the non-salient background. 

\section{Conclusion} 
\label{sec:Conclusion}
In this paper, noticeable shortcomings of existing VA performance evaluation metrics are studied and four new performance metrics that remedy these shortcomings are proposed. Also, a comprehensive analysis of several state-of-the-art VA models is performed over two databases with different distortion types and several levels of distortions in addition to the distortion-free case. Along with the widely used SAUC metric, the performance is evaluated in terms of the four proposed shuffled metrics, SNSS, SSKLD, SJSD and SEMD. For the distortion-free case, it is shown that the proposed metrics achieve a consistent performance in terms of model rankings across different datasets. Furthermore, it is shown that the proposed SJSD metric is more robust than the other metrics, including the widely used SAUC metric, in the presence of various levels and types of distortions in the sense that it exhibits the most consistent performance across distortion levels and distortion types. It is found that the performance of the VA models varies with both the type and level of distortion.  Also, it is found that, in general,  high and medium levels of distortion improve the performance of the VA models relative to the low or near zero-distortion case. Finally, the VA models' performance is also found to decrease with an increase in image complexity. 


%









\bibliographystyle{IEEEtran}
\bibliography{refs_short}

\begin{thebibliography}{10}
\providecommand{\url}[1]{#1}
\csname url@samestyle\endcsname
\providecommand{\newblock}{\relax}
\providecommand{\bibinfo}[2]{#2}
\providecommand{\BIBentrySTDinterwordspacing}{\spaceskip=0pt\relax}
\providecommand{\BIBentryALTinterwordstretchfactor}{4}
\providecommand{\BIBentryALTinterwordspacing}{\spaceskip=\fontdimen2\font plus
\BIBentryALTinterwordstretchfactor\fontdimen3\font minus
  \fontdimen4\font\relax}
\providecommand{\BIBforeignlanguage}[2]{{%
\expandafter\ifx\csname l@#1\endcsname\relax
\typeout{** WARNING: IEEEtran.bst: No hyphenation pattern has been}%
\typeout{** loaded for the language `#1'. Using the pattern for}%
\typeout{** the default language instead.}%
\else
\language=\csname l@#1\endcsname
\fi
#2}}
\providecommand{\BIBdecl}{\relax}
\BIBdecl

\bibitem{borjireview}
A.~Borji and L.~Itti, ``State-of-the-art in visual attention modeling,''
  \emph{IEEE Trans. Pattern Analysis and Machine Intelligence}, vol.~35, no.~1,
  pp. 185--207, Jan 2013.

\bibitem{borjieval}
A.~Borji, D.~Sihite, and L.~Itti, ``Quantitative analysis of human-model
  agreement in visual saliency modeling: A comparative study,'' \emph{IEEE
  Trans. Image Processing}, vol.~22, no.~1, pp. 55--69, Jan 2013.

\bibitem{judd2012benchmark}
T.~Judd, F.~Durand, and A.~Torralba, ``A benchmark of computational models of
  saliency to predict human fixations,'' \emph{MIT Tech Report}, 2012.

\bibitem{obj_recog_VAapp1}
D.~Walther and C.~Koch, ``Modeling attention to salient proto-objects,''
  \emph{Neural networks}, vol.~19, no.~9, pp. 1395--1407, 2006.

\bibitem{obj_recog_VAapp2}
S.~Frintrop, \emph{VOCUS: A visual attention system for object detection and
  goal-directed search}.\hskip 1em plus 0.5em minus 0.4em\relax Springer, 2006.

\bibitem{segment_VA_application1}
A.~Mishra, Y.~Aloimonos, and C.~L. Fah, ``Active segmentation with fixation,''
  in \emph{12th IEEE Int'l Conf. Computer Vision}, Sept 2009, pp. 468--475.

\bibitem{segment_VA_application2}
A.~Maki, P.~Nordlund, and J.-O. Eklundh, ``Attentional scene segmentation:
  integrating depth and motion,'' \emph{Computer Vision and Image
  Understanding}, vol.~78, no.~3, pp. 351--373, 2000.

\bibitem{Wang2004}
Z.~Wang, A.~Bovik, H.~Sheikh, and E.~Simoncelli, ``Image quality assessment:
  from error visibility to structural similarity,'' \emph{IEEE Trans. Image
  Processing}, vol.~13, no.~4, pp. 600--612, April 2004.

\bibitem{Sheikh2006}
H.~Sheikh and A.~Bovik, ``Image information and visual quality,'' \emph{IEEE
  Trans. Image Processing}, vol.~15, no.~2, pp. 430--444, Feb 2006.

\bibitem{Wang2003}
Z.~Wang, E.~Simoncelli, and A.~Bovik, ``Multiscale structural similarity for
  image quality assessment,'' in \emph{the Thirty-Seventh Asilomar Conf.
  Signals, Systems and Computers}, vol.~2, Nov 2003, pp. 1398--1402.

\bibitem{Engelke2011}
U.~Engelke, H.~Kaprykowsky, H.-J. Zepernick, and P.~Ndjiki-Nya, ``Visual
  attention in quality assessment,'' \emph{IEEE Signal Processing Magazine},
  vol.~28, no.~6, pp. 50--59, Nov. 2011.

\bibitem{Ninassi2007}
A.~Ninassi, O.~Le~Meur, P.~Le~Callet, and D.~Barbba, ``Does where you gaze on
  an image affect your perception of quality? applying visual attention to
  image quality metric,'' in \emph{IEEE Int'l Conf. Image Processing}, vol.~2,
  Sept 2007, pp. 169--172.

\bibitem{Larson2008}
E.~Larson, C.~Vu, and D.~Chandler, ``Can visual fixation patterns improve image
  fidelity assessment?'' in \emph{15th IEEE Int'l Conf. Image Processing}, Oct
  2008, pp. 2572--2575.

\bibitem{HantaoLiu2011}
H.~Liu and I.~Heynderickx, ``Visual attention in objective image quality
  assessment: Based on eye-tracking data,'' \emph{{IEEE} Trans. Circuits and
  Systems for Video Technology}, vol.~21, no.~7, pp. 971--982, Jul. 2011.

\bibitem{Barland2006}
R.~Barland and A.~Saadane, ``Blind quality metric using a perceptual importance
  map for jpeg-20000 compressed images,'' \emph{IEEE Int'l Conf. Image
  Processing}, pp. 2941--2944, 2006.

\bibitem{Moorthy2009}
A.~K. Moorthy and A.~C. Bovik, ``\BIBforeignlanguage{English}{{Visual
  Importance Pooling for Image Quality Assessment}},''
  \emph{\BIBforeignlanguage{English}{IEEE J. Selected Topics in Signal
  Processing}}, vol.~3, no.~2, pp. 193--201, Apr. 2009.

\bibitem{Sadaka2008}
N.~G. Sadaka, L.~J. Karam, R.~Ferzli, and G.~P. Abousleman, ``A no-reference
  perceptual image sharpness metric based on saliency-weighted foveal
  pooling,'' in \emph{IEEE Int'l Conf. Image Processing}, Oct. 2008, pp.
  369--372.

\bibitem{Gkioulekas2010}
I.~Gkioulekas, G.~Evangelopoulos, and P.~Maragos, ``Spatial bayesian surprise
  for image saliency and quality assessment,'' in \emph{IEEE Int'l Conf. Image
  Processing}, Sep. 2010, pp. 1081--1084.

\bibitem{Moorthy}
A.~K. Moorthy and A.~C. Bovik, ``Visual importance pooling for image quality
  assessment,'' \emph{{IEEE} J. Selected Topics in Signal Processing}, vol.~3,
  no.~2, pp. 193--201, Apr. 2009.

\bibitem{Liu}
H.~Liu and I.~Heynderickx, ``Visual attention in objective image quality
  assessment: based on eye-tracking data,'' \emph{{IEEE} Trans. Circuits and
  Systems for Video Technology}, vol.~21, no.~7, pp. 971--982, Jul. 2011.

\bibitem{TUDInteractions}
J.~Redi, H.~Liu, R.~Zunino, and I.~Heynderickx, ``Interactions of visual
  attention and quality perception,'' \emph{Proc. {SPIE}, Human Vision and
  Electronic Imaging XVI}, vol. 7865, no.~1, pp. 1--11, Feb. 2011.

\bibitem{Vu2008}
C.~T. Vu, E.~C. Larson, and D.~M. Chandler, ``Visual fixation patterns when
  judging image quality: Effects of distortion type, amount, and subject
  experience,'' \emph{IEEE Southwest Symposium on Image Analysis and
  Interpretation, 2008}, pp. 73--76, Mar. 2008.

\bibitem{yarbus1967eye}
A.~L. Yarbus, ``Eye movements during perception of complex objects,'' in
  \emph{Eye Movements and Vision}.\hskip 1em plus 0.5em minus 0.4em\relax
  Springer, 1967, pp. 171--211.

\bibitem{NIPS2013_5196}
S.~Mathe and C.~Sminchisescu, ``Action from still image dataset and inverse
  optimal control to learn task specific visual scanpaths,'' in \emph{Advances
  in Neural Information Processing Systems}, C.~Burges, L.~Bottou, M.~Welling,
  Z.~Ghahramani, and K.~Weinberger, Eds., 2013, pp. 1923--1931.

\bibitem{GAFFE}
U.~Rajashekar, I.~van~der Linde, A.~C. Bovik, and L.~K. Cormack, ``{GAFFE:} a
  {Gaze-Attentive} fixation finding engine,'' \emph{{IEEE} Trans. Image
  Processing}, vol.~17, no.~4, pp. 564--573, Apr. 2008.

\bibitem{Itti}
L.~Itti, C.~Koch, and E.~Niebur, ``A model of saliency-based visual attention
  for rapid scene analysis,'' \emph{IEEE Trans. Pattern Analysis and Machine
  Intelligence}, vol.~20, no.~11, pp. 1254--1259, Nov 1998.

\bibitem{GBVS}
J.~Harel, C.~Koch, and P.~Perona, ``Graph-based visual saliency,''
  \emph{Advances in Neural Information Processing Systems}, vol.~19, pp.
  545--552, 2007.

\bibitem{AIM}
N.~Bruce and J.~Tsotsos, ``Saliency, attention, and visual search: An
  information theoretic approach,'' \emph{J. Vision}, vol.~9, no.~3, pp. 1--24,
  Mar. 2009.

\bibitem{HouNIPS}
X.~Hou and L.~Zhang, ``Dynamic visual attention: searching for coding length
  increments,'' in \emph{Advances in Neural Information Processing Systems},
  2008, pp. 681--688.

\bibitem{GR}
M.~Mancas, C.~Mancas-thillou, B.~Gosselin, and B.~Macq, ``A rarity-based visual
  attention map-application to texture description,'' in \emph{{IEEE} Int'l
  Conf. Image Processing}, Oct. 2006, pp. 445 --448.

\bibitem{SDSR}
H.~J. Seo and P.~Milanfar, ``\BIBforeignlanguage{en}{Static and space-time
  visual saliency detection by self-resemblance},''
  \emph{\BIBforeignlanguage{en}{J. Vision}}, vol.~9, no.~12, p.~15, Nov. 2009.

\bibitem{SUN}
L.~Zhang, M.~H. Tong, T.~K. Marks, H.~Shan, and G.~W. Cottrell, ``{SUN:} a
  bayesian framework for saliency using natural statistics,'' \emph{J. Vision},
  vol.~8, no.~7, pp. 1 -- 20, Dec. 2008.

\bibitem{AudeTorralba}
A.~Oliva and A.~Torralba, ``\BIBforeignlanguage{en}{Modeling the shape of the
  scene: A holistic representation of the spatial envelope},''
  \emph{\BIBforeignlanguage{en}{Int'l J. Computer Vision}}, vol.~42, no.~3, pp.
  145--175, May 2001.

\bibitem{MIT}
T.~Judd, K.~Ehinger, F.~Durand, and A.~Torralba,
  ``\BIBforeignlanguage{English}{Learning to predict where humans look},'' in
  \emph{\BIBforeignlanguage{English}{12th IEEE Int'l Conf. Computer Vision}},
  Oct. 2009, pp. 2106--2113.

\bibitem{FES}
H.~Tavakoli, E.~Rahtu, and J.~Heikkila, ``Fast and efficient saliency detection
  using sparse sampling and kernel density estimation,'' in \emph{Scandinavian
  Conf. Image Analysis}.\hskip 1em plus 0.5em minus 0.4em\relax Springer Berlin
  Heidelberg, 2011, pp. 666--675.

\bibitem{FTS}
R.~Achanta, S.~Hemami, F.~Estrada, and S.~Susstrunk, ``Frequency-tuned salient
  region detection,'' in \emph{{IEEE} Conf. Computer Vision and Pattern
  Recognition {(CVPR)}}, Jun. 2009, pp. 1597--1604.

\bibitem{SigSal}
X.~Hou, J.~Harel, and C.~Koch, ``Image signature: Highlighting sparse salient
  regions,'' \emph{IEEE Trans. Pattern Analysis and Machine Intelligence},
  vol.~34, pp. 194--201, Jul. 2011.

\bibitem{SpectRes}
X.~Hou and L.~Zhang, ``Saliency detection: A spectral residual approach,'' in
  \emph{{IEEE} Conf. Computer Vision and Pattern Recognition ({CVPR})}, 2007,
  pp. 1--8.

\bibitem{AWS}
A.~Garcia-Diaz, V.~Lebor\'{a}n, X.~R. Fdez-Vidal, and X.~M. Pardo,
  ``\BIBforeignlanguage{en}{{On the relationship between optical variability,
  visual saliency, and eye fixations: a computational approach}},''
  \emph{\BIBforeignlanguage{en}{J. Vision}}, vol.~12, no.~6, pp. 1--22, Jan.
  2012.

\bibitem{BMS}
J.~Zhang and S.~Sclaroff, ``Saliency detection: A boolean map approach,'' in
  \emph{Proc. of the IEEE Int'l Conf. Computer Vision (ICCV)}, Dec. 2013, pp.
  153--160.

\bibitem{Context}
S.~Goferman, L.~Zelnik-Manor, and A.~Tal, ``Context-aware saliency detection,''
  \emph{{IEEE} Trans. Pattern Analysis and Machine Intelligence}, vol.~34,
  no.~10, pp. 1915--1926, 2012.

\bibitem{CovSal}
E.~Erdem and A.~Erdem, ``\BIBforeignlanguage{en}{Visual saliency estimation by
  nonlinearly integrating features using region covariances},''
  \emph{\BIBforeignlanguage{en}{J. Vision}}, vol.~13, no.~4, p.~11, Mar. 2013,
  {PMID:} 23509407.

\bibitem{RandomCS}
T.~N. Vikram, M.~Tscherepanow, and B.~Wrede, ``A saliency map based on sampling
  an image into random rectangular regions of interest,'' \emph{Pattern
  Recognition}, vol.~45, no.~9, pp. 3114 -- 3124, 2012.

\bibitem{SIM}
R.~Margolin, L.~Zelnik-Manor, and A.~Tal, ``\BIBforeignlanguage{en}{Saliency
  for image manipulation},'' \emph{\BIBforeignlanguage{en}{The Visual
  Computer}}, vol.~29, no.~5, pp. 381--392, May 2013.

\bibitem{koch1987salmap}
C.~Koch and S.~Ullman, ``Shifts in selective visual attention: Towards the
  underlying neural circuitry,'' in \emph{Matters of Intelligence}.\hskip 1em
  plus 0.5em minus 0.4em\relax Springer, 1987, pp. 115--141.

\bibitem{treisman1980feature}
A.~M. Treisman and G.~Gelade, ``A feature-integration theory of attention,''
  \emph{Cognitive Psychology}, vol.~12, no.~1, pp. 97--136, 1980.

\bibitem{ShannonsInformationMeasure}
C.~E. Shannon, ``A mathematical theory of communication,'' \emph{ACM SIGMOBILE
  Mobile Computing and Communications Rev.}, vol.~5, no.~1, pp. 3--55, 2001.

\bibitem{torralba2003modeling}
A.~Torralba, ``Modeling global scene factors in attention,'' \emph{J. the
  Optical Society of America. A, Optics, Image Science, and Vision}, vol.~20,
  no.~7, pp. 1407--1418, 2003.

\bibitem{Judd_2012}
T.~Judd, F.~Durand, and A.~Torralba, ``A benchmark of computational models of
  saliency to predict human fixations,'' in \emph{MIT Technical Report}, 2012.

\bibitem{Rubner}
Y.~Rubner, C.~Tomasi, and L.~J. Guibas, ``The earth mover's distance as a
  metric for image retrieval,'' \emph{Int'l J. Computer Vision}, vol.~40,
  no.~2, pp. 99--121, Nov. 2000.

\bibitem{EMDHat}
O.~Pele and M.~Werman, ``A linear time histogram metric for improved sift
  matching,'' in \emph{10th European Conf. Computer Vision (ECCV)}.\hskip 1em
  plus 0.5em minus 0.4em\relax Springer, Oct 2008, pp. 495--508.

\bibitem{FastEMD}
O.~Pele and M.~Werman, ``Fast and robust earth mover's distances,'' in
  \emph{12th Int'l Conf. Computer Vision (ICCV)}, Sept 2009, pp. 460--467.

\bibitem{Tatler2005}
B.~W. Tatler, R.~J. Baddeley, and I.~D. Gilchrist, ``{Visual correlates of
  fixation selection: Effects of scale and time},'' \emph{Vision Research},
  vol.~45, pp. 643--659, 2005.

\bibitem{Moorthy_Bovik}
A.~Mittal, A.~Moorthy, and A.~Bovik, ``Automatic prediction of saliency on
  {JPEG} distorted images,'' in \emph{2011 Third Int'l Workshop on Quality of
  Multimedia Experience (QoMEX)}, September 2011, pp. 195 --200.

\bibitem{NSS}
R.~J. Peters, A.~Iyer, L.~Itti, and C.~Koch, ``{Components of bottom-up gaze
  allocation in natural images.}'' \emph{Vision research}, vol.~45, no.~18, pp.
  2397--416, Aug. 2005.

\bibitem{margolin2014evaluate}
R.~Margolin, L.~Zelnik-Manor, and A.~Tal, ``How to evaluate foreground maps,''
  in \emph{IEEE Conf. Computer Vision and Pattern Recognition (CVPR)}, 2014,
  pp. 248--255.

\bibitem{KLD}
S.~Kullback and R.~A. Leibler, ``On information and sufficiency,'' \emph{The
  Annals of Mathematical Statistics}, pp. 79--86, 1951.

\bibitem{riche2013saliency}
N.~Riche, M.~Duvinage, M.~Mancas, B.~Gosselin, and T.~Dutoit, ``Saliency and
  human fixations: State-of-the-art and study of comparison metrics,'' in
  \emph{IEEE Int'l Conf. Computer Vision (ICCV)}, Dec 2013, pp. 1153--1160.

\bibitem{JSD}
J.~Lin, ``Divergence measures based on the {Shannon} entropy,'' \emph{IEEE
  Trans. Information Theory}, vol.~37, no.~1, pp. 145--151, 1991.

\bibitem{AUC}
D.~M. Green, J.~A. Swets \emph{et~al.}, \emph{Signal Detection Theory and
  Psychophysics}.\hskip 1em plus 0.5em minus 0.4em\relax Wiley New York, 1966,
  vol.~1.

\bibitem{JSDistance}
D.~Endres and J.~Schindelin, ``A new metric for probability distributions,''
  \emph{IEEE Trans. Information Theory}, vol.~49, no.~7, pp. 1858--1860, July
  2003.

\bibitem{TUD}
H.~Liu and I.~Heynderickx, ``Studying the added value of visual attention in
  objective image quality metrics based on eye movement data,'' in \emph{16th
  {IEEE} Int'l Conf. Image Processing {(ICIP)}}, Nov. 2009, pp. 3097 --3100.

\bibitem{JuddLowRes}
T.~Judd, F.~Durand, and A.~Torralba, ``{Fixations on low-resolution images},''
  \emph{J. Vision}, vol.~11, pp. 1--20, 2011.

\end{thebibliography}
\vspace{-0.4in}
\begin{IEEEbiography}[{\includegraphics[width=1in,height=1.25in,clip,keepaspectratio]{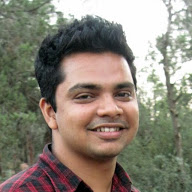}}]{Milind S. Gide} received 
	  the M.S. degree in Electrical Engineering from the University of Houston, Houston, TX, USA in 2009 . He is currently a Ph.D. student in the Image, Video, \& Usability lab and a teaching assistant in the School of Electrical, Computer \& Energy Engineering at Arizona State University. His research interests include visual attention, image quality and texture-based video coding. During Summer 2011, he worked as an intern at Sharp Labs of America, Camas, Washington in the area of texture-based video coding. 
\end{IEEEbiography}
\vspace{-0.4in}	
\begin{IEEEbiography}[{\includegraphics[width=1in,height=1.25in,clip,keepaspectratio]{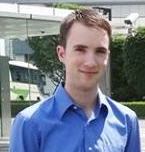}}]{Samuel F. Dodge} received the B.S.E and M.S. both in Electrical Engineering from Arizona State University, Tempe AZ USA in 2011 and 2014, respectively. In 2013 he was a visiting researcher at Peking University in Beijing, China as part of the NSF EAPSI program. He has also held internships at Intel in Chandler, Arizona and at Sony in Kanagawa, Japan working on content-based image retrieval and biomedical image processing.  He is currently a Ph.D. student in the Image, Video, \& Usability Lab at Arizona State University, and a teaching assistant in the School of Electrical, Computer and Energy Engineering. He is a recipient of the ARCS fellowship in 2013 and 2014. His research interests include computational visual saliency and applications in computer vision and image processing.  
\end{IEEEbiography}	
\vspace{-0.4in}		
\vfill
\begin{IEEEbiography}[{\includegraphics[width=1in,height=1.25in,clip,keepaspectratio]{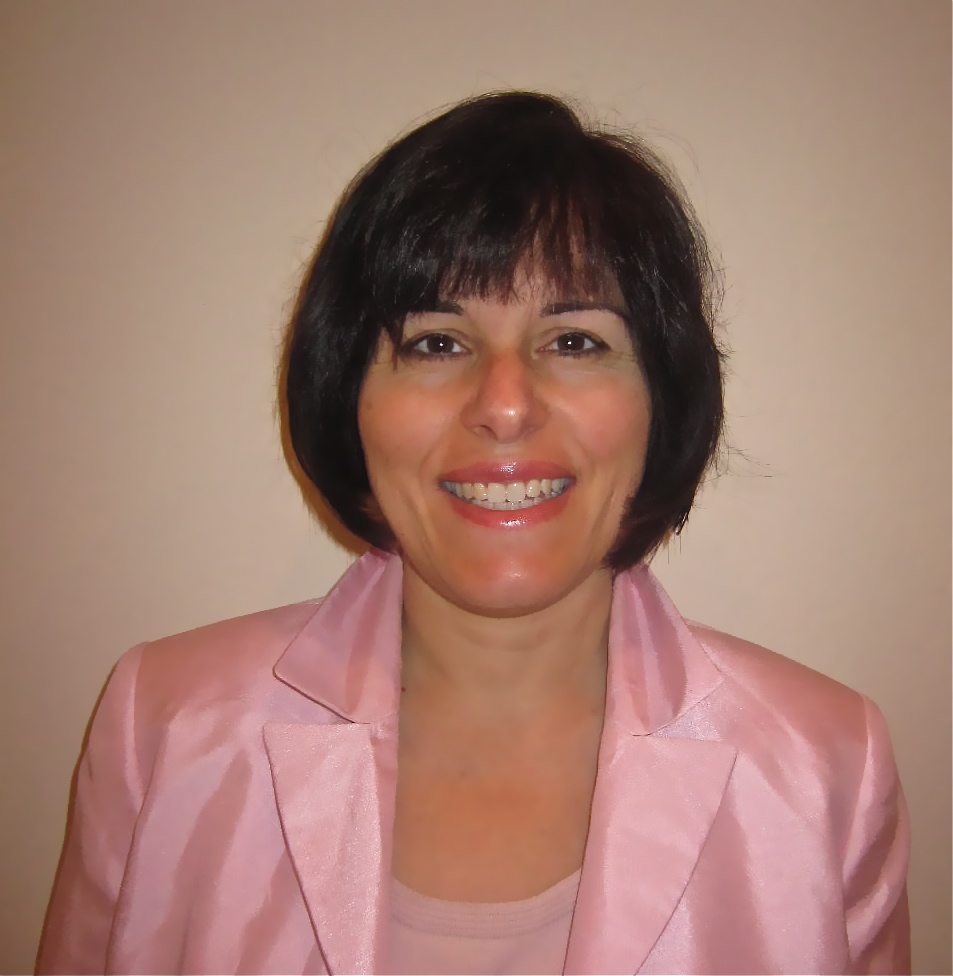}}]{Lina J. Karam} (Fellow, IEEE) received the B.E. degree in computer and communications engineering from the American University of Beirut in 1989, and the M.S. and Ph.D. degrees in electrical engineering from the Georgia Institute of Technology in 1992 and 1995, respectively. She is a Full Professor in the School of Electrical, Computer \& Energy Engineering at Arizona State University, where she directs the Image, Video, \& Usabilty (IVU) Research Laboratory. Dr. Karam was awarded a U.S. National Science Foundation CAREER Award, a NASA Technical Innovation Award, the Intel Outstanding Researcher Award, and the IEEE Phoenix Section Outstanding Faculty Award. She served on several journal editorial boards, several conference organization committees, and several IEEE technical committees. She is currently serving as the General Chair of the 2016 IEEE International Conference on Image Processing. Professor Karam is a member of the IEEE Circuits and Systems Society's DSP Technical Committee, the IEEE Signal Processing Society's (SPS) IVMSP Technical Committee, and the IEEE SPS Board of Governors.
She has over 200 technical publications and she is a co-inventor on a number of patents.
		
\end{IEEEbiography}
\vfill
\vfill
\vfill
\vfill
\end{document}